\documentclass[10pt,journal,compsoc]{IEEEtran}
\ifCLASSOPTIONcompsoc
  \usepackage[nocompress]{cite}
\else
  \usepackage{cite}
\fi
\ifCLASSINFOpdf
\else
\fi

\usepackage{stfloats}

\usepackage{caption}
\usepackage{subfigure}
\usepackage{times}
\usepackage{epsfig}
\usepackage{graphicx}
\usepackage{amsmath}
\usepackage{amssymb}
\usepackage{eso-pic}
\usepackage{helvet}
\usepackage{courier}
\usepackage{url}
\usepackage{graphicx}
\usepackage{mathrsfs}
\usepackage{multirow}
\usepackage{comment}
\usepackage{marvosym}
\usepackage{bbm}
\usepackage{enumerate}
\usepackage{soul}
\usepackage[utf8]{inputenc}
\usepackage{booktabs}
\usepackage{amsfonts}
\usepackage{nicefrac}
\usepackage{microtype}
\usepackage[numbers]{natbib}
\usepackage[breaklinks=true,bookmarks=false,colorlinks]{hyperref}
\usepackage{mathtools}
\usepackage{stmaryrd}
\usepackage{enumitem}

\newcommand{\eqn}[1]{Eq.~\eqref{#1}}

\newcommand{\tab}[1]{Table~\ref{#1}}
\newcommand{\fig}[1]{Fig.~\ref{#1}}
\newcommand{\revise}[1]{{\textcolor{black}{#1}}}

\newcommand{\myparagraph}[1]{\vspace{5pt} \noindent \textbf{#1.}}

\begin{document}


\title{PredRNN: A Recurrent Neural Network for Spatiotemporal Predictive Learning}

%
%
%
%

\author{Yunbo~Wang,
        Haixu~Wu,
        Jianjin~Zhang,
        Zhifeng~Gao,
        Jianmin~Wang,\\
        Philip~S.~Yu,~\IEEEmembership{Fellow,~IEEE},
        Mingsheng~Long,~\IEEEmembership{Member,~IEEE}%
\IEEEcompsocitemizethanks{
\IEEEcompsocthanksitem Y. Wang is with MoE Key Lab of Artificial Intelligence, AI Institute, Shanghai Jiao Tong University, China.
\IEEEcompsocthanksitem H. Wu, Z. Gao, J. Wang, P. S. Yu, and M. Long are with the School of Software, BNRist, Tsinghua University, China.
\IEEEcompsocthanksitem J. Zhang is with the Microsoft Corporation, China.
\IEEEcompsocthanksitem Y. Wang and H. Wu contributed equally to this work.
\IEEEcompsocthanksitem Corresponding author: Mingsheng Long, mingsheng@tsinghua.edu.cn.
}
}

\markboth{IEEE Transactions on Pattern Analysis and Machine Intelligence,~Vol.~XX, No.~X, March~2021}%
{Wang \MakeLowercase{\textit{et al.}}: PredRNN: A Recurrent Neural Network for Spatiotemporal Predictive Learning}

\IEEEtitleabstractindextext{%
\begin{abstract}

The predictive learning of spatiotemporal sequences aims to generate future images by learning from the historical context, where the visual dynamics are believed to have modular structures that can be learned with compositional subsystems. This paper models these structures by presenting PredRNN, a new recurrent network, in which a pair of memory cells are explicitly decoupled, operate in nearly independent transition manners, and finally form unified representations of the complex environment. Concretely, besides the original memory cell of LSTM, this network is featured by a zigzag memory flow that propagates in both bottom-up and top-down directions across all layers, enabling the learned visual dynamics at different levels of RNNs to communicate. It also leverages a memory decoupling loss to keep the memory cells from learning redundant features. We further propose a new curriculum learning strategy to force PredRNN to learn long-term dynamics from context frames, which can be generalized to most sequence-to-sequence models. We provide detailed ablation studies to verify the effectiveness of each component. Our approach is shown to obtain highly competitive results on five datasets for both action-free and action-conditioned predictive learning scenarios.

\end{abstract}

}

\maketitle

\IEEEdisplaynontitleabstractindextext

%
\IEEEpeerreviewmaketitle

\section{Introduction}

As a key application of predictive learning, generating future frames from historical consecutive frames has received growing interest in machine learning and computer vision communities. 
It benefits many practical applications and downstream tasks, such as the precipitation forecasting \cite{shi2015convolutional,wang2017predrnn}, traffic flow prediction \cite{xu2018predcnn,wang2019memory}, physical scene understanding \cite{wu2017learning,van2018relational,kipf2018neural,xu2019unsupervised}, early activity recognition \cite{wang2019eidetic}, deep reinforcement learning \cite{ha2018recurrent,hafner2018learning}, and the vision-based model predictive control \cite{finn2017deep,ebert2017self}.
Many of these existing approaches suggested leveraging RNNs \cite{rumelhart1988learning,werbos1990backpropagation} with stacked LSTM units \cite{hochreiter1997long} to capture the temporal dependencies of spatiotemporal data.
This architecture is mainly inspired by similar ideas from other well-explored tasks of sequential data, such as neural machine translation \cite{Sutskever2011Generating,Cho2014On}, speech recognition \cite{Graves2014Towards}, video action recognition \cite{Ng15,donahue2015long}, and video captioning \cite{donahue2015long}.


For stacked LSTMs, a network structure named \textit{memory cell} plays an important role in alleviating the vanishing gradient problem of RNNs. Strong theoretical and empirical evidence has shown that it can latch the gradients of hidden states inside each LSTM unit in the training process and thereby preserve valuable information of underlying temporal dynamics \cite{hochreiter1997long}.
However, the state transition pathway of LSTM memory cells may not be optimal for spatiotemporal predictive learning, as this task requires different focuses on the learned representations in many aspects from other tasks of sequential data.
First, most predictive networks for language or speech modeling \cite{Sutskever2011Generating,Cho2014On,Graves2014Towards} focus on capturing the long-term, non-Markovian properties of sequential data, rather than spatial deformations of visual appearance. But for future frames prediction, both data structures in space-time are crucial and need to be carefully considered.
Second, in other supervised tasks of video data, such as action recognition, high-level semantical features can be informative enough, and the low-level features are less important to final outputs. Due to the absence of complex structures of supervision signals, the stacked LSTMs don't need to preserve fine-grained representations from the bottom up. 
Although the existing recurrent architecture based on inner-layer memory transitions can be sufficient to capture temporal variations at each level of the network, it may not be the best choice for predictive learning, where low-level details and high-level semantics of spatiotemporal data are both significant to generating future frames.

To jointly model the spatial correlations and temporal dynamics at different levels of RNNs, we propose a new memory-prediction framework named Predictive Recurrent Neural Network (PredRNN), which extends the inner-layer transition function of memory states in LSTMs to \textit{spatiotemporal memory flow}.
The spatiotemporal memory flow traverses all nodes of PredRNN in a zigzag path of bi-directional hierarchies:
At each timestep, the low-level information is delivered vertically from the input to the output via a newly-designed memory cell, while at the top layer, the spatiotemporal memory flow brings the high-level memory state to the bottom layer at the next timestep.
By this means, the top-down expectations interact with the bottom-up signals to both analyze those inputs and generate predictions of subsequently expected inputs, which is different from stacked LSTMs, where the memory state is latched inside each recurrent unit.

Accordingly, we define the central building block of PredRNN as the Spatiotemporal LSTM (ST-LSTM), in which the proposed spatiotemporal memory flow interacts with the original, unidirectional memory state of LSTMs. 
The intuition is that if we expect a vivid imagination of multiple future images, we need a unified memory mechanism to cope with both short-term deformations of spatial details and long-term dynamics: On one hand, the new design of the spatiotemporal memory cell enables the network to learn complex transition functions within short neighborhoods of consecutive frames by increasing the depth of non-linear neurons between time-adjacent RNN states. It thus significantly improves the modeling capability of ST-LSTM for short-term dynamics. 
On the other hand, ST-LSTM still retains the temporal memory cell of LSTMs and closely combines it with the proposed spatiotemporal memory cell, in pursuit of both long-term coherence of hidden states and their quick response to short-term dynamics.

This journal paper extends our previous work \cite{wang2017predrnn} in three technical aspects:
First, we propose a decoupling loss to maximize the distance of memory states between the two memory cells in ST-LSTM. It is largely inspired by the theoretical argument that distributed representation has a great potential to match the underlying properties of data distribution \cite{pascanu2013number}. In other words, we implicitly train the two memory cells to focus on different aspects of spatiotemporal variations.
Second, we extend PredRNN to support action-conditioned video prediction, such that it models the effects of various actions taken by an agent in high-dimensional pixel observations.
\revise{
Third, another challenge we observed for video prediction is that the widely used sequence-to-sequence training procedure \cite{Sutskever2011Generating} prevents the models from learning long-term dynamics in the context frames.
To this end, we introduce a new curriculum learning procedure named Reverse Scheduled Sampling, which forces our model to learn more about \textit{jumpy} frame dependencies by randomly hiding real observations in the input sequence with certain probabilities changing over the training phase. 
We believe that the proposed training procedure can be easily extended to other video prediction models that also follow the sequence-to-sequence framework \cite{Sutskever2011Generating}.
}

Our approach achieves state-of-the-art performance on five datasets: the Moving MNIST dataset, the KTH action dataset, a radar echo dataset for precipitation forecasting, the Traffic4Cast dataset of high-resolution traffic flows, and the action-conditioned BAIR dataset with robot-object interactions. The last two of them are new in this version of the manuscript.
We perform ablation studies to understand the effectiveness of each component of PredRNN, which is complementary to existing methods and further improves them when integrated.
We release the code to facilitate future research at \url{https://github.com/thuml/predrnn-pytorch}.

\section{Related work}

\begin{table*}[t]
  \caption{
  The evolution of convolutional recurrent units in video prediction models. 
  %
  \revise{The ``\textit{Standard}'' training scheme is to follow the conventional sequence-to-sequence framework \cite{Sutskever2014Sequence} that feeds the historical real observations to the sequence encoder and feeds the previously generated frames to the decoder during the entire training phase. ``\textit{SS}'' is short for \textit{scheduled sampling} \cite{bengio2015scheduled}.}
  }
  \vskip -0.05in
  \label{tab:model_compare}
  \centering
  \begin{tabular}{llll}
    \toprule
    Model & Architecture & Recurrent unit & \revise{Training (Encoder + Decoder)} \\
    \midrule
    ConvLSTM \cite{shi2015convolutional} & Seq-to-Seq with stacked layers \cite{Sutskever2014Sequence} & Convolutional LSTM & \revise{Standard \cite{Sutskever2014Sequence} + Standard \cite{Sutskever2014Sequence}}  \\
    TrajGRU \cite{shi2017deep} & Seq-to-Seq with reversed forecasting layers & Trajectory GRU & \revise{Standard + Standard} \\
    CDNA \cite{Finn2016Unsupervised} & Action-conditioned ConvLSTM network & ConvLSTM \cite{shi2015convolutional} & \revise{Standard + SS \cite{bengio2015scheduled}} \\
    PredRNN \cite{wang2017predrnn} (\textit{Conf. version}) & Spatiotemporal memory flow & Spatiotemporal LSTM & \revise{Standard + SS} \\
    PredRNN++ \cite{wang2018predrnn++} & Gradient highway & Causal LSTM & \revise{Standard + SS} \\
    E3D-LSTM \cite{wang2019eidetic} & Temporal self-attention on memory cells & Eidetic 3D LSTM & \revise{Standard + SS } \\
    CrevNet \cite{yu2020efficient} & Two-way autoencoder + Reversible prediction & ConvLSTM \cite{shi2015convolutional}; ST-LSTM \cite{wang2017predrnn} & \revise{Standard + Standard} \\
    Conv-TT-LSTM \cite{su2020convolutional} & Seq-to-Seq with stacked layers & Conv. Tensor-Train LSTM & \revise{Standard + SS } \\
    PredRNN-V2 (\textit{Ours}) & Spatiotemporal memory flow & ST-LSTM (decoupled memory) & \revise{Reverse scheduled sampling + SS }\\
    \bottomrule
  \end{tabular}
  \vspace{-10pt}
\end{table*}

For spatiotemporal predictive learning, different inductive biases are encoded into neural networks by using different network architectures \cite{oprea2020review}, which, in general, can be roughly divided into three groups: feed-forward models based on CNNs  \cite{Oh2015Action,Mathieu2015Deep,tulyakov2018mocogan}, recurrent models \cite{srivastava2015unsupervised,shi2015convolutional,babaeizadeh2017stochastic}, and other models including the combinations of the convolutional and recurrent networks, as well as the Transformer-based and flow-based methods \cite{weissenborn2019scaling,kumar2019videoflow}.

\myparagraph{Feed-forward models based on CNNs}
The use of convolutional layers has introduced the inductive bias of group invariance over space to spatiotemporal prediction.
Oh \textit{et al.} \cite{Oh2015Action} defined a convolutional autoencoder for next-frame prediction in Atari games. 
Xue \textit{et al.} \cite{xue2016visual}
presented a probabilistic model that encodes motion information as convolutional kernels and learns to produce a probable set of future frames by learning their conditional distribution.
Zhang \textit{et al.} \cite{zhang2017deep} used a CNN with residual connections for traffic prediction, which considers the closeness, period, trend, and external factors of the traffic flows.
Recent literature proposed to train the predictive CNNs with adversarial learning \cite{Goodfellow2014Generative,Denton2015Deep} to reduce the uncertainty of the learning process and improve the sharpness of the generated frames \cite{Mathieu2015Deep,bhattacharjee2017temporal,liang2017dual,tulyakov2018mocogan,wu2020future,gur2020hierarchical,liu2021deep}.
Compared with the recurrent models, feed-forward models typically show higher computational efficiency on large-scale GPUs.
However, these models learn complicated state transitions as compositions of simpler ones by stacking convolutional layers, thus often failing to capture long-term dependency between distant frames.

\myparagraph{Recurrent models}
Recent advances in RNNs typically show a greater ability to model the dynamics of historical observations.
Ranzato \textit{et al.} \cite{Ranzato2014Video} defined an RNN inspired by language modeling, predicting future frames in a discrete space of patch clusters. 
Srivastava \textit{et al.} \cite{srivastava2015unsupervised} proposed a sequence-to-sequence video prediction model borrowed from neural machine translation \cite{Sutskever2011Generating}.
Later on, some work improved the modeling of temporal relations and extended the prediction horizon by organizing 2D recurrent states in hierarchical architectures \cite{villegas2017learning,wichers2018hierarchical,kim2019variational}.
To model the temporal uncertainty, some work proposed to integrate variational inference with 2D recurrence
\cite{babaeizadeh2017stochastic,denton2018stochastic,lee2018stochastic,villegas2019high,castrejon2019improved,franceschi2020stochastic,wu2021greedy}. 
Another line of work is to factorize the content and motion of videos. 
%
Typical approaches include using optical flows \cite{Villegas2017Decomposing,bei2021learning}, adversarial training schemes \cite{denton2017unsupervised}, reasoning about object-centric content and pose vectors \cite{hsieh2018learning,bodla2021hierarchical}, differentiable clustering methods \cite{van2018relational}, amortized inference \cite{zablotskaia2020unsupervised,greff2019multi}, and solvers of neural differential equations \cite{guen2020disentangling}.
However, the above methods 
mainly used 2D RNNs to model video dynamics in low-dimensional space, which inevitably causes the loss of visual information in real scenarios. In this paper, we are more focused on improving the spatiotemporal modeling capability of the recurrent predictive backbones.

\myparagraph{Convolutional RNNs}
\tab{tab:model_compare} summarizes the evolution of recent video prediction models that combine the advantages of convolutional and recurrent architectures. 
Shi \textit{et al.} \cite{shi2015convolutional} proposed the Convolutional LSTM (ConvLSTM), which uses convolutions to replace the matrix multiplication in the recurrent transitions of the original LSTM. 
Finn \textit{et al.} \cite{Finn2016Unsupervised} designed an action-conditioned ConvLSTM network for visual planning and control.
Shi \textit{et al.} \cite{shi2017deep} combined convolutions with GRUs \cite{Cho2014On} and extended the receptive fields of state-to-state transitions with non-local neural connections. 
Wang \textit{et al.} \cite{wang2019eidetic} enriched each RNN state by incorporating a time dimension that covers a short snippet, and proposed to model the dynamics using 3D convolutions and temporal self-attention. 
Su \textit{et al.} \cite{su2020convolutional} improved the efficiency of higher-order ConvLSTMs based on low-rank tensor factorization. 
In summary, the convolutional RNNs jointly model the visual appearances and temporal dynamics, and is also a foundation for the follow-up approaches \cite{patraucean2015spatio,Finn2016Unsupervised,Lotter2016Deep,Kalchbrenner2016Video,shi2017deep,wang2018predrnn++,byeon2018contextvp,oliu2018folded,xu2018structure,wang2019memory,wang2019eidetic,yu2020efficient,su2020convolutional,wu2021motionrnn}. 
This paper improves upon the existing video prediction models with (i) a zigzag recurrent transition mechanism named \textit{spatiotemporal memory flow}, (ii) a new convolutional recurrent unit with a pair of decoupled memory cells, and (iii) a new training procedure for sequence-to-sequence predictive learning. 

\section{Preliminaries}

\subsection{Spatiotemporal Predictive Learning}

Suppose we are monitoring a dynamical system of $J$ measurements over time, where each measurement is recorded at all locations in a spatial region represented by an $M \times N$ grid. 
From the spatial view, the observation of these $J$ measurements at any time can be represented by a tensor of $\mathcal{X} \in \mathbb{R}^{J \times M \times N}$. 
%
From the temporal view, the observations over $T$ timesteps form a sequence of $\mathcal{X}_\text{in}=\{\mathcal{X}_1,\ldots,\mathcal{X}_T\}$. 
Given $\mathcal{X}_\text{in}$, the spatiotemporal predictive learning is to predict the most probable length-$K$ sequence in the future, $\widehat{\mathcal{X}}_\text{out}=\{\widehat{\mathcal{X}}_{T+1},\ldots,\widehat{\mathcal{X}}_{T+K}\}$.
In this paper, we train neural networks parametrized by $\theta$. Concretely, we use stochastic gradient descent to find a set of parameters $\theta^{\star}$ that maximizes the log-likelihood of producing the true target sequence $\mathcal{X}_\text{out}$ given the input data $\mathcal{X}_\text{in}$ for all training pairs $\{(\mathcal{X}_\text{in}^n,\mathcal{X}_\text{out}^n)\}_n$:
\begin{equation}
\begin{split}
   \theta^{\star} & = \mathop {\arg \max }\limits_{\theta} \sum_{(\mathcal{X}_\text{in}^n,\mathcal{X}_\text{out}^n)} \log P(\mathcal{X}_\text{out}^n | \mathcal{X}_\text{in}^n; \theta).
\end{split}
\end{equation}


In this paper, we take video prediction as a typical experimental domain, where the observed data at each timestep is an RGB image and the number of measured channels is $3$.
Another domain is precipitation nowcasting, where the observed data is a sequence of radar echo maps in a certain geographic region\footnote{
We usually visualize radar echoes by mapping them to color images.}. 

\subsection{Convolutional LSTM}

The Convolutional LSTM (ConvLSTM) \cite{shi2015convolutional} models the spatial and temporal data structures simultaneously by explicitly encoding the spatial information into tensors, and applying convolution operators to both the state-to-state and input-to-state recurrent transitions.
It overcomes the limitation of vector-variate representations in standard LSTM, where the spatial correlation is not explicitly modeled. 
The input state $\mathcal{X}_t$, memory state $\mathcal{C}_t$, and hidden state $\mathcal{H}_t$ at each timestep are 3D tensors in $\mathbb{R}^{J \times M \times N}$. The first dimension is either the number of measurements (for input states) or the number of feature maps (otherwise), and the last two dimensions are the numbers of spatial rows $M$ and columns $N$. 
Like standard LSTM, the input gate $i_t$, forget gate $f_t$, output gate $o_t$, and input-modulation gate $g_t$ control the information flow across $\mathcal{C}_t$, such that the gradient will be kept from quickly vanishing by being trapped in the memory cell.
ConvLSTM determines the future state of a certain cell in the $M \times N$ grid based on the input frame and past states of its local neighbors:
\begin{equation}\label{equ:convlstm}
\begin{split}
    g_t & = \tanh(W_{xg} \ast \mathcal{X}_t+W_{hg} \ast \mathcal{H}_{t-1}+b_g) \\
  i_t & = \sigma(W_{xi} \ast \mathcal{X}_t+W_{hi} \ast \mathcal{H}_{t-1} + {W_{ci}} \odot {\mathcal{C}_{t - 1}} + b_i) \\
  f_t & = \sigma(W_{xf} \ast \mathcal{X}_t+W_{hf} \ast \mathcal{H}_{t-1} + {W_{cf}} \odot {\mathcal{C}_{t - 1}} +b_f) \\
  \mathcal{C}_t & = f_t \odot \mathcal{C}_{t-1}+i_t \odot g_t \\
  o_t & = \sigma(W_{xo} \ast \mathcal{X}_t+W_{ho} \ast \mathcal{H}_{t-1} + {W_{co}} \odot {\mathcal{C}_t} +b_o) \\
  \mathcal{H}_t & = o_t \odot \tanh(\mathcal{C}_t),
\end{split}
\end{equation}
where $\sigma$ is the Sigmoid activation function, $\ast$ and $\odot$ denote the convolution operator and the Hadamard product. 
%
%
We observe that the ConvLSTM network can be improved from three aspects.

\myparagraph{Challenge I}
For a stacked ConvLSTM network, the input sequence $\mathcal{X}_\text{in}$ is fed into the bottom layer, and the output sequence $\mathcal{X}_\text{out}$ is generated at the top one. With hidden states $\mathcal{H}_t$ being delivered from the bottom up, spatial representations are encoded layer by layer. However, the memory states $\mathcal{C}_t$ are merely updated along the arrow of time within each ConvLSTM layer, being less dependent on the hierarchical visual features at other layers.
Thus, the first layer at the current timestep may largely ignore what had been memorized by the top layer at the previous timestep. 


\myparagraph{Challenge II}
In ConvLSTM, the output hidden state $\mathcal{H}_t$ is dependent on the memory state $\mathcal{C}_t$ and the output gate $o_t$, which means that the memory cell is forced to cope with long-term and short-term dynamics simultaneously. Therefore, the modeling capability of $\mathcal{C}_t$ may greatly limit the overall performance of the model for complex spatiotemporal variations.

\myparagraph{Challenge III}
\revise{The ConvLSTM network follows the training procedure of sequence-to-sequence RNNs \cite{Sutskever2014Sequence}. During training, it always takes as inputs real observations at encoding timesteps but has to rely more on the long-term information from the previous context frames at the forecasting timesteps.
Since both the encoding and forecasting parts of ConvLSTM use the same set of model parameters, the one-step training in the encoding phase may prevent the forecaster from learning the jumpy frame dependencies, thus affecting the performance of long-term prediction. 
}

\begin{figure*}[t]
  \centering
  \includegraphics[width=0.9\textwidth]{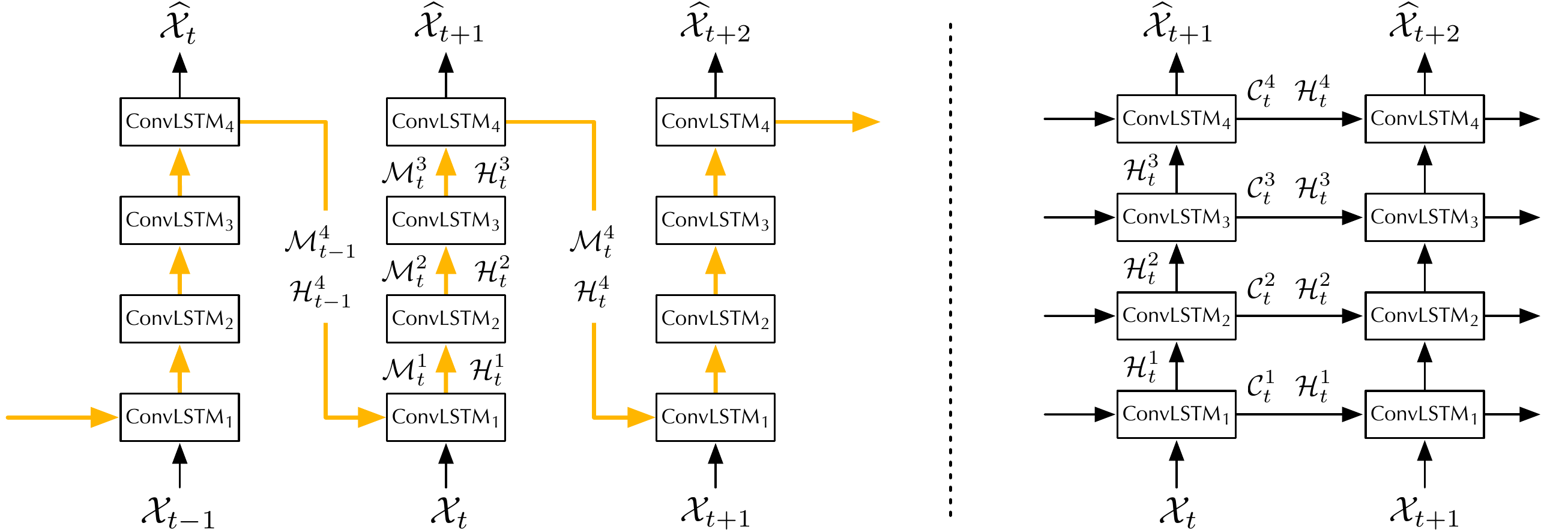}
  \vskip -0.1in
  \caption{\textbf{Left:} the spatiotemporal memory flow architecture that uses ConvLSTM as the building block. The orange arrows show the deep-in-time path of memory state transitions. \textbf{Right:} the original ConvLSTM network proposed by Shi \textit{et al.} \cite{shi2015convolutional}.
  }
  \label{fig:rnn_compare}
  \vspace{-10pt}
\end{figure*}

\section{Method}
To tackle the above challenging problems, we propose the Predictive Recurrent Neural Network (PredRNN).
For \textbf{Challenge I}, we present a new RNN architecture named the spatiotemporal memory flow, which improves the state transition functions of convolutional RNNs. 
For \textbf{Challenge II}, we propose the Spatiotemporal LSTM (ST-LSTM) which serves as the building block of PredRNN. It leverages a pair of memory cells that are jointly learned and explicitly decoupled to cover long- and short-term dynamics of spatiotemporal variations.
For \textbf{Challenge III}, we improve the training procedure to encourage the predictive network to learn non-Markovian dynamics from longer periods of observation frames.
Furthermore, we present an \textbf{action-conditioned} PredRNN, which allows simulating the spatiotemporal variations of the environment in response to the actions of the agent in decision-making scenarios.

\subsection{Spatiotemporal Memory Flow}

PredRNN employs a stack of convolutional recurrent units to learn unified representations of the spatial correlations and temporal dynamics from the input sequence, and then transforms these features back to the data space to predict future spatiotemporal frames. 
We initially adopt the original ConvLSTM layer as the basic building block of PredRNN and apply a novel memory state transition method between the recurrent nodes.
In the original ConvLSTM network \cite{shi2015convolutional} shown in \fig{fig:rnn_compare} (right), the memory states $\mathcal{C}_t^l$ are constrained inside individual recurrent layers and updated along the arrow of time. 
Only the hidden states $\mathcal{H}_t^l$ can be delivered from the observation to the final prediction.
This temporal memory flow has been widely used in supervised learning tasks such as video classification, where the hidden representations are more and more abstract and class-specific, starting at the bottom.
However, in predictive learning, the input frames and the expected outputs share the same data space, \emph{i.e.}, they may have very close data distributions in the spatial domain and very related underlying dynamics in the temporal domain.
Therefore, it becomes important to make the predictions more effectively dependent on the memory representations learned at different levels of recurrent layers. 
If we want to frame the future vividly, we need both high-level understandings of global motions and more detailed memories of the subtle variations in the input sequence.

Considering that the memory cell of ConvLSTM can latch the gradients\footnote{Like the standard LSTM, the memory cell of ConvLSTM was originally designed to alleviate the gradient vanishing problem during training.} and thereby store valuable information across recurrent nodes, we improve the above horizontal memory flow by updating the memory state in the zigzag direction, so that it can better deliver knowledge from input to output. 
We show the key equations of this memory flow using ConvLSTM as the building block:
\begin{equation}\label{equ:flow}
  \begin{split}
  g_t & = \tanh(W_{xg} \ast \mathcal{X}_t\mathbbm{1}_{\{l=1\}} + W_{hg} \ast \mathcal{H}_t^{l-1}) \\
  i_t & = \sigma(W_{xi} \ast \mathcal{X}_t\mathbbm{1}_{\{l=1\}} + W_{hi} \ast \mathcal{H}_t^{l-1} + W_{ci} \ast \mathcal{M}_t^{l-1}) \\
  f_t & = \sigma(W_{xf} \ast \mathcal{X}_t\mathbbm{1}_{\{l=1\}} + W_{hf} \ast \mathcal{H}_t^{l-1} + W_{cf} \ast \mathcal{M}_t^{l-1}) \\
  \mathcal{M}_t^l & = f_t \odot \mathcal{M}_t^{l-1} + i_t \odot g_t \\
  o_t & = \sigma(W_{xo} \ast \mathcal{X}_t\mathbbm{1}_{\{l=1\}} + W_{ho} \ast \mathcal{H}_t^{l-1} + W_{co} \ast \mathcal{M}_t^l) \\ 
  \mathcal{H}_t^l & = o_t \odot \tanh(\mathcal{M}_t^l).
  \end{split} 
\end{equation}

We name this new memory state transition method the \textit{spatiotemporal memory flow}, and show its transition direction by the orange arrows in \fig{fig:rnn_compare} (left).
In this way, the memory states in different ConvLSTM layers are no longer independent, and all nodes in the entire recurrent network jointly maintain a memory bank denoted by $\mathcal{M}$. 
The input gate, input modulation gate, forget gate, and output gate are no longer dependent on the hidden state and the temporal memory state from the previous timestep at the same layer. Instead, they rely on the hidden state $\mathcal{H}_t^{l-1}$ and the spatiotemporal memory state $\mathcal{M}_t^{l-1} (l \in \{1,...,L\})$ supplied by the previous layer at current timestep (see \fig{fig:rnn_compare} (left)). 
In particular, the bottom recurrent unit ($l=1$) receives state values from the top layer at the previous timestep: $\mathcal{H}_t^{l-1}=\mathcal{H}_{t-1}^L$, $\mathcal{M}_t^{l-1}=\mathcal{M}_{t-1}^L$. The four layers in this figure have different sets of convolutional parameters regarding the input-to-state and state-to-state transitions. 
They thereby read and update the values of the memory state based on their individual understandings of the spatiotemporal dynamics as the information flows through the current node.
Note that we replace the notation for memory state from $\mathcal{C}$ to $\mathcal{M}$ to emphasize that it flows in the zigzag direction in PredRNN, instead of the horizontal direction in standard recurrent networks. 
Different from ConvLSTM which uses Hadamard product $\odot$ for the computation of the gates, we adopt convolution operators $\ast$ for finer-grained memory transitions. An additional benefit of this change is that the learned PredRNN can be deployed directly on the input sequence of different spatial resolutions.

The spatiotemporal memory flow provides a recurrent highway for hierarchical visual representations that can reduce the loss of information from the bottom layers to the top of the network. 
Besides, by introducing more nonlinear gated neurons within temporal neighborhoods, it expands the deep transition path of hidden states and enhances the modeling capability of the network for short-term dynamics.
In contrast, the original ConvLSTM network requires larger convolution kernels for input-to-state and state-to-state transitions in order to capture faster motion, resulting in an unnecessary increase in model parameters.

Alternatively, we can understand the spatiotemporal memory flow from the perspective of \textit{memory networks} \cite{graves2014neural,sukhbaatar2015end,graves2016hybrid}. Here, the proposed spatiotemporal memory state $\mathcal{M}$ can be viewed as a simple version of the so-called \textit{external memory} with continuous memory representations, and the stacked recurrent layers can be viewed as multiple computational steps. The layer-wise forget gates, input gates, and input modulation gates respectively determine the read and write functions, as well as the content to be written. One advantage of the classic \textit{memory networks} is to capture long-term structure (even with multiple temporal hops) within sequences. Our spatiotemporal memory flow is analogous to their mechanism, as it enables our model to consider different levels of video representations before making a prediction.

\begin{figure*}[t]
  \centering
  \includegraphics[width=\textwidth]{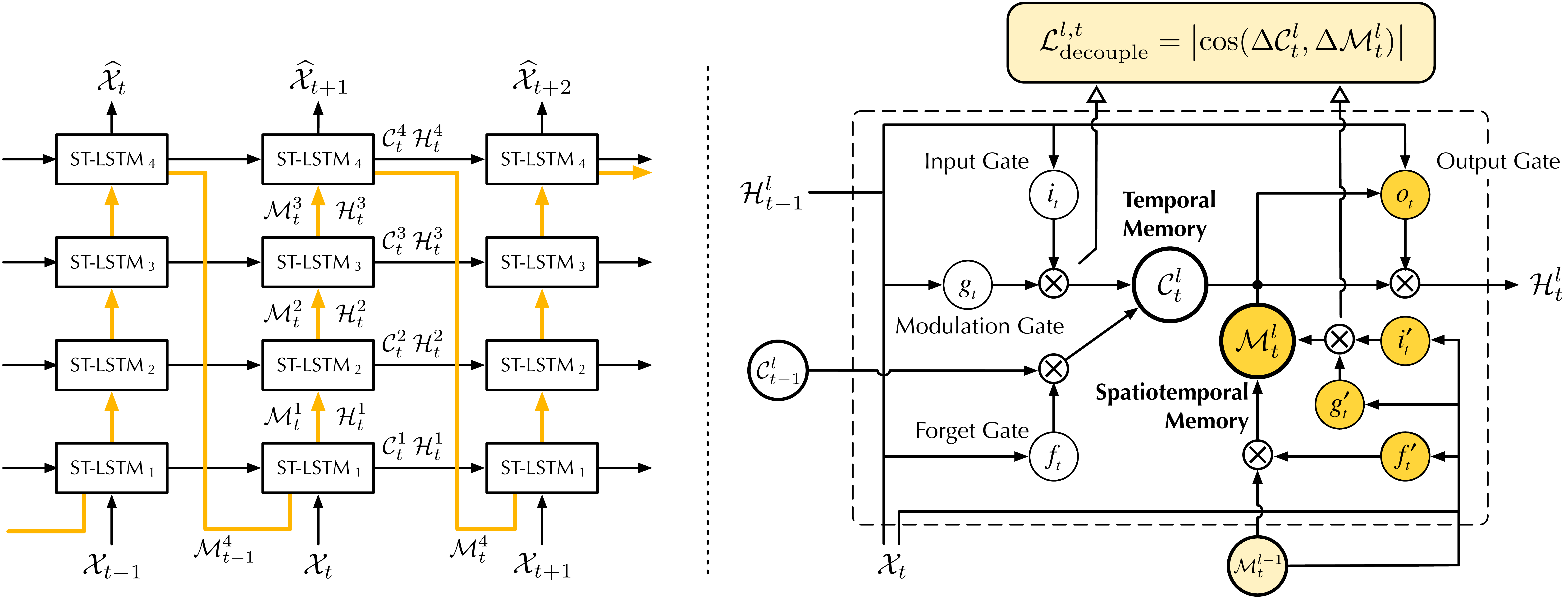}
  \vskip -0.1in
  \caption{\textbf{Left:} the main architecture of PredRNN, in which the orange arrows denote the state transition paths of $\mathcal{M}_t^l$, namely the spatiotemporal memory flow.
  \textbf{Right:} the ST-LSTM unit with twisted memory states that serves as the building block of the proposed PredRNN, where the orange circles denote the unique structures compared with ConvLSTM.
  }
  \label{fig:st-lstm}
  \vspace{-10pt}
\end{figure*}

\subsection{Spatiotemporal LSTM with Memory Decoupling}

As described previously, the spatiotemporal memory state is updated first upwards across layers then forwards to the next timestep.
It stretches the state transition path across time and adds extra neurons between horizontally adjacent nodes at the same level of the network. It thus enables the network to learn complex non-linear transition functions of short-term motions.
However, this deep-in-time architecture may also bring in the gradient vanishing problem. The roundabout memory transition path may make it difficult to capture long-term dependencies. 
For both short-term recurrence depth and long-term coherence, we introduce a double-flow memory transition mechanism that combines the original memory cell $\mathcal{C}$ and the new memory cell $\mathcal{M}$, which derives a recurrent unit named Spatiotemporal LSTM (ST-LSTM): 
\begin{equation}\label{equ:stlstm}
  \begin{split}
  g_t & = \tanh(W_{xg} \ast \mathcal{X}_t + W_{hg} \ast \mathcal{H}_{t-1}^l) \\
  i_t & = \sigma(W_{xi} \ast \mathcal{X}_t + W_{hi} \ast \mathcal{H}_{t-1}^l) \\
  f_t & = \sigma(W_{xf} \ast \mathcal{X}_t + W_{hf} \ast \mathcal{H}_{t-1}^l) \\
  \mathcal{C}_t^l & = f_t \odot \mathcal{C}_{t-1}^l + i_t \odot g_t \\
  g_t^\prime & = \tanh(W_{xg}^\prime \ast \mathcal{X}_t + W_{mg} \ast \mathcal{M}_t^{l-1}) \\
  i_t^\prime & = \sigma(W_{xi}^\prime \ast \mathcal{X}_t + W_{mi} \ast \mathcal{M}_t^{l-1}) \\
  f_t^\prime & = \sigma(W_{xf}^\prime \ast \mathcal{X}_t + W_{mf} \ast \mathcal{M}_t^{l-1}) \\
  \mathcal{M}_t^l & = f_t^\prime \odot \mathcal{M}_t^{l-1} + i_t^\prime \odot g_t^\prime \\
  o_t & = \sigma(W_{xo} \ast \mathcal{X}_t + W_{ho} \ast \mathcal{H}_{t-1}^l + W_{co} \ast \mathcal{C}_t^l + W_{mo} \ast \mathcal{M}_t^l) \\ 
  \mathcal{H}_t^l & = o_t \odot \tanh(W_{1\times1} \ast [\mathcal{C}_t^l, \mathcal{M}_t^l]).\\
  \end{split} 
\end{equation}

In \fig{fig:st-lstm}, we present the final PredRNN model by taking ST-LSTM as the building block in place of ConvLSTM. 
There are two memory states: First, $\mathcal{C}_t^l$ is the temporal memory that transits within each ST-LSTM unit from the previous node at $t-1$ to the current timestep. 
We adopt the original gates for $\mathcal{C}_t^l$ from the standard LSTM, but remove the Hadamard terms $W_{\bullet\bullet} \odot \mathcal{C}_{t-1}^l$ from the computation of $i_t$ and $f_t$, due to empirical performance and model efficiency.
Second, $\mathcal{M}_t^l$ is the spatiotemporal memory, which transits vertically to the current node from the lower $l-1$ ST-LSTM unit at the same timestep. 
In particular, we assign $\mathcal{M}_{t-1}^L$ to $\mathcal{M}_t^0$ for the bottom ST-LSTM where $l=1$. 
We construct another set of input gate $i_t^\prime$, forget gate $f_t^\prime$, and input modulation gate $g_t^\prime$ for $\mathcal{M}_t^l$, because the memory transition functions in distinct directions are supposed to be controlled by different signals.

The final hidden states $\mathcal{H}_t^l$ of each node are dependent on a combination of the horizontal and zigzag memory states: we concatenate the $\mathcal{C}_t^l$ and $\mathcal{M}_t^l$, and then apply the $1\times1$ convolutional layer for dimensionality reduction, which makes the hidden state $\mathcal{H}_t^l$ have the same dimensions as the memory states. 
In addition to simple concatenation, pairs of memory states are finally twisted and unified by output gates with bidirectional control signals (horizontal and vertical), resulting in comprehensive modeling of long-term and short-term dynamics.
This dual-memory mechanism benefits from the compositional advantage with distributed representations \cite{hinton1984distributed,bengio2000taking}.  
Intuitively, due to different mechanisms of state transitions, the pair of memory cells in ST-LSTM are expected to deal with different aspects of spatiotemporal variations:
\begin{itemize}[leftmargin=*]
    \item $\mathcal M$ introduces a deeper transition path that zigzags across ST-LSTM units. With the forget gate $f_t^\prime$ and the input-related modules $i_t^\prime \odot g_t^\prime$, it improves the ability to model complex short-term dynamics from one timestep to the next, and allows $\mathcal H$ to transit adaptively at different rates.
    \item $\mathcal C$ operates on a slower timescale. It provides a shorter gradient path between distant hidden states, thus facilitating the learning process of long-term dependencies.
\end{itemize}

\begin{figure}[t]
  \centering
  \includegraphics[width=\columnwidth]{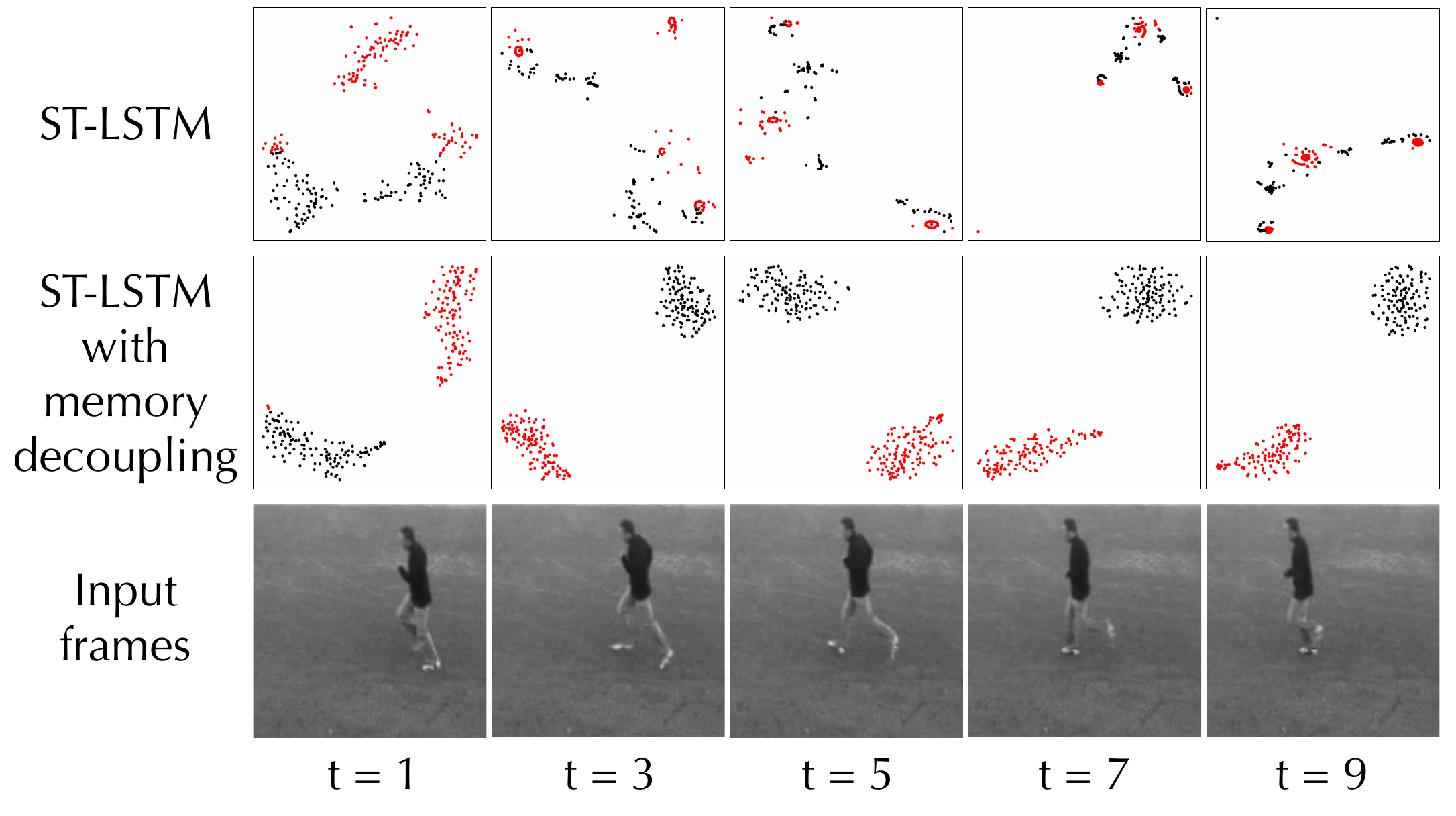}
  \vskip -0.1in
  \caption{Visualization of $\Delta \mathcal{C}_{t}^{l}$ (red points) and $\Delta \mathcal{M}_{t}^{l}$ (black points) using t-SNE \cite{maaten2008visualizing} on the KTH action dataset. Models are respectively trained without or with memory decoupling.}
  \label{fig:decouple}
  \vspace{-10pt}
\end{figure}

However, as shown in \fig{fig:decouple}, we visualized the increments of memory states at each timestep (\textit{i.e.}, $\Delta \mathcal{C}_{t}^{l}$ and $\Delta \mathcal{M}_{t}^{l}$) using t-SNE \cite{maaten2008visualizing}, and found that they were not automatically separated as expected.
In fact, the two memory states are often so intertwined that they are difficult to decouple spontaneously through their respective network architectures.
To some extent, it results in the inefficient utilization of network parameters.

Building upon the first version of ST-LSTM \cite{wang2017predrnn}, we present a new decoupling loss, as shown in \fig{fig:st-lstm} (right), that keeps $\mathcal{C}$ and $\mathcal{M}$ from learning redundant features.
In each ST-LSTM unit, we first add the convolutional layer upon the increments of $\mathcal{C}_t^l$ and $\mathcal{M}_t^l$ at every timestep, and leverage a new decoupling loss to explicitly extend the distance between them in latent space. 
By this means, different memory states are trained to focus on different aspects of spatiotemporal variations.
The overall memory decoupling method can be formulated as follows:
\begin{equation}\label{equ:stlstm_regu}
  \begin{split}
  \Delta \mathcal{C}_t^l & = W_{\text{decouple}} \ast (i_t \odot g_t) \\
  \Delta \mathcal{M}_t^l & = W_{\text{decouple}} \ast (i_t^\prime \odot g_t^\prime) \\
  \mathcal{L}_{\text{decouple}} & = \sum_{t}\sum_{l}\sum_{c}\frac{\lvert\left< \Delta \mathcal{C}_t^l, \Delta \mathcal{M}_t^l\right>_{c}\rvert}{\Vert \Delta \mathcal{C}_t^l\Vert_{c}\cdot\Vert\Delta \mathcal{M}_t^l\Vert_{c}},\\
  \end{split}
\end{equation}
where $W_{\text{decouple}}$ denotes $1\times1$ convolutions shared by all ST-LSTM units; $\left<\cdot,\cdot\right>_{c}$ and $\Vert\cdot\Vert_{c}$ are respectively dot product and $\ell_2$ norm of flattened feature maps, which are calculated for each channel $c$.
At training time, the increments of memory states, $i_t \odot g_t$ and $i_t^\prime \odot g_t^\prime$, are derived from \eqn{equ:stlstm}.
Notably, the new parameters are only used at training time and are removed from the entire model at inference. That is, there is no increase in model size compared to the previous version of ST-LSTM \cite{wang2017predrnn}.
By defining the decoupling loss with the cosine similarity, our approach encourages the increments of the two memory states to be orthogonal at any timestep. It unleashes the respective strengths of $\mathcal{C}$ and $\mathcal{M}$ for long-term and short-term dynamic modeling.
As shown by t-SNE visualization in \fig{fig:decouple}, at test time, $\Delta \mathcal{C}_t^l$ and $\Delta \mathcal{M}_t^l$ can be easily separated.

The decoupling loss is largely inspired by the theoretical evidence that using reasonable inductive bias to construct distributed representations can bring a great performance boost if it matches properties of the underlying data distribution \cite{pascanu2013number}.
There is a similar idea in ensemble learning that generally, to form a good ensemble, the base learners should be as more accurate as possible, and as more diverse as possible \cite{krogh1995neural}. 
The diversity of base learners can be enhanced in different ways, such as sub-sampling the training examples, manipulating the input attributes, and employing randomness into training procedures \cite{zhou2009ensemble}. 
%
It is worth noting that the proposed memory decoupling method has not been used by existing ensemble learning algorithms, though it is inspired by the general idea of enhancing the diversity of base learners. 
We use it to diversify the pairs of memory states, intuitively in pursuit of a more disentangled representation of long- and short-term dynamics in predictive learning. 

The final PredRNN model is trained end-to-end in a fully unsupervised manner. The overall training objective is:
\begin{equation}\label{equ:final_loss}
  \begin{split}
  \mathcal{L}_\text{final} =  \sum_{t=2}^{T+K} \big\| \widehat{\mathcal{X}}_t - \mathcal{X}_t\big\|_2^2 + \mathcal{L}_{\text{decouple}},\\
  \end{split} 
\end{equation} 
where the first term is the frame reconstruction loss that works on the network output at each timestep, the second term is the memory decoupling regularization from \eqn{equ:stlstm_regu}.


\subsection{Action-Conditioned ST-LSTM}

\begin{figure}[t]
  \centering
  \includegraphics[width=\columnwidth]{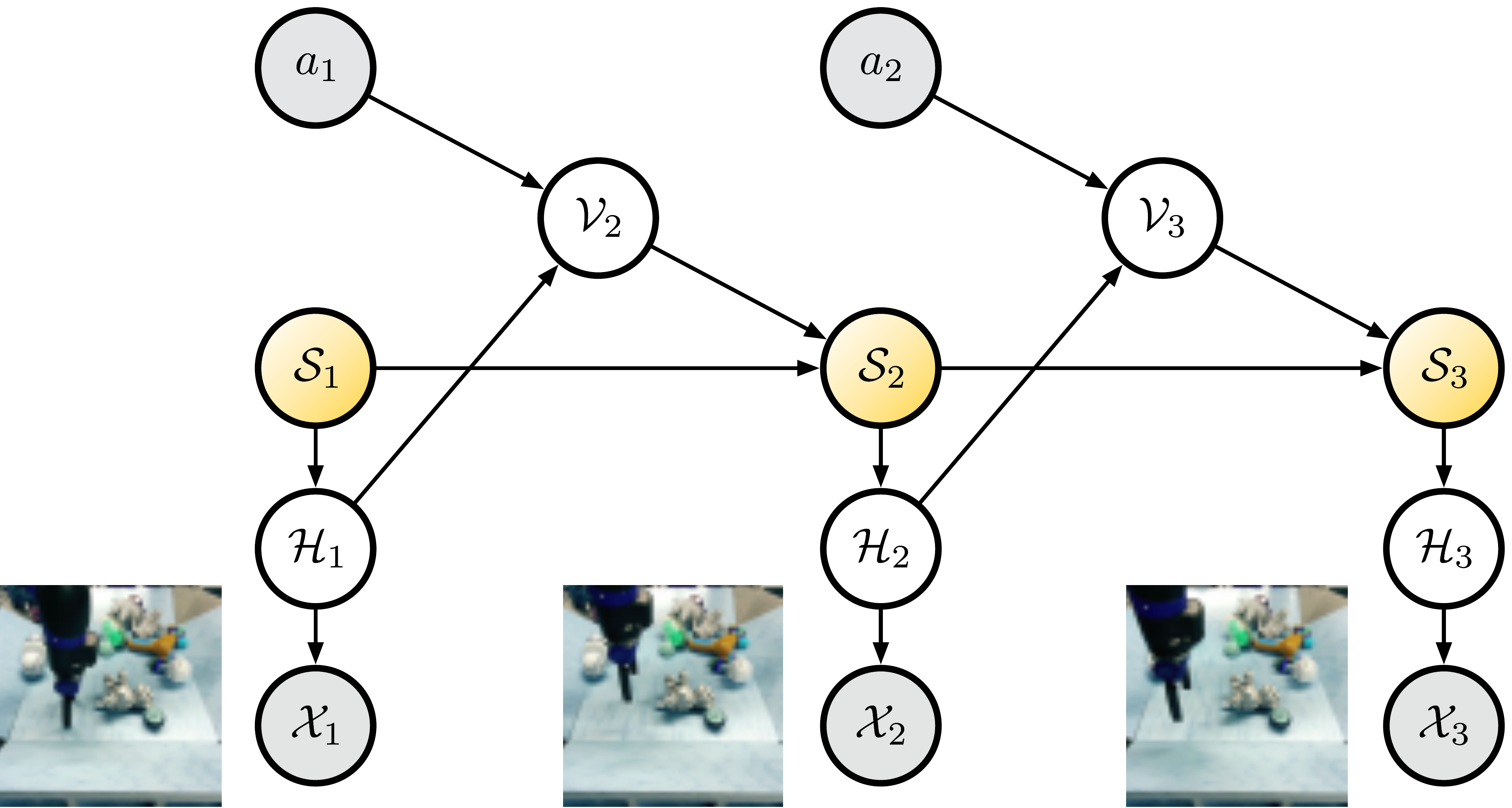}
  \caption{Graphical model of the action-conditioned PredRNN. For simplicity, we display only one ST-LSTM layer and use $\mathcal{S}_t$ as the combination of the memory states $\mathcal{C}_t$ and $\mathcal{M}_t$ in \eqn{equ:action}.
  }
  \label{fig:action}
  \vspace{-10pt}
\end{figure}

Action-conditioned video prediction models allow for flexibility in learning the policies of agent-based decision making systems \cite{Oh2015Action,Finn2016Unsupervised,chiappa2017recurrent,babaeizadeh2017stochastic,ebert2017self,hafner2018learning,hafner2020dream}.
Inspired by the work of Recurrent Environment Simulator \cite{chiappa2017recurrent}, we extend ST-LSTMs, which can be denoted by $\text{ST-LSTM}(\mathcal{X}_t, \mathcal{H}_{t-1}^l, \mathcal{C}_{t-1}^l, \mathcal{M}_t^{l-1})$, to action-conditioned video prediction with
\begin{equation}\label{equ:action}
  \begin{split}
  \text{Action fusion: \ } \mathcal{V}_{t}^l & = (W_{hv} \ast \mathcal{H}_{t-1}^l)\odot(W_{av} \ast \mathcal{A}_{t-1}) \\
  \mathcal{H}_t^{l},\mathcal{C}_t^{l},\mathcal{M}_t^{l} &=\text{ST-LSTM}(\mathcal{X}_t, \mathcal{V}_{t}^l, \mathcal{C}_{t-1}^l, \mathcal{M}_t^{l-1}),\\
  \end{split} 
\end{equation}
where $\mathcal{V}_{t}^l$ encodes previous hidden state $\mathcal{H}_{t-1}^l$ and the action taken at the previous timestep $\mathcal{A}_{t-1}$. 
In our experiments, $\mathcal{A}_{t}$ has continuous values and has been transformed to the same resolution as the hidden state. 
By incorporating $\mathcal{V}_{t}^l$ in each ST-LSTM unit, at both encoding and forecasting timesteps, PredRNN learns to simulate the consequences of future action sequences over long time periods.
Different from the original Recurrent Environment Simulator, we validate the effectiveness of action-conditioned ST-LSTM on a dataset collected with a real robot \cite{ebert2017self}.

\subsection{Training with Reverse Scheduled Sampling}
\label{method_curri}

\begin{figure*}[t]
  \centering
  \includegraphics[width=0.95\textwidth]{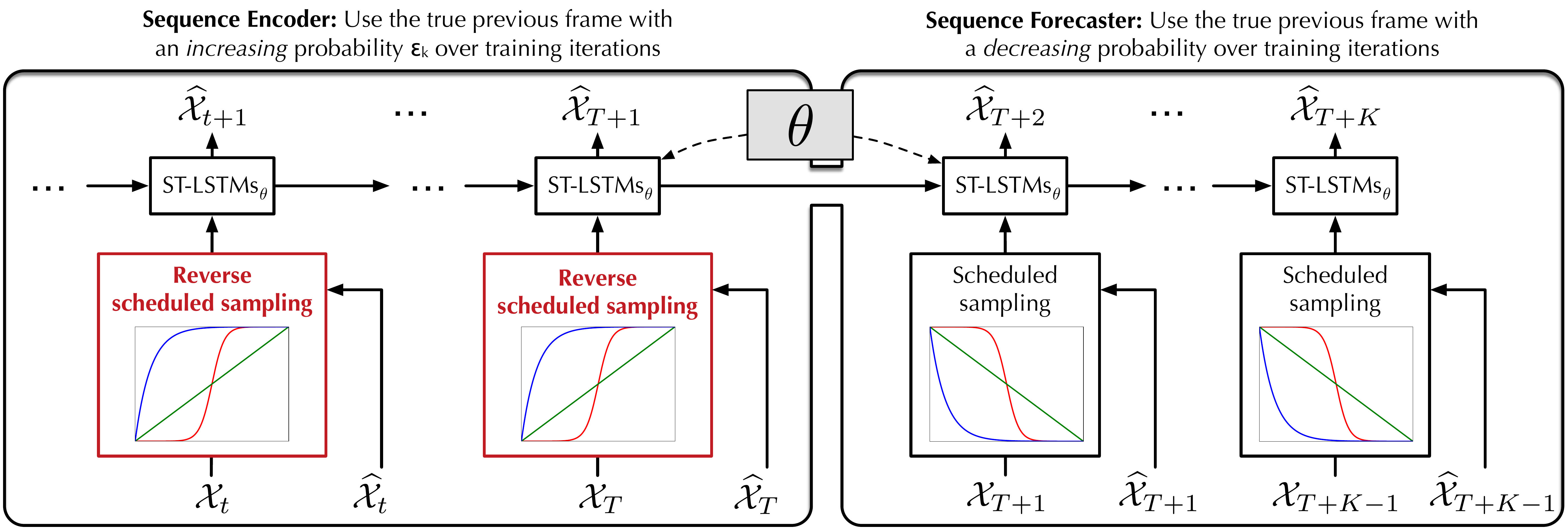}
  \vskip -0.05in
  \caption{In the training phase, we apply the \textit{reverse scheduled sampling} at the encoding timesteps to force the model to learn long-term dynamics from context frames. This training scheme can be easily combined with the original scheduled sampling strategy, which is used to close the training-inference gap at forecasting timesteps. Both the encoder and the forecaster are parametrized by $\theta$.
  }
  \label{fig:rss}
  \vspace{-10pt}
\end{figure*}

As shown in \fig{fig:rss}, we propose a new curriculum learning strategy for PredRNN, which consists of two components:
\begin{itemize}[leftmargin=*]
    \item \textbf{Reverse scheduled sampling (RSS)}: It is used at the encoding timesteps and forces the model to learn more about long-term dynamics by randomly hiding real observations with decreasing probabilities as the training phase progresses.
    \item \textbf{Scheduled sampling (SS)} \cite{bengio2015scheduled}: It is used at the forecasting timesteps to alleviate the inconsistency of data flow between the training and inference phases.
\end{itemize}

\myparagraph{Difficulty in learning long-term dynamics}
We follow the common practice \cite{Finn2016Unsupervised,denton2018stochastic} that uses a unified predictive model $f_\theta(\cdot)$ for sequence encoding and forecasting.
We use $\mathcal{X}_\text{in}=\{\mathcal{X}_1,\ldots,\mathcal{X}_T\}$ to denote the context frames and $\mathcal{X}_\text{out}=\{\mathcal{X}_{T+1},\ldots,\mathcal{X}_{T+K}\}$ for ground-truth output frames. We use $\widehat{\mathcal{X}}_{t+1}$, $t\in\{1,\ldots,T+K-1\}$, to denote the prediction at each timestep. During inference, as well as in the typical training scheme, we have
\begin{equation}
\label{eq:inference}
   \widehat{\mathcal{X}}_{t+1} = 
   \begin{cases}
   f_\theta(\mathcal{X}_t,\mathcal{H}_{t-1}, \mathcal{S}_{t-1}) & \text{if} \ t\le T, \\
   f_\theta(\widehat{\mathcal{X}}_t,\mathcal{H}_{t-1},\mathcal{S}_{t-1}) & \text{otherwise}, \\
   \end{cases}
\end{equation}
where $\mathcal{S}_{t-1}=\{\mathcal{C}_{t-1},\mathcal{M}_{t-1}\}$ represents the temporal dynamics as a combination of previous memory states. 
The main difference between the encoding part ($t \le T$) and the forecasting part ($t>T$) is whether to use the true frame $\mathcal{X}_t$ or the previous prediction $\widehat{\mathcal{X}}_t$.
In the encoder, the model is mainly learned to make a one-step prediction, because $\mathcal{X}_t$ tends to be a more informative signal than $\mathcal{H}_{t-1}$  and $\mathcal{S}_{t-1}$, keeping PredRNN from digging deeper into the non-Markovian properties of historical observations. 
But for the forecaster, since $\widehat{\mathcal{X}}_t$ does not contain new observations, the model has to learn long-term dynamics from the latter. 
Such a training inconsistency between the encoder and the forecaster may lead to an ineffective optimization of $\theta$ and hamper the model from learning long-term dynamics from $\mathcal{X}_\text{in}$.

\begin{figure}[t]
\centering
\vspace{-5pt}
\subfigure[Strategy 1]{
\includegraphics[width=0.472\columnwidth]{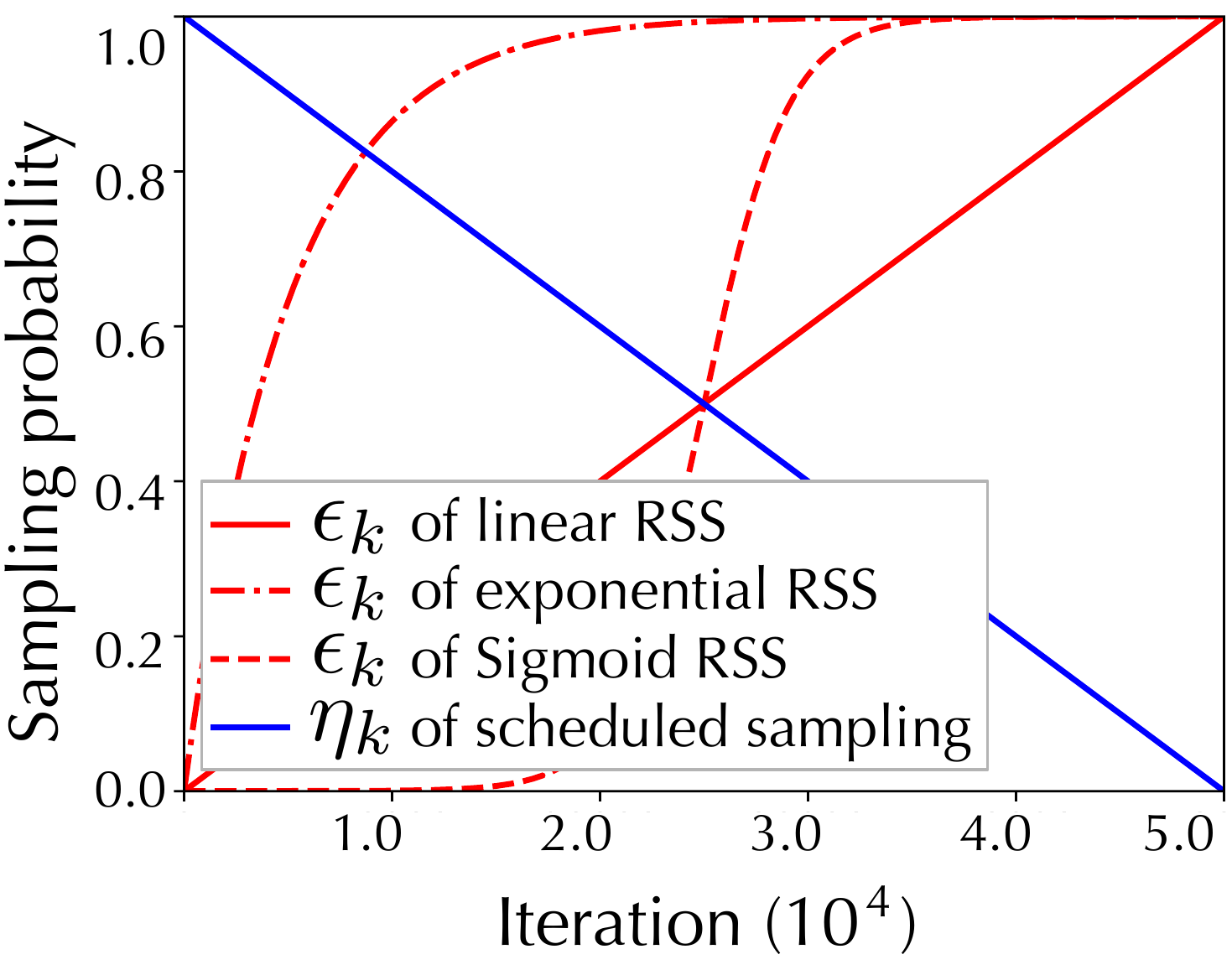}
\label{fig:Strategy_1}
}
\subfigure[Strategy 2]{
\includegraphics[width=0.472\columnwidth]{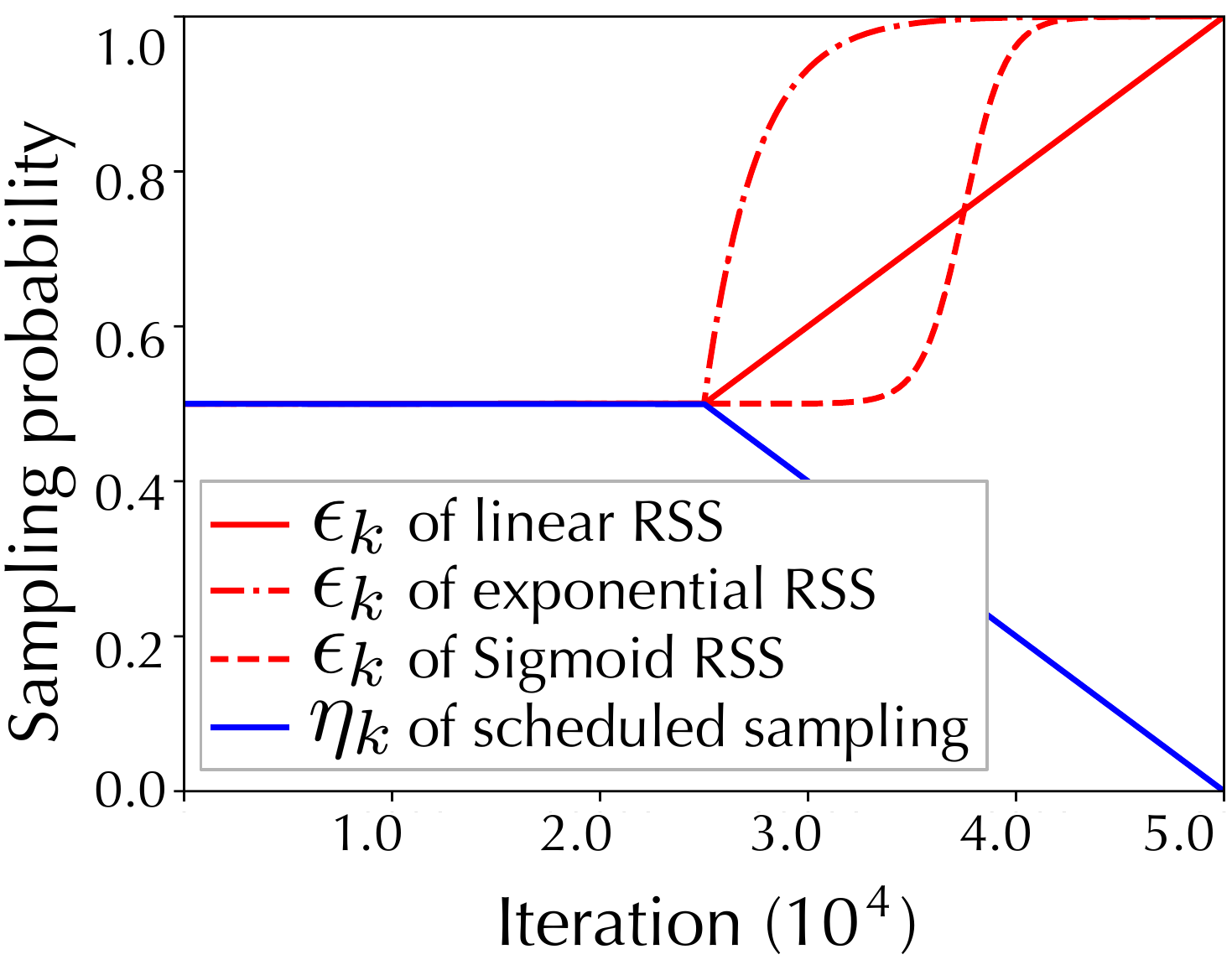}
\label{fig:Strategy_2}
}
\vskip -0.15in
\caption{Two feasible strategies to combine the reverse scheduled sampling (in red) and the original scheduled sampling (in blue).
}
\label{fig:rss_schemes}
\vspace{-10pt}
\end{figure}

\myparagraph{Reverse scheduled sampling}
To force the recurrent model to learn long-term dynamics from historical observations, we propose the reverse scheduled sampling, a curriculum learning strategy applied to the input frames at encoding timesteps. 
At forecasting timesteps, we follow the previous literature \cite{Finn2016Unsupervised,denton2018stochastic} to apply the original schedule sampling \cite{bengio2015scheduled} to the inputs.
Finally, we have
\begin{equation}
\label{eq:rss}
   \widehat{\mathcal{X}}_{t+1} =
   \begin{cases}
   f_\theta(\widehat{\mathcal{X}}_t \xmapsto[]{\text{RSS}} \mathcal{X}_t, \mathcal{H}_{t-1}, \mathcal{S}_{t-1}) & \text{if} \ t\le T, \\
   f_\theta(\mathcal{X}_t \xmapsto[]{\text{SS}} \widehat{\mathcal{X}}_t,, \mathcal{H}_{t-1} \mathcal{S}_{t-1}) & \text{otherwise}, \\
   \end{cases}
\end{equation}
where $\xmapsto[]{\text{RSS}}$ indicates a gradual change throughout the training phase from taking the previous prediction  $\widehat{\mathcal{X}}_t$ as the input to taking the true frame $\mathcal{X}_t$.
At encoding timesteps, the changing schedule is in the reverse order of the scheduled sampling strategy denoted by $\xmapsto[]{\text{SS}}$ at forecasting timesteps.
Concretely, for RSS, there is a probability ($\epsilon_k \in [0,1]$) of sampling the true frame $\mathcal{X}_t\in\mathcal{X}_\text{in}$ or a corresponding probability ($1-\epsilon_k$) of sampling $\widehat{\mathcal{X}}_t$, such that over the entire encoder, a sequence of sampling outcomes can be seen as a Bernoulli process with $T$ independent trials. 
$\epsilon_k$ is an \textit{increasing} function of the number of training iterations $k$, starting from $\epsilon_s$ and increasing to $\epsilon_e$, which has the following forms: 
\begin{itemize}[leftmargin=*]
\item Linear increase: $\epsilon_k = \mathrm{min}\{\epsilon_s+\alpha_l\times k, \epsilon_e\}$;
\item Exponential increase: $\epsilon_k = \epsilon_e-(\epsilon_e-\epsilon_s)\times \mathrm{exp}(-\frac{k}{\alpha_e})$;
\item Sigmoid increase: $\epsilon_k = \epsilon_s+(\epsilon_e-\epsilon_s)\times\frac{1}{1+\mathrm{exp}(\frac{\beta_s-k}{\alpha_s})}$,
\end{itemize}
where $\alpha_l, \alpha_e, \alpha_s >0$ denote the increasing factors and $\beta_s > 0$ denotes the starting point of the sigmoid function. These hyper-parameters jointly decide the increasing curve of $\epsilon_k$.
Examples of such schemes are shown in \fig{fig:rss_schemes} in the red curves. 
The encoder is trained with a progressively simplified curriculum. It gradually changes from generating multi-step future frames, which is challenging due to the absence of some historical observations, to making one-step predictions, just as the encoder does at test time. 
Such a training scheme encourages the model to extract long-term, non-Markovian dynamics from the input sequence.


\myparagraph{Entire training scheme}
\fig{fig:rss_schemes} shows two feasible strategies to jointly use RSS and the original scheduled sampling method, where $\eta_k$ denotes the probability of the original scheduled sampling at forecasting timesteps.
For simplicity, we make $\eta_k$ decay linearly, although other scheduled sampling schemes could be employed (such as an exponential decay).
The biggest difference between the two strategies lies in whether the variation ranges of the sampling probabilities of encoder and forecaster are close at the early stage of training.
Empirically, the second strategy performs slightly better (as shown in \tab{tab:mnist_rss} in Section \ref{sec:mnist}).
This suggests that in the early training stages, it is helpful to sample true frames with similar probabilities in the encoding and forecasting parts of PredRNN.

Like scheduled sampling, the RSS training strategy can be widely used in most sequence-to-sequence models beyond PredRNN to enhance the long-term modeling capability.

\section{Experiments}

In this section, we evaluate our approach on five spatiotemporal prediction datasets. We strongly encourage readers to view \url{https://github.com/thuml/predrnn-pytorch} for the source code.

\subsection{Experimental Setups}

\myparagraph{Datasets} 
We valuate PredRNN on the following datasets for both synthetic and real-world scenarios:
\begin{itemize}[leftmargin=*]
    \item \textbf{Moving MNIST}: 
    This dataset contains handwritten digits that are sampled from the static MNIST, placed at random locations, and initialized with a random speed. They bounce off the edges of the image at a certain angle. For a fair comparison, we follow two training setup from ConvLSTM \cite{shi2015convolutional} and CrevNet \cite{yu2020efficient}. In the first setup, we use a fixed training set of $10{,}000$ samples; while in the second one, we generate training data on the fly, which provides a larger number of training samples. We use a test set of $5{,}000$ sequences where the digits are sampled from a different subset of static MNIST. 
    \item \textbf{KTH} \cite{Sch2004Recognizing}: This dataset contains $6$ types of human actions, \textit{i.e.}, walking, jogging, running, boxing, hand-waving, and hand-clapping, performed by $25$ persons in $4$ different scenes. The videos last $4$ seconds on average and have a frame rate of $25$ FPS. We resize the frames to a resolution of $128\times128$. We adopt the protocol from \cite{Villegas2017Decomposing}, \textit{i.e.}, persons $1$-$16$ for training; persons $17$-$25$ for testing, and obtain a training set of $108{,}717$ sequences and a test set of $4{,}086$ sequences.
    \item \textbf{Radar echo dataset}: This dataset contains $10{,}000$ consecutive radar maps recorded every $6$ minutes at Guangzhou, China. We transform the radar maps to pixel values and represent them as $128\times128$ gray-scale images. We divide the dataset into $7{,}800$ training sequences and $1{,}800$ test sequences.
    \item \textbf{Traffic4Cast} \cite{Traffic4cast-web-2019}: This dataset records the GPS trajectories of consecutive traffic flows in Berlin, Moscow, and Istanbul in the form of video frames in 2019. The size of each frame is $495 \times 436 \times 3$. The value of each pixel corresponds to the traffic information in an area of $100$m$\times100$m in $5$ minutes, including mean speed, volume, and major traffic direction. 
    \item \textbf{BAIR} \cite{ebert2017self}: This dataset contains the action-conditioned videos collected by a Sawyer robotic arm pushing a variety of objects. 
    At each timestep, we have a $64\times64$ RGB image and the action vector of the commanded gripper pose. We have a training set of $822{,}016$ sequences and a test set of $4{,}864$ sequences.
\end{itemize}

\begin{figure*}[t]
\centering
    \subfigure[Frame-wise MSE ($\downarrow$)]{
    \includegraphics[width=0.65\columnwidth]{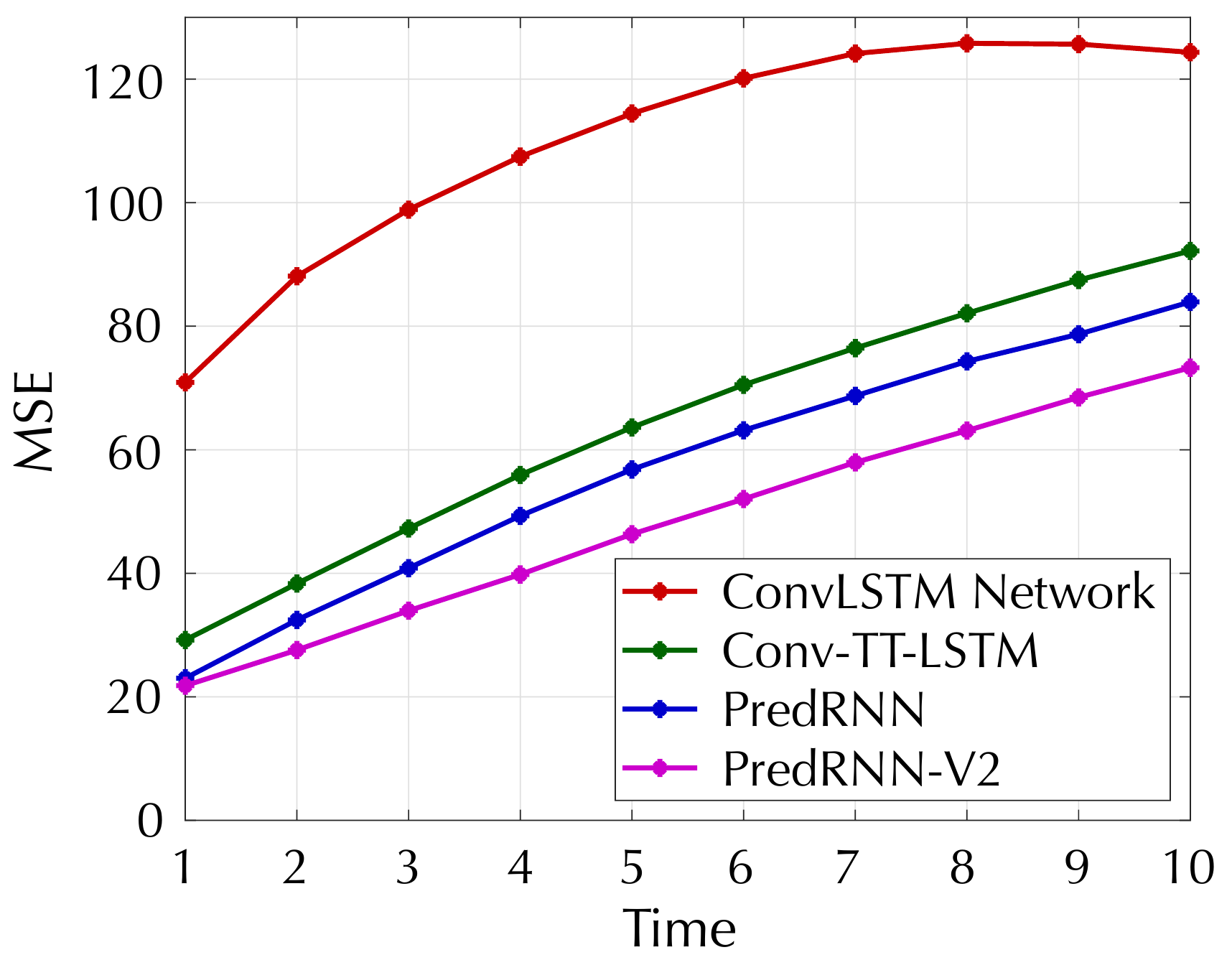}
    }
    \subfigure[Frame-wise SSIM ($\uparrow$)]{
    \includegraphics[width=0.65\columnwidth]{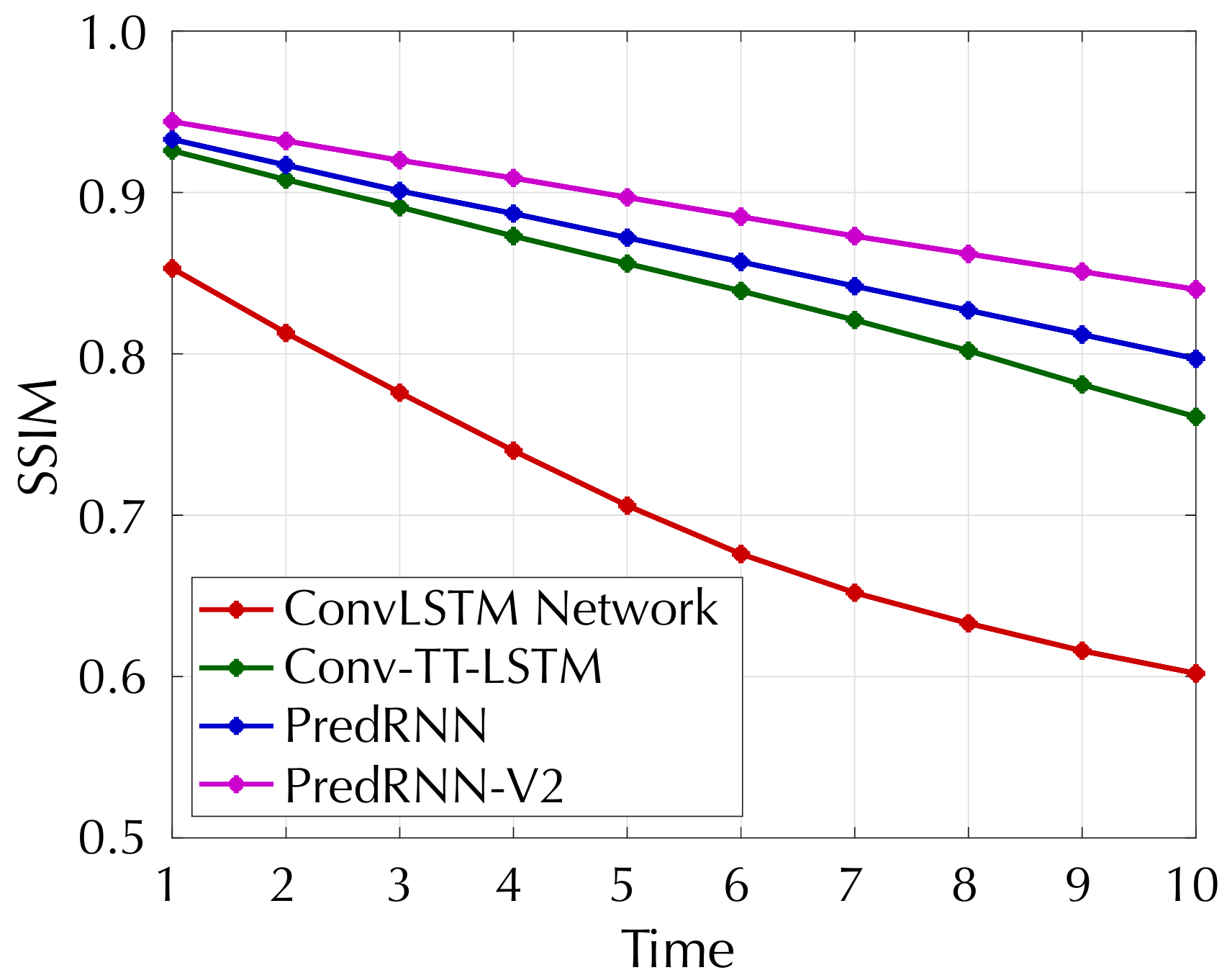}
    }
    \subfigure[Frame-wise LPIPS ($\downarrow$)]{
    \includegraphics[width=0.65\columnwidth]{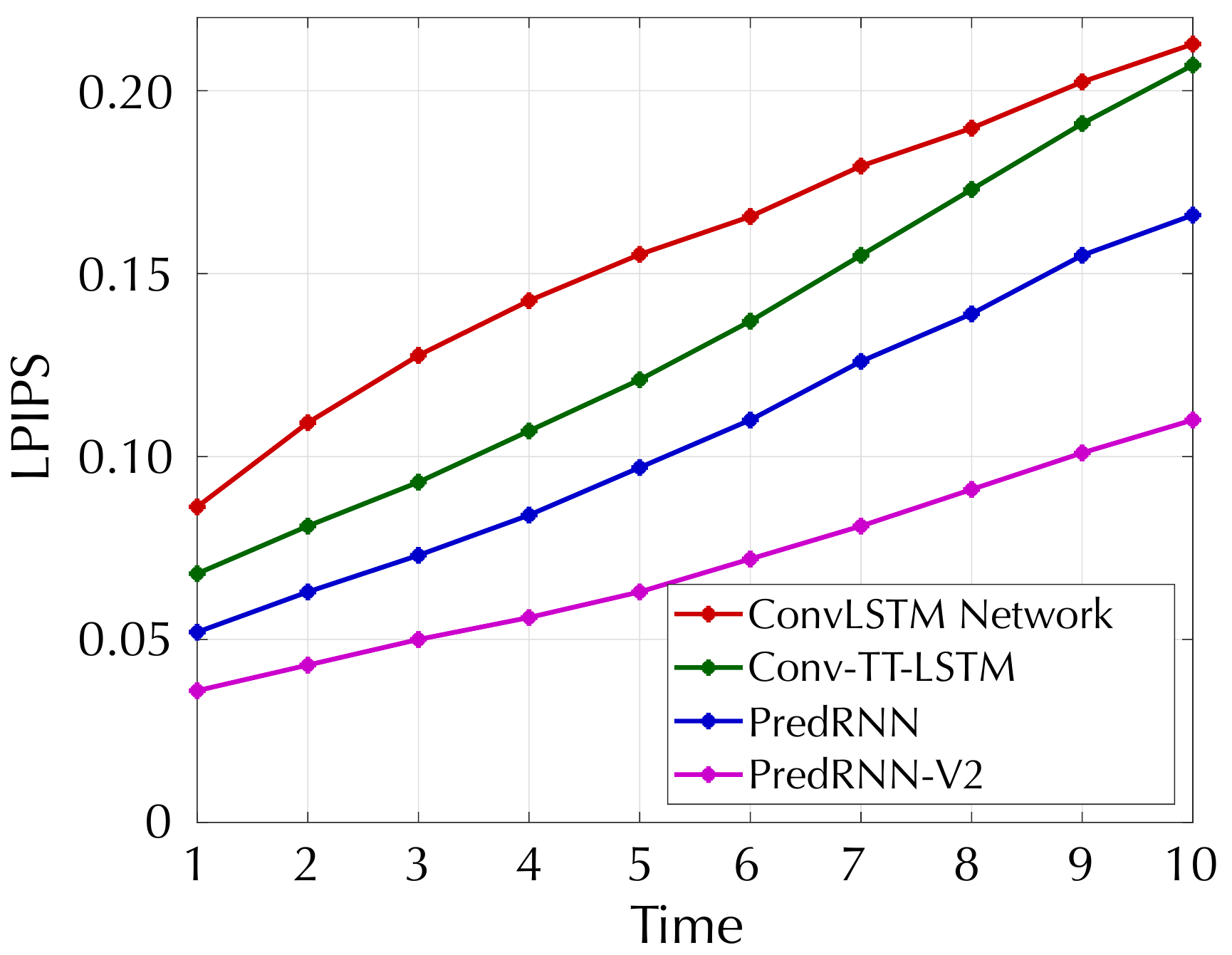}
    }
\vskip -0.1in
\caption{Frame-wise results on the Moving MNIST test set produced by models trained on the fixed training set.}
\label{fig:mnist_frame}
\vspace{-15pt}
\end{figure*}

\begin{figure*}[t]
  \centering
  \includegraphics[width=\textwidth]{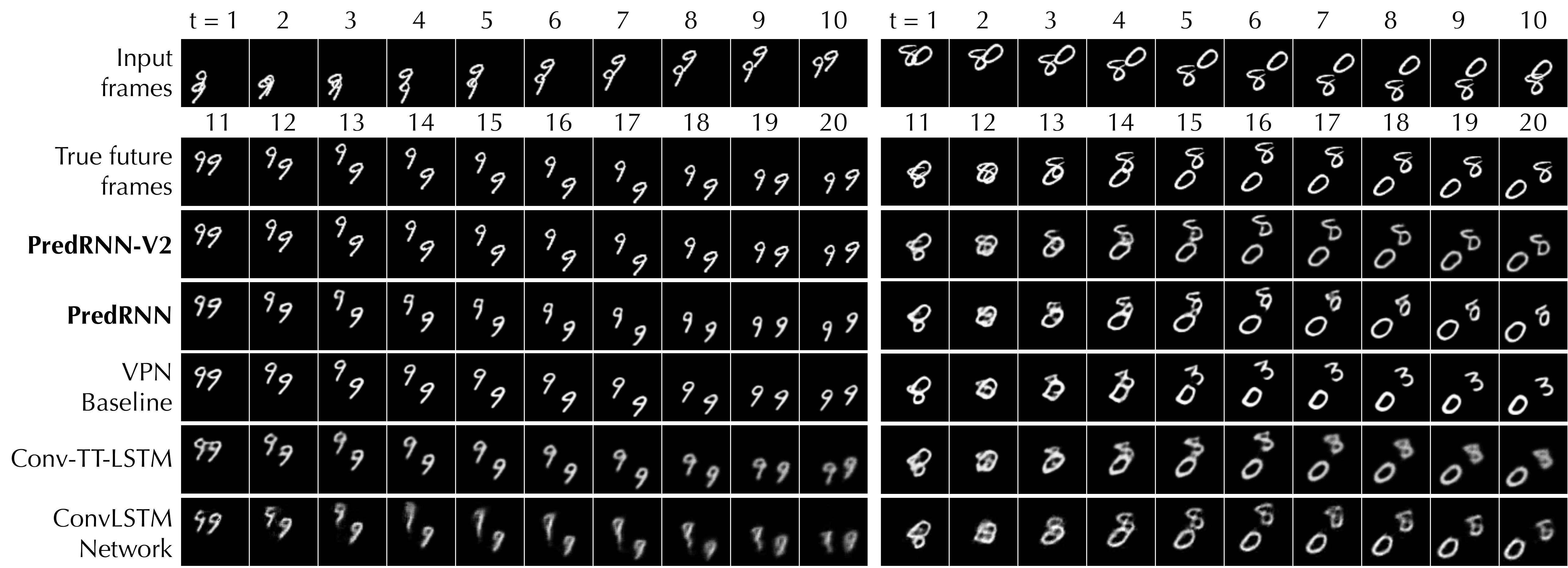}
  \vskip -0.05in
  \caption{Prediction examples on the Moving MNIST test set.}
  \label{fig:mnist_result}
  \vspace{-15pt}
\end{figure*}

\begin{table}[t]
\vskip 0.05in
  \caption{Performance on the Moving MNIST test set, averaged over $10$ prediction timesteps. All models are trained with a fixed training set of $10{,}000$ sequences. For a fair comparison, we also report the model size and computation efficiency. ConvLSTM* denotes a $4$-layer network with a $256$-channel hidden state each.
  }
  \vskip -0.05in
  \label{tab:mnist_mse}
  \centering
  \setlength{\tabcolsep}{3.8pt}
  \begin{tabular}{lcccccc}
    \toprule
    \multirow{2}{*}{Model} & MSE & SSIM  & LPIPS & \#Params. & FLOPS & Mem.  \\
    & ($\downarrow$) & ($\uparrow$) &  ($\downarrow$) & (MB) & (G) & (GB) \\
    \midrule
    ConvLSTM \cite{shi2015convolutional} & 103.3& 0.707  & 0.156 & 16.60 & 80.7 & 2.58 \\
    ConvLSTM* \cite{shi2015convolutional} & 62.8 & 0.846  & 0.126 & 65.96 & 320.8 & 5.62 \\
    CDNA \cite{Finn2016Unsupervised} & 97.4& 0.721  & - & - & - & - \\
    VPN Baseline \cite{Kalchbrenner2016Video} & 64.1& 0.870  & - & - & - & - \\
    MIM \cite{wang2019memory} & 52.0 & 0.874 & 0.079 & 37.37 & 181.7 & 4.22 \\
    Conv-TT-LSTM \cite{su2020convolutional} & 64.3 & 0.846  & 0.133 & 23.91 & 116.3 & 6.17 \\
    \midrule
    PredRNN-$\mathcal{M}$-Only & 74.0 & 0.851 & 0.109 & 18.30 & 89.02 & 2.37 \\
    PredRNN (\textit{Conf.}) & 56.8 & 0.867 & 0.107 & 23.85 & 115.9 & 3.52 \\
    PredRNN-V2 & \textbf{48.4} & \textbf{0.891} & \textbf{0.071} & 23.86 & 116.6 & 3.97 \\
    \bottomrule
  \end{tabular}
  \vspace{-10pt}
\end{table}

\myparagraph{Compared models} 
We use the ConvLSTM network \cite{shi2015convolutional} as the primary baseline model, and compare PredRNN with the current states of the art for each dataset, including: 
\begin{itemize}[leftmargin=*]
    \item CrevNet \cite{yu2020efficient} for Moving MNIST and Traffic4Cast.
    \item Conv-TT-LSTM \cite{su2020convolutional} for the KTH action dataset.
    \item TrajGRU \cite{shi2017deep} for precipitation forecasting.
    \item SV2P \cite{babaeizadeh2017stochastic} for action-conditioned video prediction.
\end{itemize}
Notably, some existing approaches have extended PredRNN in different aspects \cite{wang2018predrnn++,wang2019eidetic} or use ST-LSTM as the network backbone \cite{yu2020efficient}.
The differences between them are summarized in \tab{tab:model_compare}. Particularly for these approaches, the proposed contributions of memory decoupling and reverse scheduled sampling are shown to be easily combined with them and enable them to approach state-of-the-art performance.
To facilitate the discussion of ablation study, we refer to different versions of PredRNN as:
\begin{itemize}[leftmargin=*]
    \item \textbf{PredRNN-$\mathcal{M}$-Only}: This model improves the ConvLSTM network with the spatiotemporal memory flow ($\mathcal{M}$). The architecture is shown in \fig{fig:rnn_compare} (left).
    \item \textbf{PredRNN}: This model uses ST-LSTMs as the building blocks. It was proposed in our conference paper \cite{wang2017predrnn}.
    \item \textbf{PredRNN-V2}: This is the final proposed model that improves the training process of the original PredRNN with memory decoupling and reverse scheduled sampling.
\end{itemize}

\myparagraph{Implementation details} 
We use the ADAM optimizer \cite{Kingma2014Adam} to train the models with a mini-batch of $2$ to $16$ sequences according to different datasets. 
Unless otherwise specified, we set the learning rate to $10^{-4}$ and stop the training process after $80{,}000$ iterations.
We typically use four ST-LSTM layers in PredRNN to strike a balance between prediction quality and training efficiency. We set the number of channels of each hidden state to $128$ and the size of convolutional kernels inside the ST-LSTM unit to $5\times 5$.
%

\subsection{Moving MNIST Dataset}
\label{sec:mnist}

In this dataset, future trajectories of moving digits are predictable based on enough historical observations, where part of the challenge is to infer the underlying dynamics with random initial velocities.
Moreover, the frequent occlusion of multiple digits leads to complex and short-term variations of local spatial information, which brings more difficulties to spatiotemporal prediction.
We train the models to predict the next ten frames given the first ten.

\myparagraph{Quantitative results with a fixed training set}
\tab{tab:mnist_mse} shows the results of all compared models averaged per frame. We adopt evaluation metrics that were widely used by previous methods: the Mean Squared Error (MSE), the Structural Similarity Index Measure (SSIM) \cite{Wang2004Image}, and the Learned Perceptual Image Patch Similarity (LPIPS) \cite{zhang2018unreasonable}.
The difference between these metrics is that MSE estimates the absolute pixel-wise errors, SSIM measures the similarity of structural information within the spatial neighborhoods, while LPIPS is based on deep features and aligns better to human perceptions.
\fig{fig:mnist_frame} provides the corresponding frame-wise comparisons. The final PredRNN model significantly outperforms all previous approaches, and all the proposed techniques have their contributions. 
First, with the proposed spatiotemporal memory flow, the PredRNN-$\mathcal{M}$-only model reduces the per-frame MSE of the ConvLSTM baseline from $103.3$ down to $74.0$. 
Second, by using the ST-LSTM in place of the ConvLSTM unit, our model further reduces the MSE down to $56.8$. 
Finally, the employment of the memory decoupling and the reverse scheduled sampling techniques brings another $14.8\%$ improvement in MSE (from $56.8$ to $48.4$). 
In \tab{tab:mnist_mse}, we also show the model size, computational efficiency, and memory usage for the sake of fair comparisons. 
Note that PredRNN performs better and is more efficient than a large version of the ConvLSTM network (denoted by ConvLSTM*) with a doubled number of channels in the hidden states.

\begin{table}[t]
\vskip 0.05in
  \caption{We follow the setup of CrevNet to train the predictive models on dynamically generated Moving MNIST. The results of CrevNet are taken directly from the paper \cite{yu2020efficient}. Models marked by * have three $64$-channel ST-LSTM layers with fewer parameters.
  }
  \vskip -0.05in
  \setlength{\tabcolsep}{4pt}
  \label{tab:mnist_fly}
  \centering
  \begin{tabular}{lcccccc}
    \toprule
    \multirow{2}{*}{Model} & \multicolumn{2}{c}{MSE ($\downarrow$)} & \multicolumn{2}{c}{SSIM ($\uparrow$)} & \#Params. \\
    & 2-digit & 3-digit & 2-digit & 3-digit & (MB) \\
    \midrule
    CrevNet w/ ConvLSTM & 38.5 & 57.2 & 0.928 & 0.886 & 2.77 \\
    CrevNet w/ ST-LSTM & 22.3 & 40.6 & 0.949 & 0.916 & 5.00 \\
    PredRNN* & 21.5 & 39.6 & 0.931 & 0.884 & 4.40\\
    PredRNN-V2* & 19.9 & 34.8 & 0.939 & 0.900 & 4.41\\
    PredRNN & 15.7 & 26.7 & 0.951 & 0.918 & 23.85 \\
    PredRNN-V2 & \textbf{13.4} & \textbf{24.7} & \textbf{0.958} & \textbf{0.928} & 23.86\\
    \bottomrule
  \end{tabular}
  \vspace{-10pt}
\end{table}

\myparagraph{Quantitative results with a dynamic training set}
To further compare the performance of PredRNN with the state of the art on the Moving MNIST dataset, we follow the experimental setups of recent approaches \cite{yu2020efficient,Kalchbrenner2016Video} where models are trained on dynamically generated data with two or three flying digits. 
The training process is stopped after $100{,}000$ iterations.
As shown in \tab{tab:mnist_fly}, PredRNN and PredRNN-V2 yield better results than those in \tab{tab:mnist_mse}, as they can have a better understanding of the spatiotemporal dynamics of the dataset with the growth of training steps. 
Notably, CrevNet also uses ST-LSTM as the recurrent unit. According to the results from its original text \cite{yu2020efficient}, it is superior to another ConvLSTM-based variant. More importantly, it is shown to be further improved by memory decoupling and reverse scheduled sampling.

\begin{figure*}[t]
  \centering
  \subfigure[$\big\|\nabla_{\mathcal{H}_t^1}\mathcal{L}_{T+K}\big\|, \ t\geq 1, T=10, K=10$]{
    \includegraphics[ height=4.22cm]{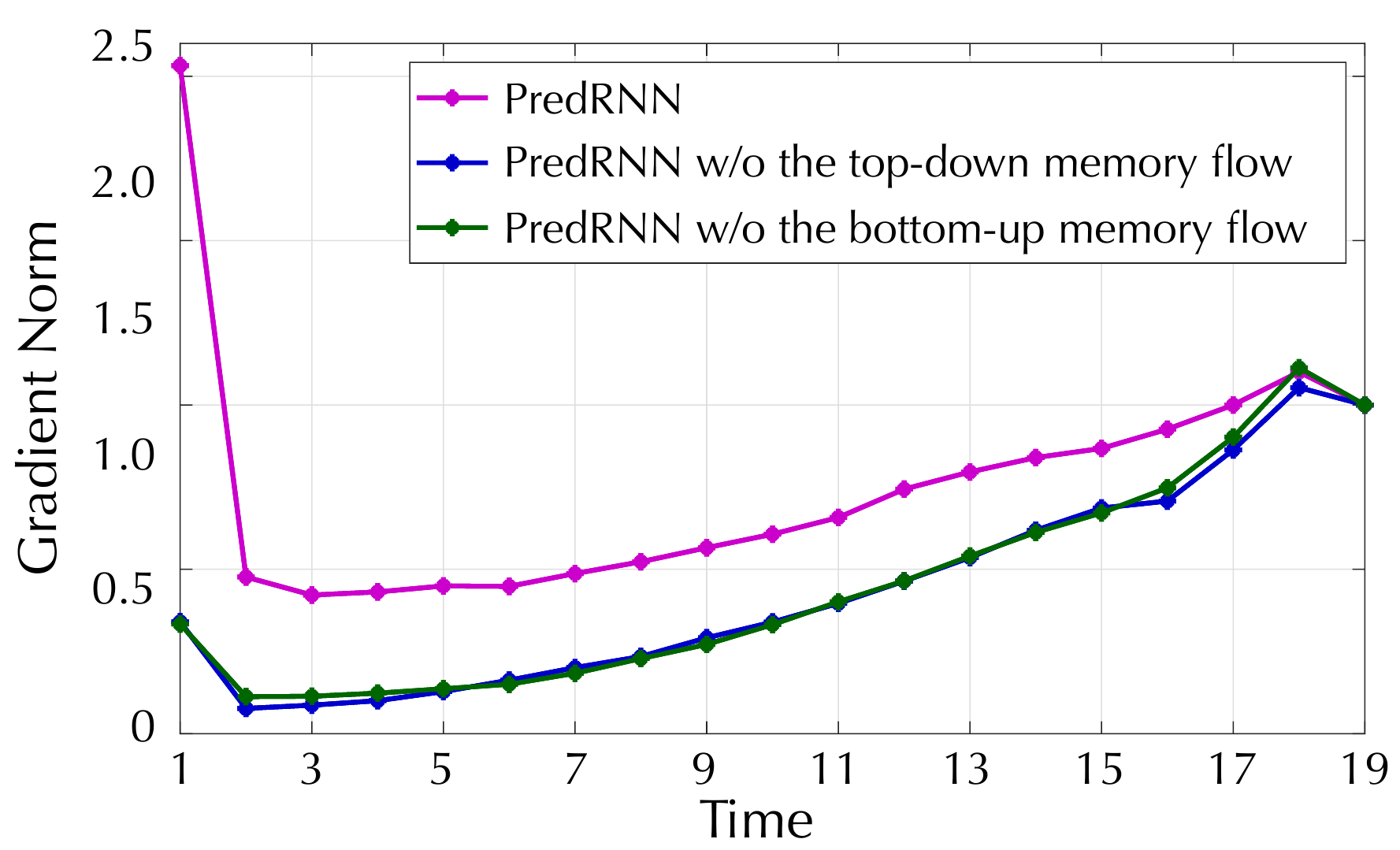}
    \label{fig:grad_a}
  }
  \hfil
  \subfigure[$\frac{1}{T-1}\sum_{\tau=2}^T\big\| \nabla_{\mathcal{H}_\tau^1}\mathcal{L}_t \big\|, \ t\geq T+1$]{
    \includegraphics[ height=4.21cm]{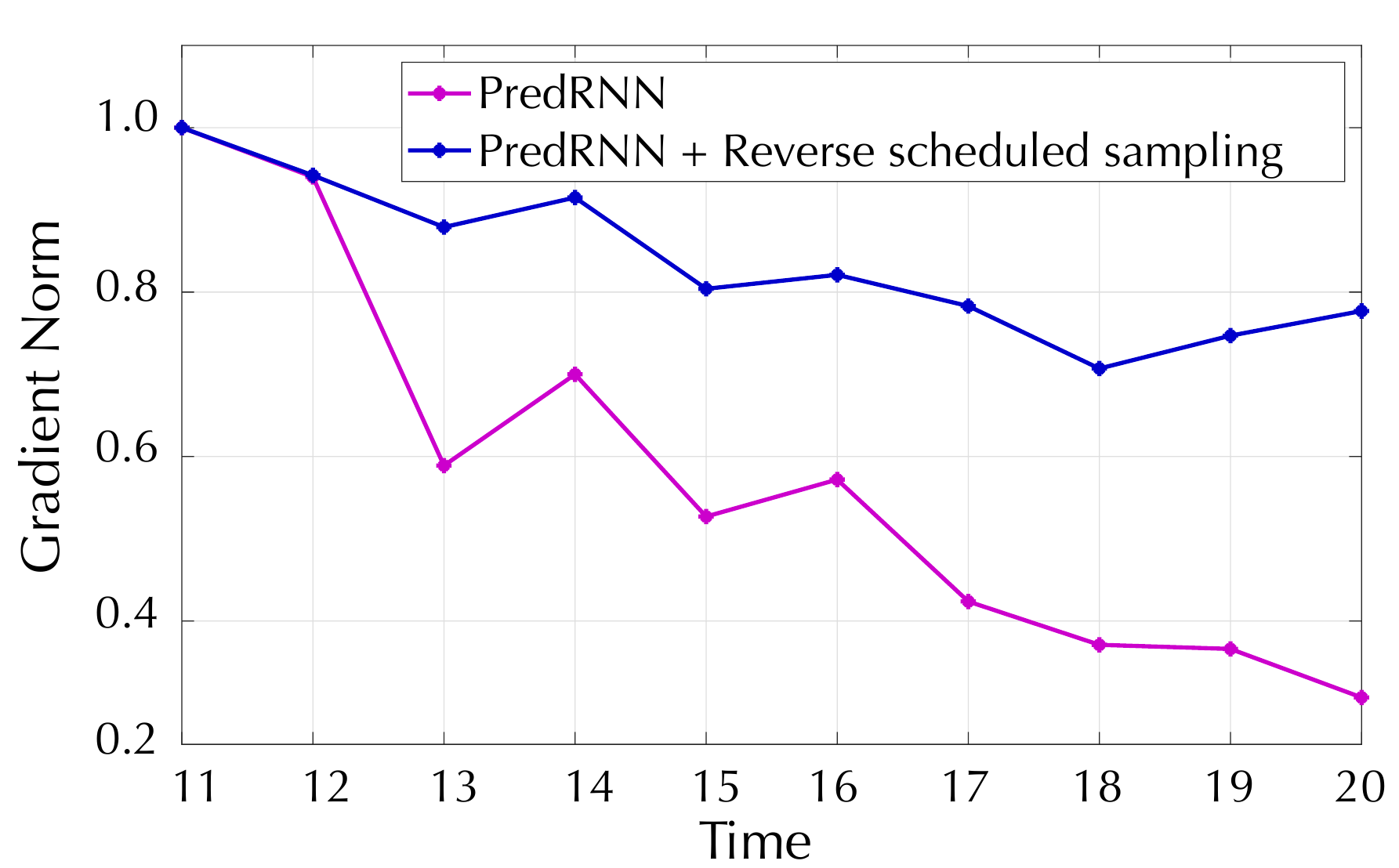}
    \label{fig:grad_b}
  }
  \vspace{-10pt}
\caption{
\revise{Gradient analyses on the Moving MNIST test set. We use well-trained models and average the results over $100$ sequences randomly sampled from the test set. \textbf{(a)} It presents the gradients of previous states concerning the loss at the last forecasting timestep, showing that the proposed memory flow benefits long-term modeling by alleviating the vanishing gradients. \textbf{(b)} It presents the accumulated gradients of the encoding states concerning the losses over the forecasting timesteps, showing the importance of RSS in encoding long-term dynamics of the inputs. We here use the RSS training strategy with an exponentially increasing $\epsilon_k$ (from \fig{fig:Strategy_2}). Both \textit{PredRNN} and \textit{PredRNN+RSS} take as input the historical real observations at encoding timesteps during testing.}
}
\label{fig:grad}
\vspace{-10pt}
\end{figure*}

\begin{figure}[t]
  \centering
  \includegraphics[width=0.85\columnwidth]{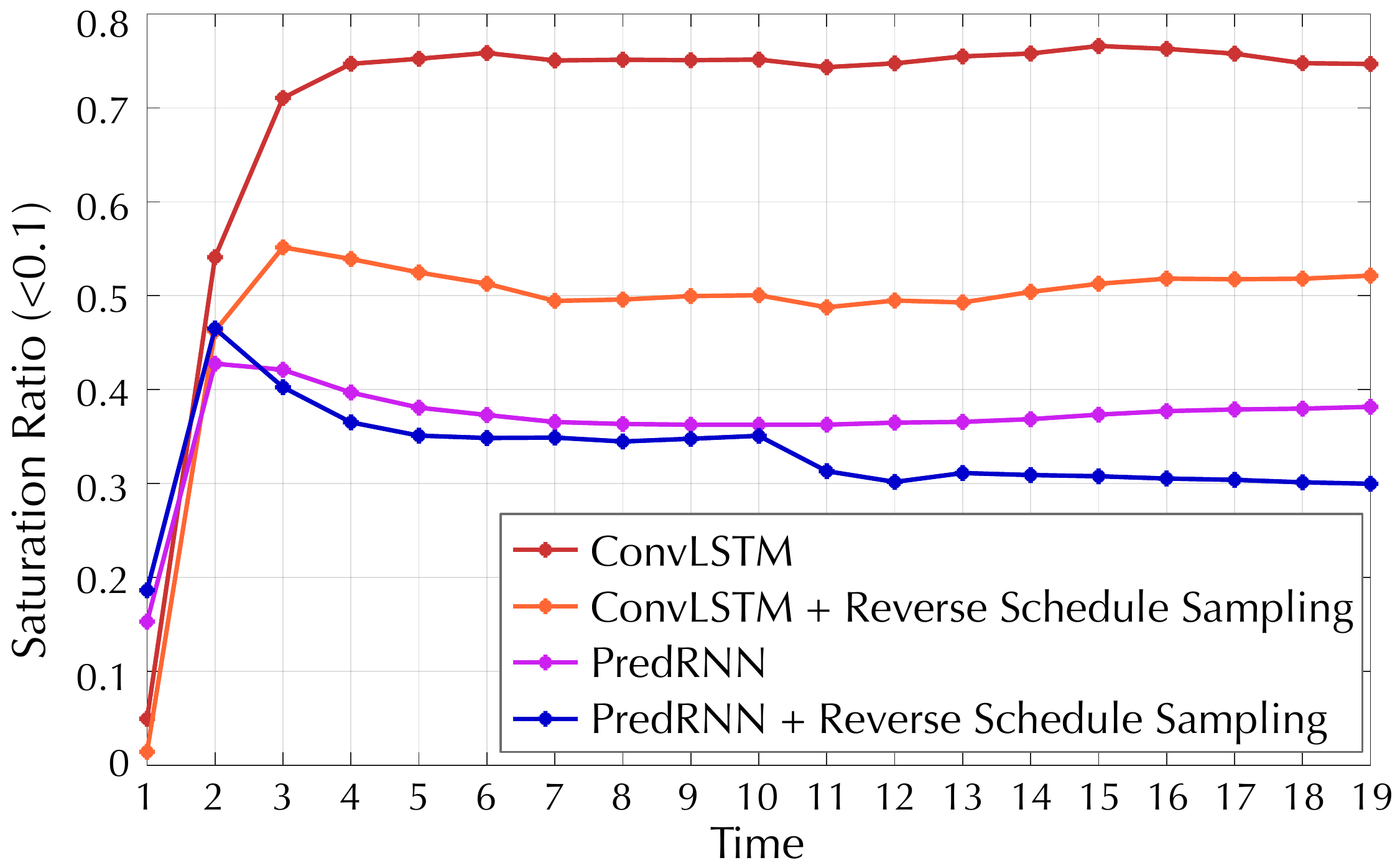}
  \vskip -0.1in
  \caption{
  \revise{An analysis of the ratios of the saturated forget gates ($f_t<0.1$) in all transitions of temporal memory states $\mathcal{C}_{t}^{l}$ ($l\in\{1,2,3,4\}$). We obtain the results by evaluating $100$ samples from the Moving MNIST test set. The results demonstrate the effect of the RSS and the zigzag memory flow in long-term modeling.}
  }
  \label{fig:c_m_gate}
  \vspace{-10pt}
\end{figure}

\myparagraph{Qualitative comparison} 
\fig{fig:mnist_result} shows two examples randomly sampled from the test set, where most of the frames produced by the compared models are severely blurred, especially for long-term predictions. 
In contrast, PredRNN produces clearer images, which means it can be more certain of future variations due to its stronger long-term modeling capabilities.
When we look closely, we can see that the most challenging issues on this dataset are to make accurate predictions of the future trajectories of moving digits and to maintain the correct shape of each digit after occlusions (in the second example, the compared VPN model incorrectly predicts the digit $8$ as $3$ after the occlusion of $8$ and $0$).
In both cases, the original PredRNN and the newly proposed PredRNN-V2 progressively improve the quality of the prediction results.

\begin{table}[t]
\vskip 0.05in
  \caption{\revise{An ablation study on the zigzag memory flow. We compare the baseline PredRNN with two alternatives that truncate the memory flow at specific positions in the transition path of $\mathcal{M}_t^l$.}}
  \vskip -0.05in
  \label{tab:ablation_flow}
  \centering
  \begin{tabular}{lcc}
    \toprule
    Model & MSE ($\downarrow$) \\
    \midrule
    PredRNN (w/o RSS or memory decoupling) & 56.8 \\
    - Truncated at $\mathcal{M}_t^1 \rightarrow \ldots \rightarrow \mathcal{M}_t^L$ (bottom-up) & 57.6 \\
    - Truncated at $\mathcal{M}_{t}^L \rightarrow \mathcal{M}_{t+1}^1$ (top-down)
    & 59.7 \\
    \bottomrule
  \end{tabular}
  \vspace{-10pt}
\end{table}

\begin{figure*}[t]
  \centering
  \subfigure[Responses to sudden step changes]{
    \includegraphics[width=0.98\columnwidth]{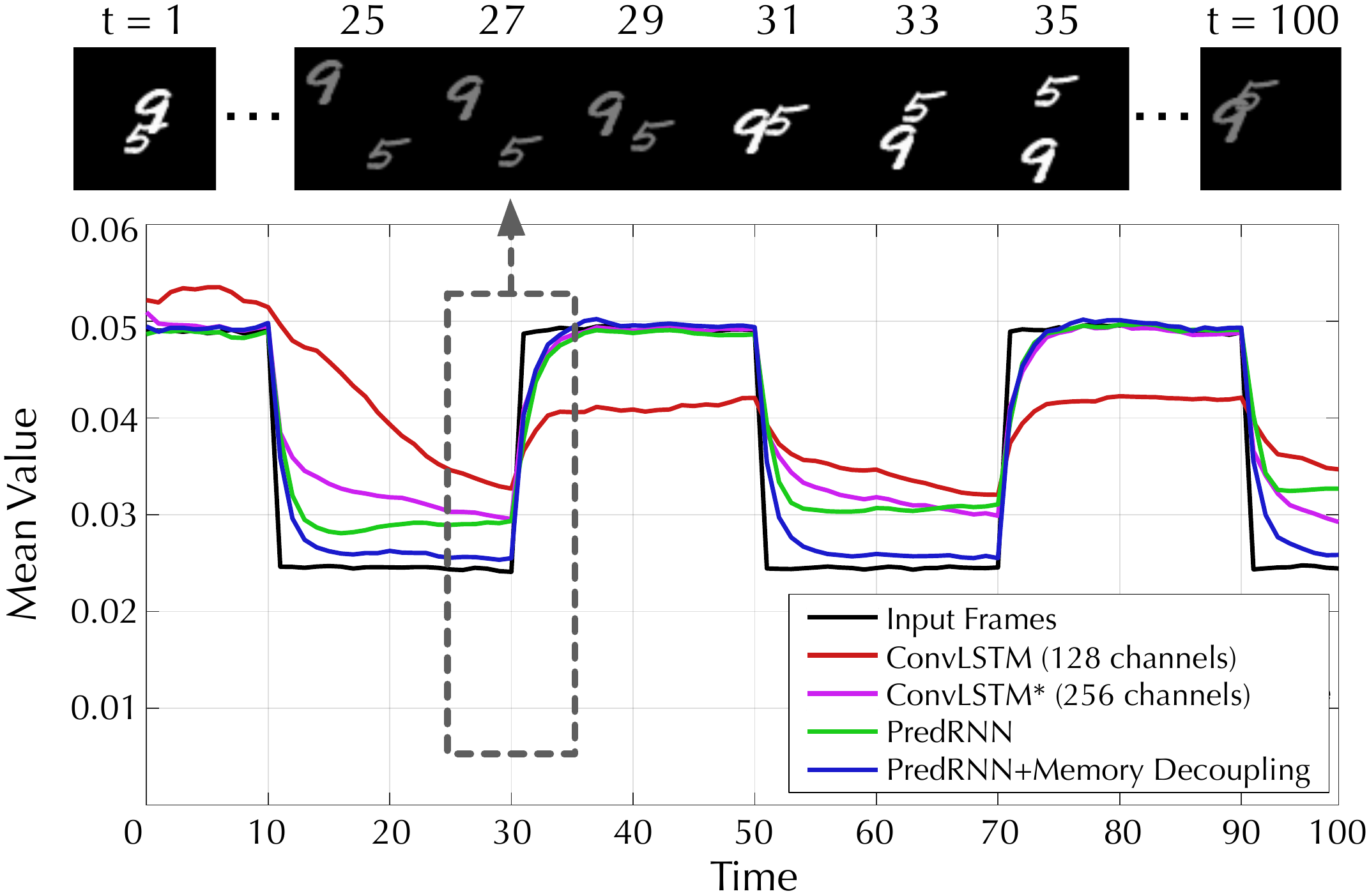}
  }\label{fig:short_term}
  \subfigure[Responses to long-term and short-term dynamics]{
    \includegraphics[width=0.98\columnwidth]{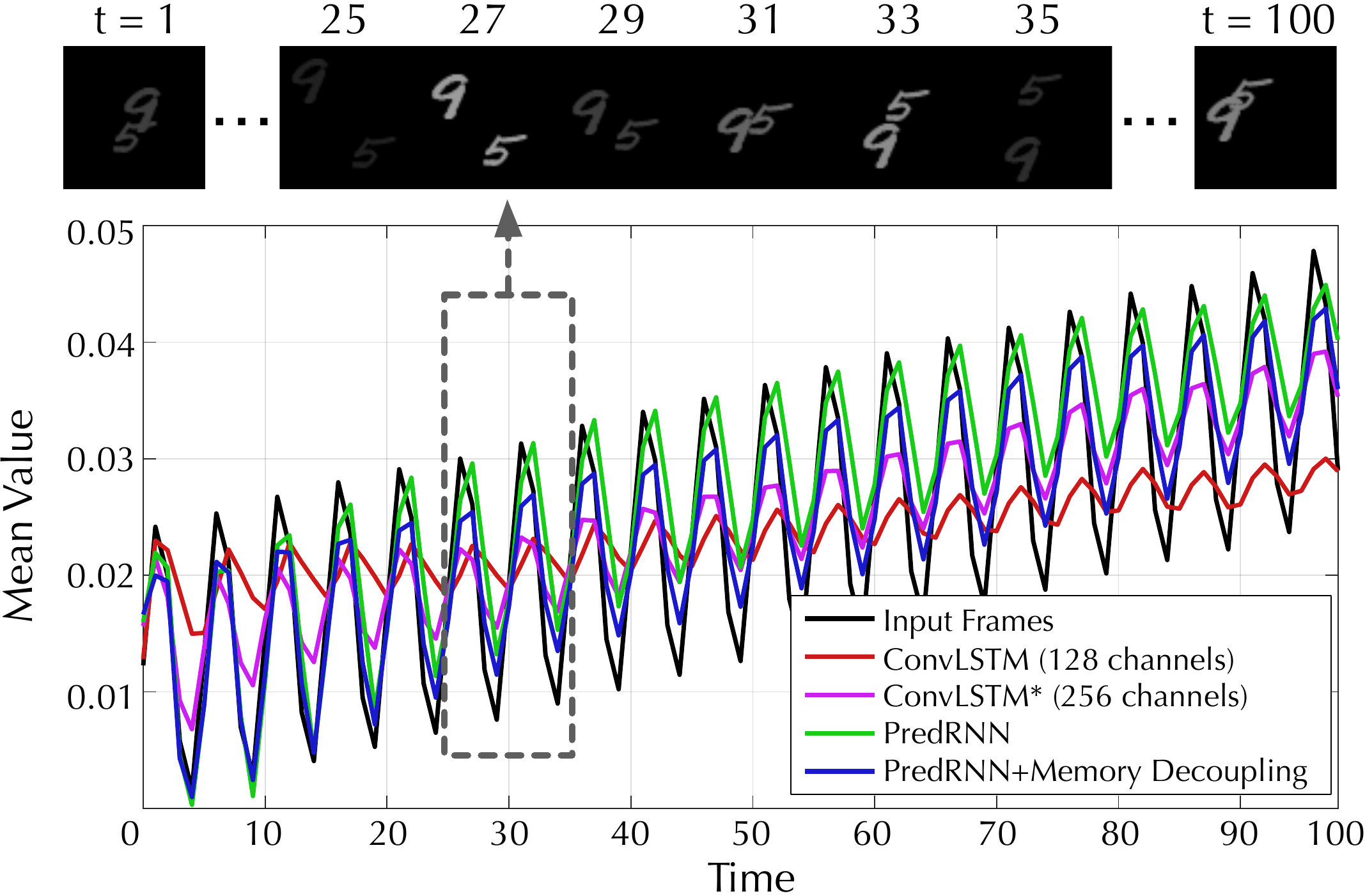}
  }\label{fig:long_term}
  \vskip -0.1in
\caption{Model performance under different types of temporal dynamics, 
\revise{where the models take as inputs the real observations at all timesteps, and the Y axis represents the mean value of predictions averaged over $100$ test samples.
We show two ground-truth sequences as examples at the top of the figure for a better intuitive understanding of the two types of pixel intensity changes. For each type, we apply the same rate of changes to all test samples.} 
With memory-decoupled ST-LSTMs, our model is shown to (a) respond more rapidly to unexpected, sudden variations, and (b) simultaneously capture temporal dynamics at different timescales. 
}
\label{fig:long_short_term}
\vspace{-10pt}
\end{figure*}

\myparagraph{Ablation studies on the memory flow} 
\revise{
To understand the benefit of the spatiotemporal memory flow, we compare the performance of (i) truncating the bottom-up state transition path of PredRNN from $\mathcal M_t^1$ to $\mathcal M_t^L$ and (ii) truncating the top-down transition path from $\mathcal M_t^L$ to $\mathcal M_{t+1}^1$. 
\fig{fig:grad_a} shows the normalized gradient values of $\big\|\nabla_{\mathcal{H}_t^1}\mathcal{L}_{T+K}\big\|$ for $1 \leq t < 20$, which demonstrates the benefit of the zigzag memory flow in learning long-term data trends due to the ability of alleviating the problem of vanishing gradients.
\tab{tab:ablation_flow} gives the quantitative results in MSE, indicating that the top-down transition path contributes more to the final performance. 
Furthermore, in \fig{fig:c_m_gate}, we study the effect of the memory flow in long-term modeling by analyzing the saturation ratio among all elements in the forget gates ($f_t<0.1$) for the temporal memory cell $\mathcal{C}_t^l$.
A high saturation ratio indicates that the model tends to block the information flow of long-term trends.
Compared with ConvLSTM, the spatiotemporal memory flow releases the modeling capability of $\mathcal{C}_t^l$ for long-term dynamics.
}

\myparagraph{Ablation study on the memory decoupling}
To show that memory decoupling facilitates both long-term and short-term dependencies, as shown in \fig{fig:long_short_term}, we make the pixel intensity of the images change regularly or irregularly over time. 
Thanks to the decoupled memory cells of ST-LSTMs, our approach can respond to sudden changes more rapidly and adapt to video dynamics at different timescales. 
%
\revise{
As shown by the purple curves, we also include the results of a larger ConvLSTM network with a model size significantly larger than that of \textit{PredRNN+Memory Decoupling} ($65.96$ MB vs. $23.86$ MB). Nonetheless, our model responds more quickly to the rapidly changing pixel intensity.}
In \tab{tab:v2_general}, we use the MIM model \cite{wang2019memory}, which is also based on the ST-LSTM unit, to specifically show the generality of the memory decoupling loss to different network backbones.

\begin{figure*}[t]
  \centering
  \subfigure[Frame-wise PSNR ($\uparrow$)]{
    \includegraphics[width=0.6\columnwidth]{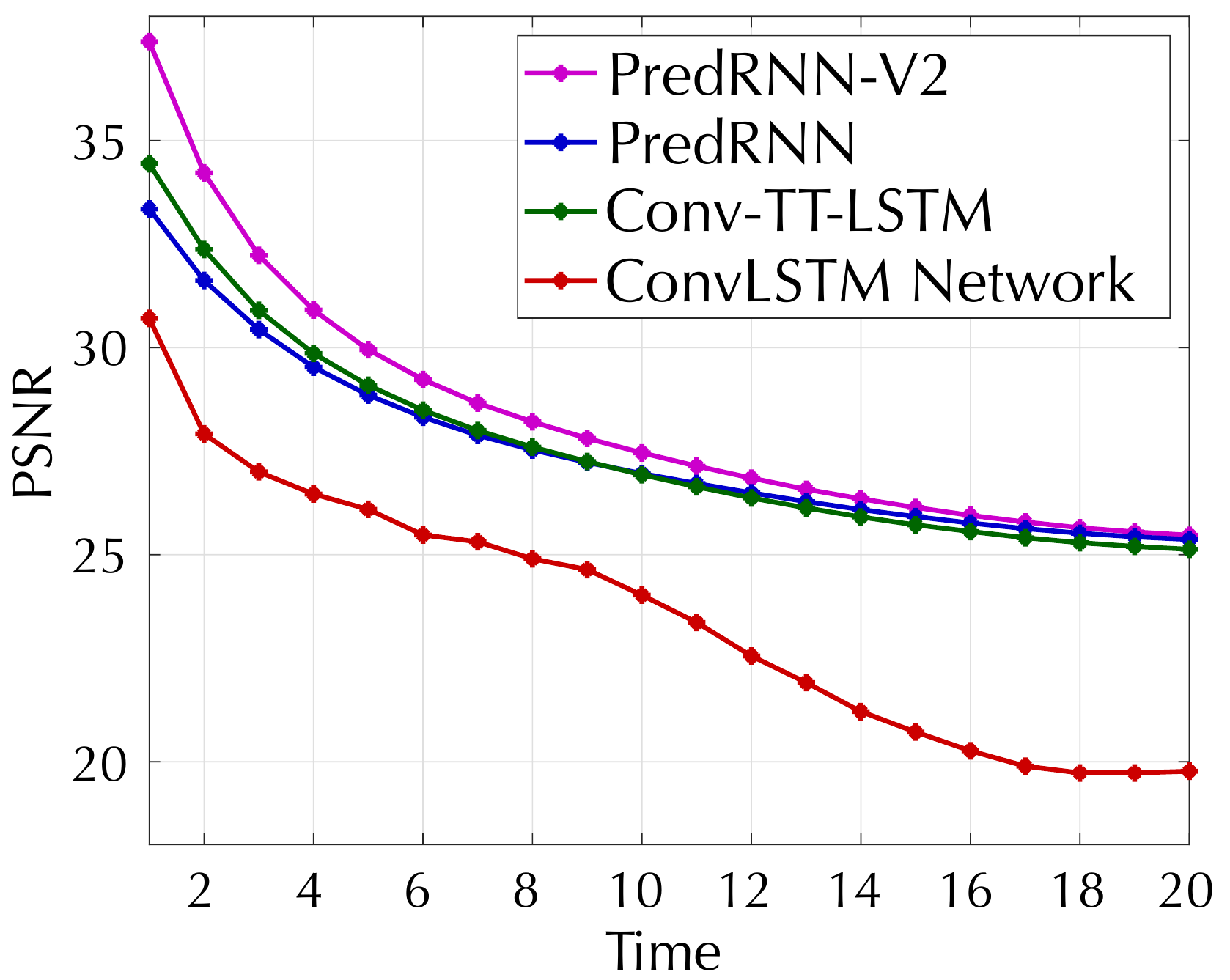}
  }
  \subfigure[Frame-wise SSIM ($\uparrow$)]{
    \includegraphics[width=0.6\columnwidth]{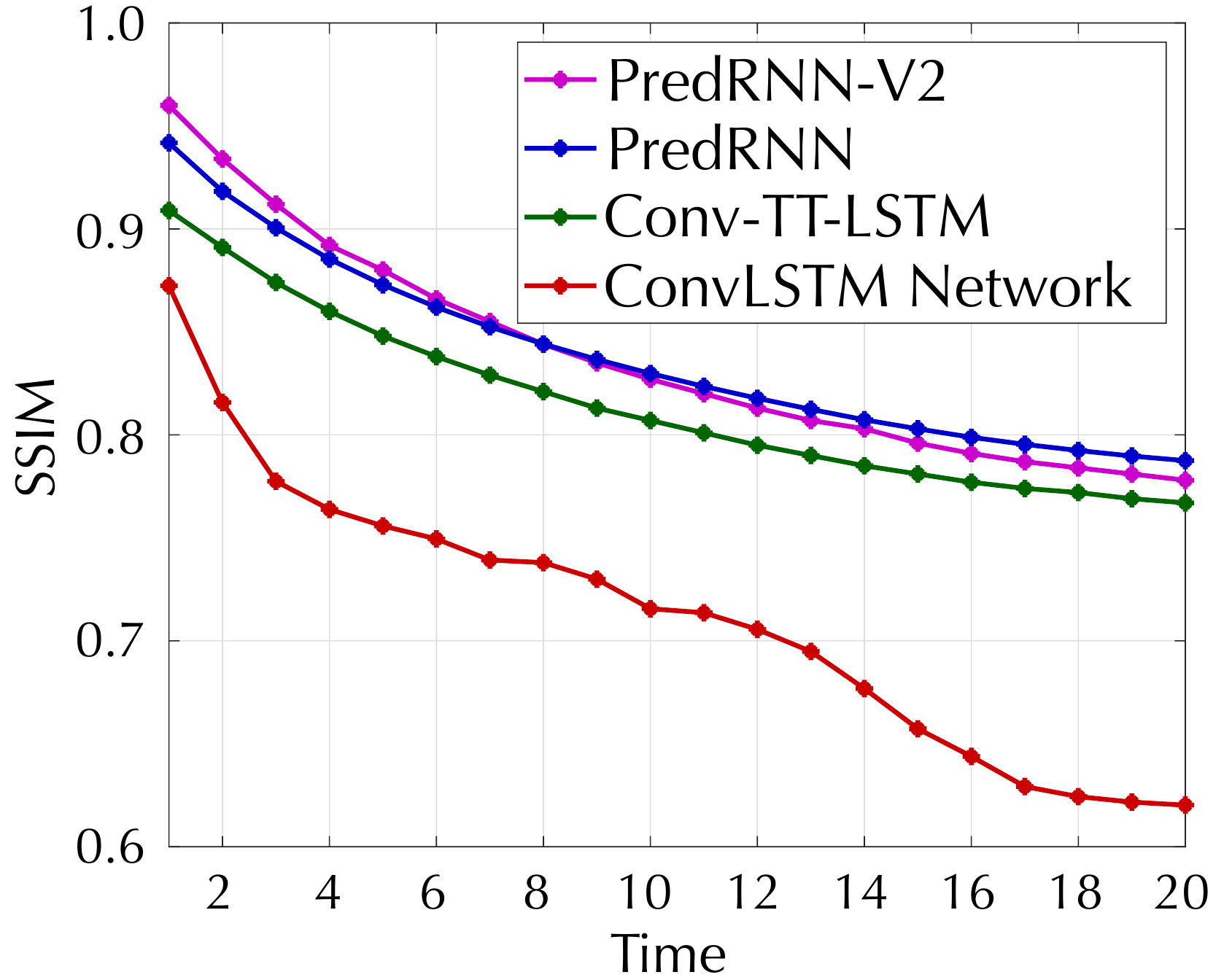}
}
  \subfigure[Frame-wise LPIPS ($\downarrow$)]{
    \includegraphics[width=0.6\columnwidth]{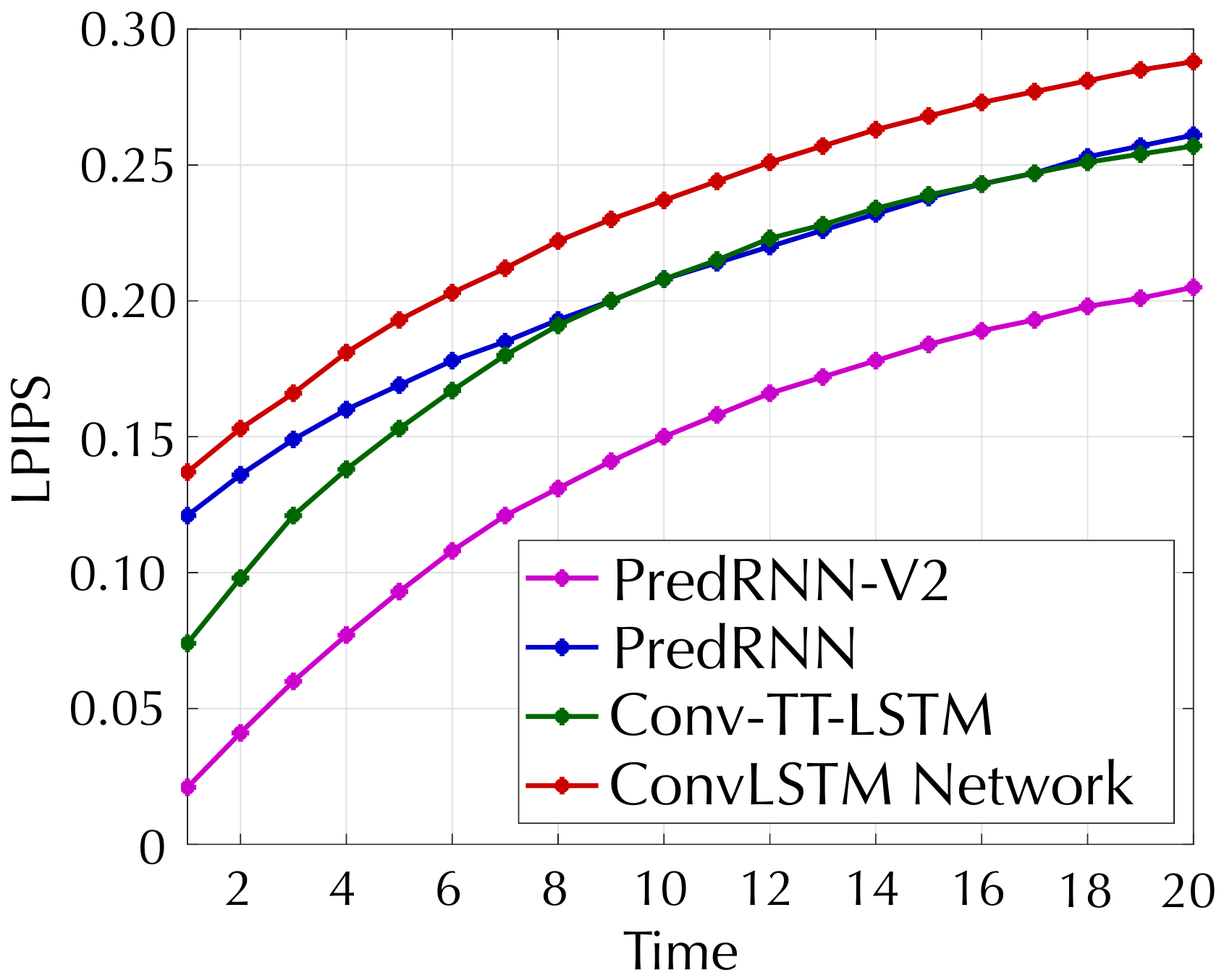}
}
\vskip -0.15in
\caption{Frame-wise results on KTH. The prediction horizon is $10$ timesteps at training time and $20$ timesteps at test time.}
\label{fig:kth_frame}
\vspace{-15pt}
\end{figure*}

\begin{table}[t]
\vskip 0.05in
  \caption{Effect of memory decoupling and reverse scheduled sampling on different network backbones (sorted by publish date).
  }
  \vskip -0.05in
  \setlength{\tabcolsep}{6pt}
  \label{tab:v2_general}
  \centering
  \begin{tabular}{lcccc}
    \toprule
    Backbone & Decoupling & RSS & MSE ($\downarrow$) & Perf. gain \\
    \midrule
    \multirow{2}{*}{ConvLSTM \cite{shi2015convolutional}} & $\times$ & $\times$ & 103.3 & \multirow{2}{*}{\revise{37.8\%}} \\
     & \revise{$\times$} & \revise{$\bigcirc$} & \revise{64.2} &  \\
    \midrule
    \multirow{3}{*}{PredRNN} & $\times$ & $\times$ & 56.8 & \multirow{3}{*}{14.8\%}\\
    & $\bigcirc$ & $\times$ & 51.1 & \\
    & $\bigcirc$ & $\bigcirc$ & 48.4 & \\
    \midrule
    \multirow{2}{*}{PredRNN++ \cite{wang2018predrnn++}} & $\times$ & $\times$ & 46.5 & \multirow{2}{*}{3.7\%} \\
    & $\bigcirc$ & $\bigcirc$ & 44.8 \\
    \midrule
    \multirow{2}{*}{E3D-LSTM \cite{wang2019eidetic}} & $\times$ & $\times$ & 44.2 & \multirow{2}{*}{4.8\%} \\
    & $\bigcirc$ & $\bigcirc$ & 42.1 \\
    \midrule
    \multirow{3}{*}{MIM \cite{wang2019memory}} & $\times$ & $\times$ & 52.0 & \multirow{3}{*}{11.9\%} \\
    & $\bigcirc$ & $\times$ & 47.9 \\
    & $\bigcirc$ & $\bigcirc$ & 45.8 \\
    \midrule
    \multirow{2}{*}{Conv-TT-LSTM \cite{su2020convolutional}} & $\times$ & $\times$ & 64.3 & \multirow{2}{*}{8.2\%} \\
    & $\times$ & $\bigcirc$ & 59.0 \\
    \midrule
    \multirow{2}{*}{MotionRNN \cite{wu2021motionrnn}} & $\times$ & $\times$ & 52.4 & \multirow{2}{*}{5.7\%} \\
    & $\bigcirc$ & $\bigcirc$ & 49.4 \\
    \bottomrule
  \end{tabular}
  \vspace{-10pt}
\end{table}

\begin{table}[t]
\vskip 0.05in
  \caption{Ablation study on the training schemes. For RSS, $\epsilon_k$ changes from $\epsilon_s$ to $\epsilon_e$. Models are trained w/o the decoupling loss.}
  \vskip -0.05in
  \label{tab:mnist_rss}
  \centering
  \begin{tabular}{lcccccc}
    \toprule
    Method & $\epsilon_s$ & $\epsilon_e$ & RSS mode & MSE ($\downarrow$) \\
    \midrule
    PredRNN & - & - & - & 57.3 \\
    + Scheduled sampling \cite{bengio2015scheduled} & - & - & - & 56.8 \\
    \midrule
    \multirow{3}{*}{+ 1st strategy in \fig{fig:Strategy_1}} & \multirow{3}{*}{0.0} & \multirow{3}{*}{1.0} & Linear & 51.8 \\
     &  &  & Sigmoid & 53.9 \\
     &  &  & Exponential & 51.8 \\
    \midrule
    \multirow{3}{*}{+ 2nd strategy in \fig{fig:Strategy_2}} & \multirow{3}{*}{0.5} & \multirow{3}{*}{1.0} & Linear & 51.9 \\
     &  &  & Sigmoid & 50.9 \\
     &  &  & Exponential & \textbf{50.6} \\
    \bottomrule
  \end{tabular}
  \vspace{-10pt}
\end{table}

\myparagraph{Ablation study on the reverse scheduled sampling}
We compare different combinations of the original and the reverse scheduled sampling techniques.
From \tab{tab:mnist_rss}, the second strategy in \fig{fig:rss_schemes} with an exponentially increased $\epsilon_k$ performs best. It is because the encoder-forecaster discrepancy can be effectively reduced by keeping their probabilities of sampling the true context frames close to each other in the early stage of training.
To further demonstrate that the reverse scheduled sampling can contribute to learning long-term dynamics, \revise{we perform empirical analyses on the long-term gradients in \fig{fig:grad_b} and the saturation ratio of forget gates in \fig{fig:c_m_gate}.}
For the gradient analysis, we evaluate the gradients of the encoder's hidden states with respect to the loss functions at different output timesteps, and average the results over the entire input sequence: $\frac{1}{T-1}\sum_{\tau=2}^T\big\| \nabla_{\mathcal{H}_\tau^1}\mathcal{L}_t \big\|$, $t\in [T+1,T+K]$.
The normalized gradient curves show that the context information can be encoded more effectively by using reverse scheduled sampling.
%
\revise{In \fig{fig:c_m_gate}, we find that with RSS, both the ConvLSTM network and PredRNN maintain lower saturation ratios for the forget gates in the temporal memory cell $\mathcal{C}_{t}^{l}$, which indicates that RSS improves the learning process of long-term dynamics.
}

\myparagraph{Generality of the proposed techniques}
\tab{tab:v2_general} shows the results of applying the proposed techniques to some existing models that are also based on convolutional RNNs. It is worth noting that, first, our original ST-LSTM has been the foundation of some recent architectures, thus allowing further improvements introduced by memory decoupling.
Besides, the proposed reverse scheduled sampling method is shown to be a general training approach for recurrent models, enabling them to approach the state of the art.
In particular, although PredRNN++ is also based on ST-LSTMs, it only achieves a $3\%$ performance gain. The reason is that, on one hand, like RSS, the \textit{gradient highway unit} of PredRNN++ is also to improve the long-term modeling capability; On the other hand, it models the state transitions from $\mathcal{C}_t^l$ to $\mathcal{M}_t^l$, which implicitly allows $\mathcal{M}$ to learn new information beyond what has been leaned by $\mathcal{C}$. 
However, the two components mentioned above in PredRNN++ introduce additional parameters and increase the computational burden. In contrast, memory decoupling and RSS increase the number of model parameters by a negligible amount but have greater benefits for the final performance.

\subsection{KTH Action Dataset}

In this dataset, all compared models are trained across the $6$ action categories by predicting $10$ future frames from $10$ observations. At test time, the prediction horizon is expanded to $20$ timesteps.

\begin{figure*}[t]
  \centering
  \includegraphics[width=\textwidth]{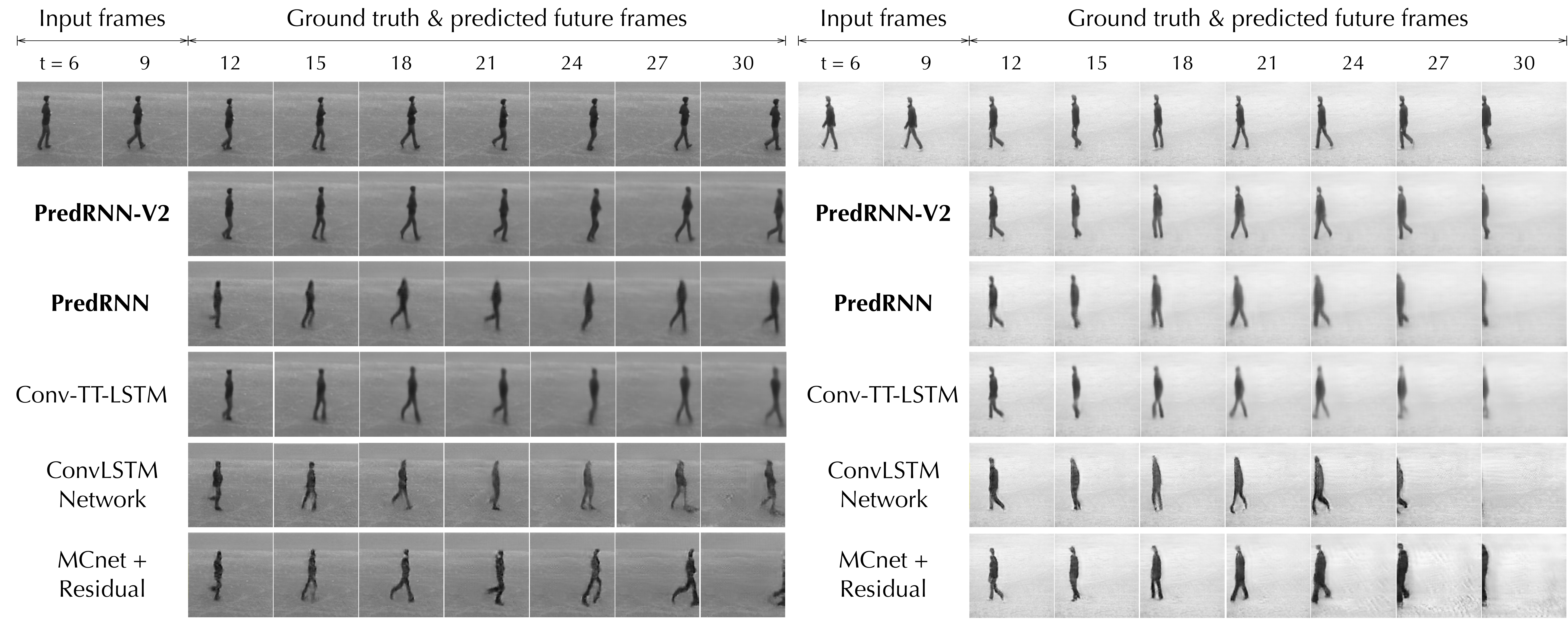}
  \vskip -0.05in
  \caption{Prediction examples on the KTH test set, where we predict $20$ frames into the future based on the past $10$ frames.}
  \label{fig:kth_results}
  \vspace{-10pt}
\end{figure*}

\begin{table}[t]
\vskip 0.05in
  \caption{Results on KTH averaged over $20$ future timesteps.
  }
  \vskip -0.05in
  \label{tab:kth}
  \centering
  \begin{tabular}{lccc}
    \toprule
    Model & PSNR ($\uparrow$) & SSIM ($\uparrow$) & LPIPS ($\downarrow$)  \\
    \midrule
    ConvLSTM \cite{shi2015convolutional} & 23.58 & 0.712 & 0.231 \\
    MCnet + Residual \cite{Villegas2017Decomposing} & 26.29 & 0.806 & - \\
    TrajGRU \cite{shi2017deep} & 26.97 & 0.790 & - \\
    DFN \cite{de2016dynamic} & 27.26 & 0.794 & - \\
    Conv-TT-LSTM \cite{su2020convolutional} & 27.62 & 0.815 & 0.196 \\
    \midrule
    PredRNN & 27.55 & \textbf{0.839} & 0.204  \\
    PredRNN-V2 & \textbf{28.37} & 0.838 & \textbf{0.139}  \\
    \bottomrule
  \end{tabular}
  \vspace{-10pt}
\end{table}

We adopt the Peak Signal to Noise Ratio (PSNR) from the previous literature as the third evaluation metric, in addition to SSIM and LPIPS. Like MSE, PSNR also estimates the pixel-level similarity of two images (higher is better). 
The evaluation results of different methods are shown in \tab{tab:kth}, and the corresponding frame-wise comparisons are shown in \fig{fig:kth_frame}, from which we have two observations: First, our models show significant improvements in both short-term and long-term predictions over the ConvLSTM network. 
Second, with the newly proposed memory decoupling and reverse scheduled sampling, PredRNN-V2 improves the conference version by a large margin in LPIPS (from 0.204 to 0.139). 
As mentioned above, LPIPS is more sensitive to perceptual human judgments, indicating that PredRNN-V2 has a stronger ability to generate high-fidelity images.
In accordance with these results, we can see from the visual examples in
\fig{fig:kth_results} that our approaches (especially PredRNN-V2) obtain more accurate predictions of future movement and body details.
By decoupling memory states, PredRNN-V2 learns to capture the complex spatiotemporal variations from different timescales.

\subsection{Precipitation Forecasting from Radar Echoes}

The accurate prediction of the movement of radar echoes in the next $0$-$2$ hours is the foundation of precipitation forecasting. In this dataset, each sequence contains $10$ input frames and $10$ output frames, covering the historical data for the past hour and that for the future hour. It is challenging because the echoes tend to have non-rigid shapes and may move, accumulate or dissipate rapidly due to complex atmospheric physics, which makes it important to learn the dynamics in a unified spatiotemporal feature space.

\begin{figure*}[t]
  \centering
  \includegraphics[width=0.98\textwidth]{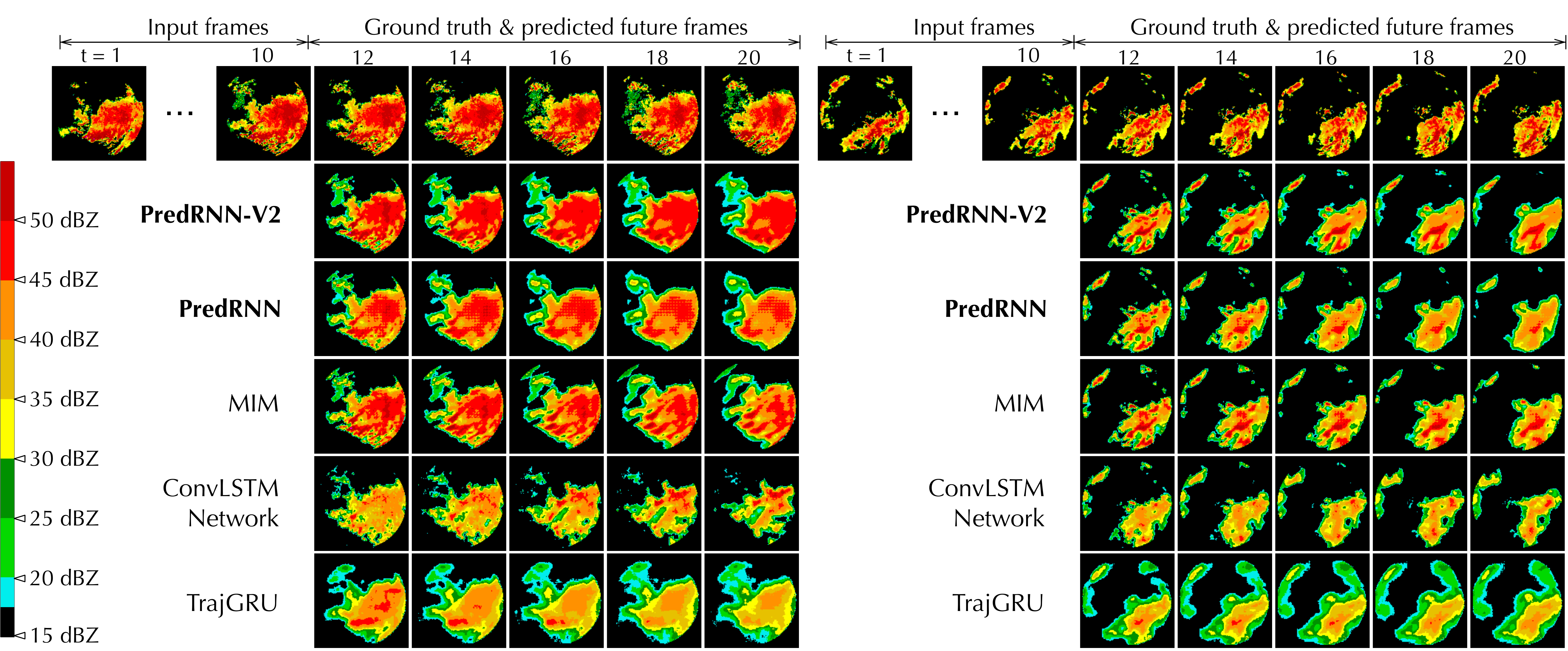}
  \vskip -0.05in
  \caption{Prediction examples on the radar echo test set, in which $10$ future frames are generated from the past $10$ observations.}
  \label{fig:radar}
  \vspace{-10pt}
\end{figure*}

\begin{table}[t]
\vskip 0.05in
  \caption{Quantitative results on the radar echo dataset.}
  \vskip -0.05in
  \label{tab:radar}
  \centering
  \begin{tabular}{lcccc}
    \toprule
    Model & MSE ($\downarrow$) & CSI-30 ($\uparrow$) & CSI-40 ($\uparrow$) & CSI-50 ($\uparrow$) \\
    \midrule
    TrajGRU \cite{shi2017deep} & 68.3 & 0.309 & 0.266 & 0.211 \\
    ConvLSTM \cite{shi2015convolutional}  & 63.7 & 0.381 & 0.340 & 0.286  \\
    MIM \cite{wang2019memory} & 39.3 & 0.451 & 0.418 & 0.372 \\
    \midrule
    PredRNN & 39.1 & 0.455 & 0.417 & 0.358  \\
    PredRNN-V2 & \textbf{36.4} & \textbf{0.462} & \textbf{0.425} & \textbf{0.378} \\
    \bottomrule
  \end{tabular}
  \vspace{-5pt}
\end{table}

\begin{table}[t]
    \vskip 0.05in
    \caption{Quantitative results on the Traffic4cast dataset.}
    \vskip -0.05in
    \label{tab:traffic4cast_final_result}
    \centering
    \begin{tabular}{lcccc}
    \toprule
    Backbone & Recurrent unit & Decoupl. & RSS & MSE ($10^{-3}$) \\
    \midrule
    U-Net \cite{ronneberger2015u} & None & $\times$ & $\times$ &
    6.992  \\
    \midrule
    \multirow{2}{*}{U-Net + PredRNN} 
    & ST-LSTM & $\times$ & $\times$ & 7.035 \\
    & ST-LSTM & $\bigcirc$ & $\bigcirc$ & \textbf{5.135} \\
    \midrule
    \multirow{3}{*}{CrevNet \cite{yu2020efficient}} & ConvLSTM & $\times$ & $\times$ &
    6.789  \\
    & ST-LSTM & $\times$ & $\times$ & 6.613 \\
    & ST-LSTM & $\bigcirc$ & $\bigcirc$ & 6.506 \\
    \bottomrule
    \end{tabular}
    \vspace{-10pt}
\end{table}

In \tab{tab:radar}, we compare PredRNN with three existing methods that have been shown effective for precipitation forecasting.
Following a common practice, we evaluate the predicted radar maps with the Critical Success Index (CSI). Concretely, we first transform the pixel values back to echo intensities in dBZ, and then take $30$, $40$ and $50$ dBZ as thresholds to compute: $\text{CSI}=\frac{\text{hits}}{\text{hits}+\text{misses}+\text{false\_alarms}}$, where \textit{hits} indicates the true positive, \textit{misses} indicates the false positive, and \textit{false\_alarms} is the false negative.
From \tab{tab:radar}, PredRNN consistently achieves the best performance over all CSI thresholds.
In \fig{fig:radar}, we visualize the predicted radar frames by mapping them into RGB space, where areas with echo values over $40$ dBZ tend to have severe weather. 

\begin{figure}[t]
  \centering
  \includegraphics[width=\columnwidth]{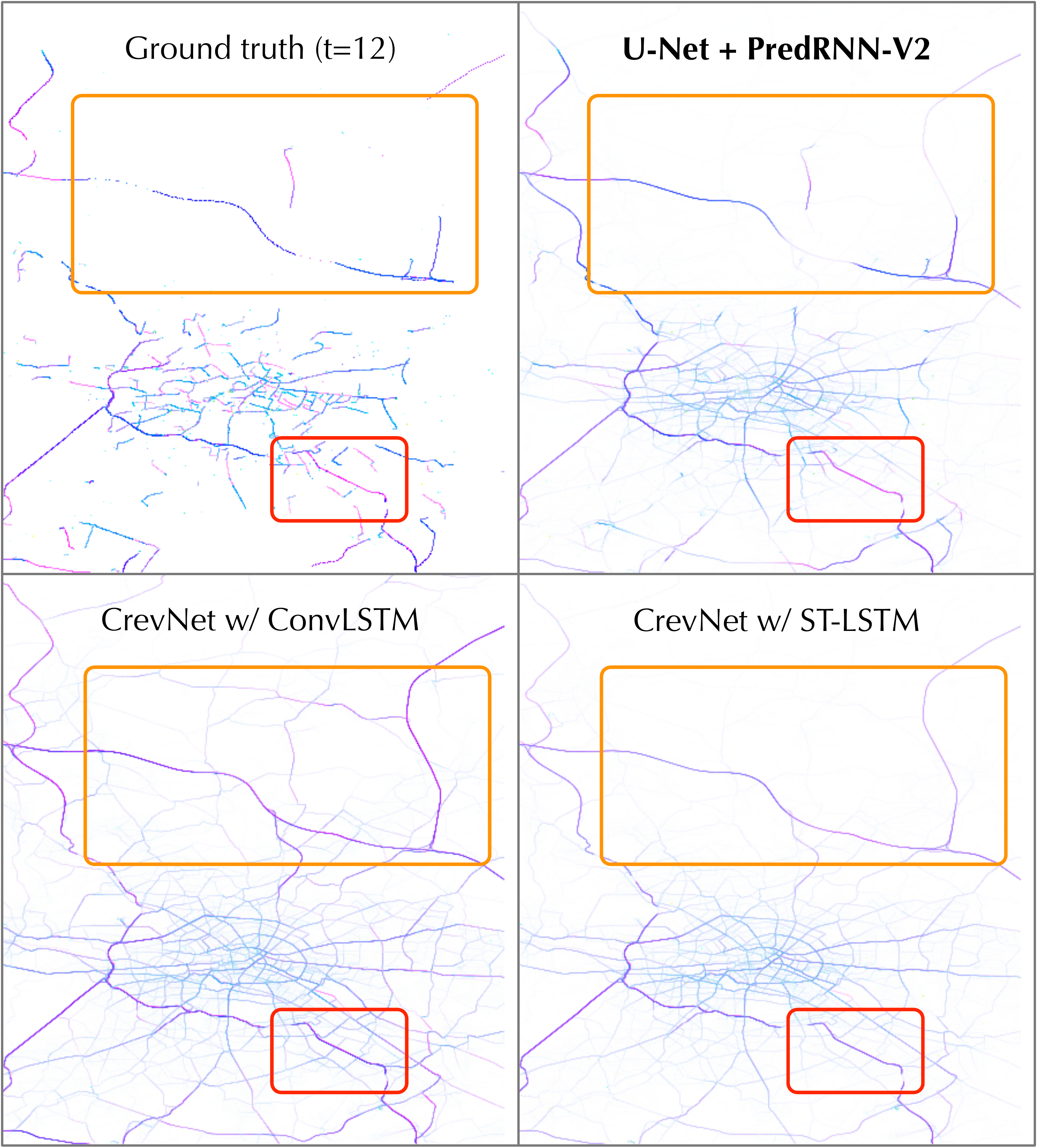}
   \vskip -0.02in
  \caption{Showcases at the last forecasting timestep on Traffic4cast. Please note the differences in areas in the orange and red boxes.}
  \label{fig:traffic4cast_showcase}
  \vspace{-5pt}
\end{figure} 

\subsection{Traffic4Cast Dataset}

We evaluate the proposed methods on Traffic4Cast \cite{Traffic4cast-web-2019}, a real-world dataset with complex and drastically changing information. 
We follow the experimental setups from CrevNet \cite{yu2020efficient}, which is a competitive model on this dataset and use the data of Berlin to construct a training set of $82{,}080$ frames and a test set of $2{,}160$ frames.
The models are trained to predict $3$ future frames from $9$ context frames. All pixel values are normalized to $[0,1]$.

To cope with high-dimensional input frames, we apply the autoencoder architecture of U-Net \cite{ronneberger2015u} to the network backbone of PredRNN. Specifically, the decoder of U-Net contains four ST-LSTM layers, and the CNN encoder takes both traffic flow maps and spatiotemporal memory states as inputs.
We also apply the proposed methods to CrevNet, which uses either ConvLSTM or ST-LSTM as the alternatives of recurrent units in the autoencoder architecture. Therefore, it can be further improved by memory decoupling and reverse scheduled sampling.

In \tab{tab:traffic4cast_final_result}, we follow the standard evaluation metric of MSE from the competition and have the following observations.
Initially, the use of ST-LSTM improves the performance of the predictive models based on U-Net and CrevNet. 
Besides, the proposed memory decoupling and reverse scheduled sampling achieve further improvements over the models based on vanilla ST-LSTMs. 
\fig{fig:traffic4cast_showcase} provides the qualitative comparisons of different variants of our methods based on the architecture of CrevNet.  
As shown in the red box at $t=11$, the final model with memory-decoupled ST-LSTM and reverse scheduled sampling predicts the rapid changes of traffic flows most accurately.

\subsection{Action-Conditioned BAIR Robot Pushing Dataset}

\begin{figure}[t]
\centering
\subfigure[Frame-wise SSIM ($\uparrow$)]{
\includegraphics[width=0.47\columnwidth]{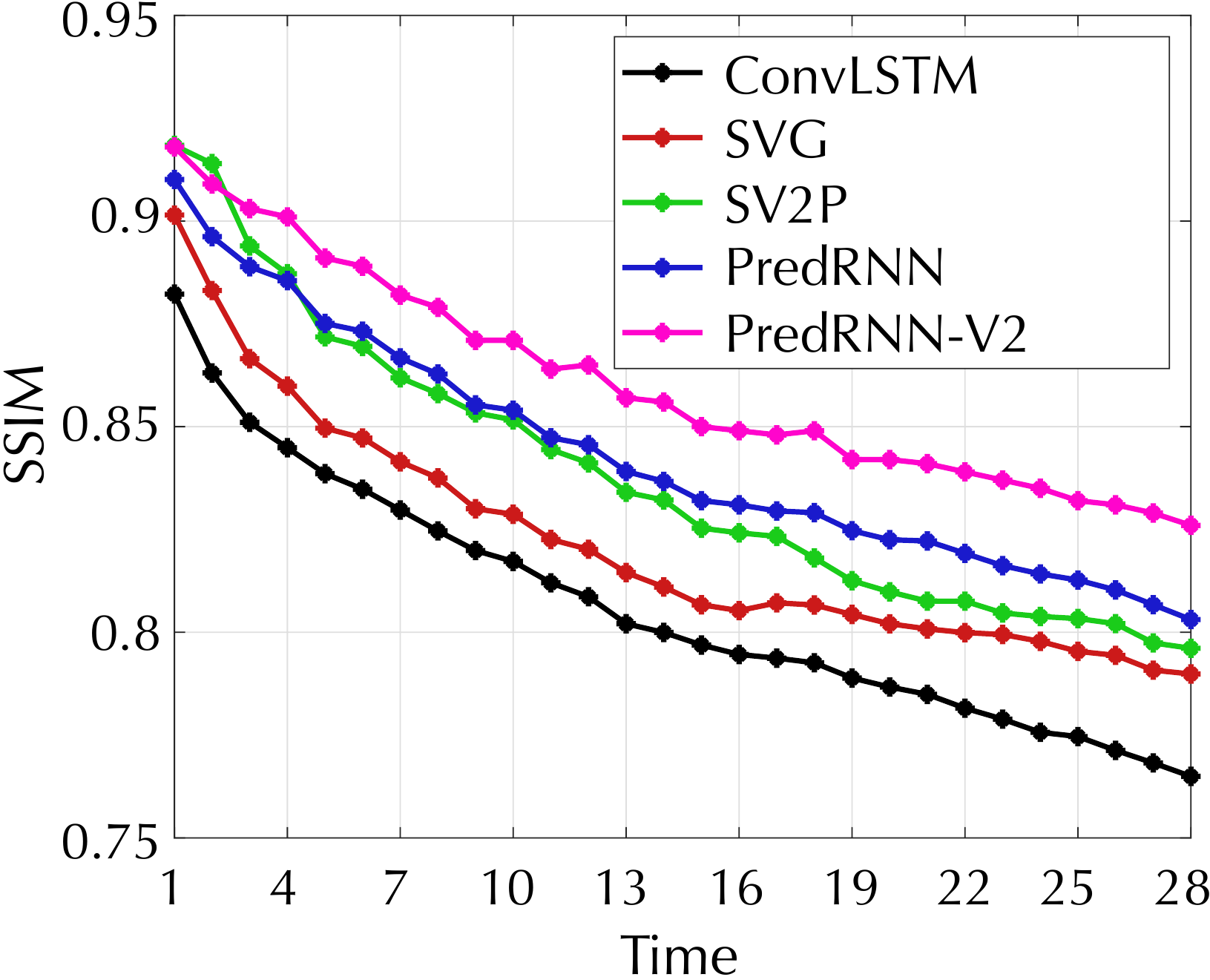}
\label{fig:bair_frame_ssim}
}
\subfigure[Frame-wise PSNR ($\uparrow$)]{
\includegraphics[width=0.47\columnwidth]{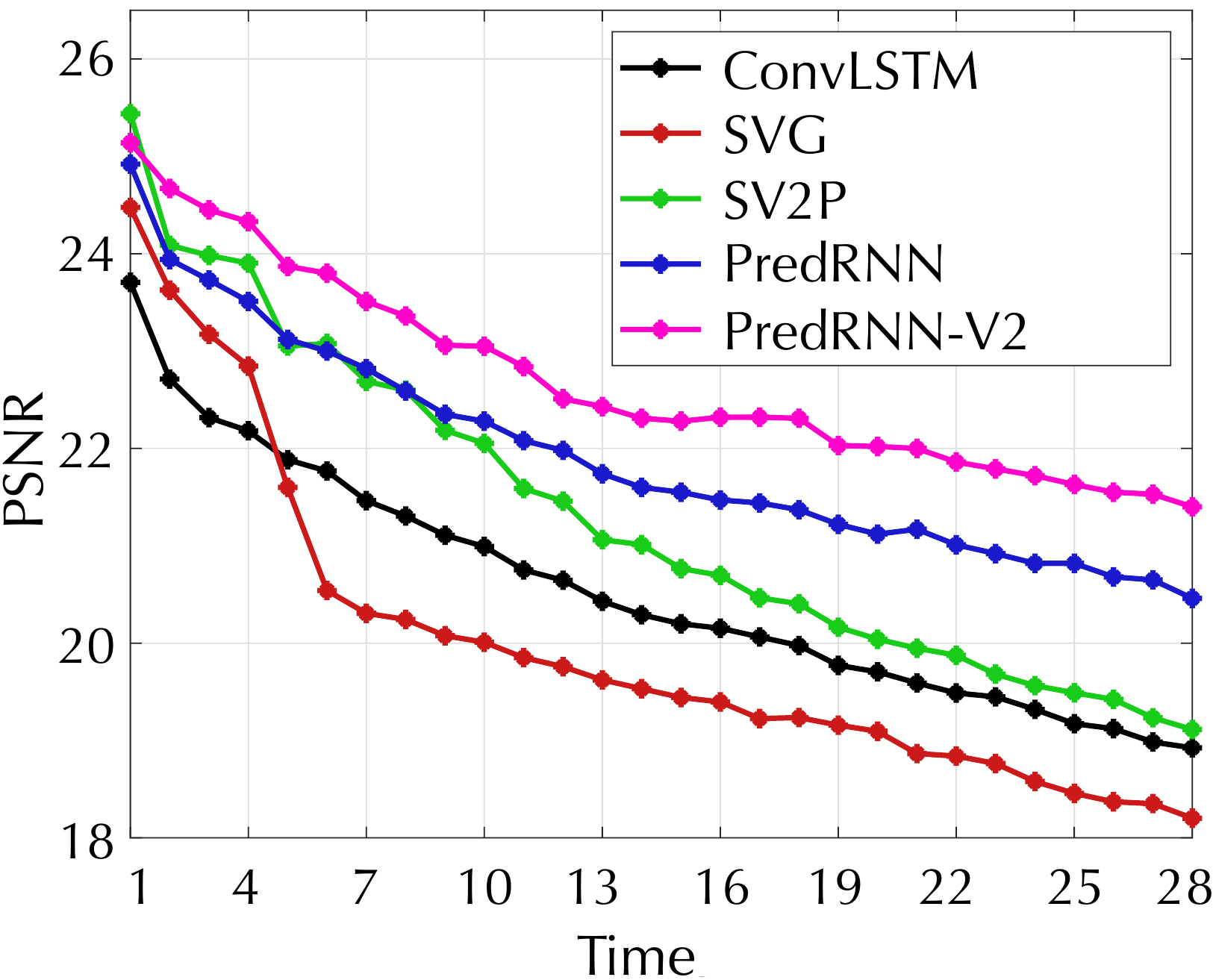}
\label{fig:bair_frame_psnr}
}
\vskip -0.1in
\caption{Frame-wise SSIM and PSNR on the action-conditioned BAIR dataset. Note that the prediction time horizon is $10$ during the training phase and extended to $28$ during testing phase.}
\label{fig:bair_results}
\end{figure}

\begin{table}[t]
\vskip 0.05in
  \caption{Quantitative results on the BAIR robot pushing dataset averaged over $28$ future timesteps in both action-free and action-conditioned experimental settings.
  }
  \vskip -0.05in
  \label{tab:bair}
  \centering
  \begin{tabular}{llccc}
    \toprule
    Inputs & Model & SSIM ($\uparrow$) & PSNR ($\uparrow$) \\
    \midrule
    \multirow{2}{*}{Action-Free}& PredRNN & 0.536 & 16.21 \\
    & PredRNN-V2 & 0.577 & 16.80 \\
    \midrule
    \multirow{5}{*}{Action-Conditioned} & ConvLSTM \cite{wang2019memory} & 0.807 & 20.52 \\
    & SVG \cite{denton2018stochastic} & 0.821 & 19.97  \\
    & SV2P \cite{babaeizadeh2017stochastic} & 0.837 & 21.32 \\
    & PredRNN & 0.843 & 21.93 \\
    & PredRNN-V2 & \textbf{0.861} & \textbf{22.74} \\
    \bottomrule
  \end{tabular}
  \vspace{-5pt}
\end{table}

\begin{figure}[t]
  \centering
  \includegraphics[width=\columnwidth]{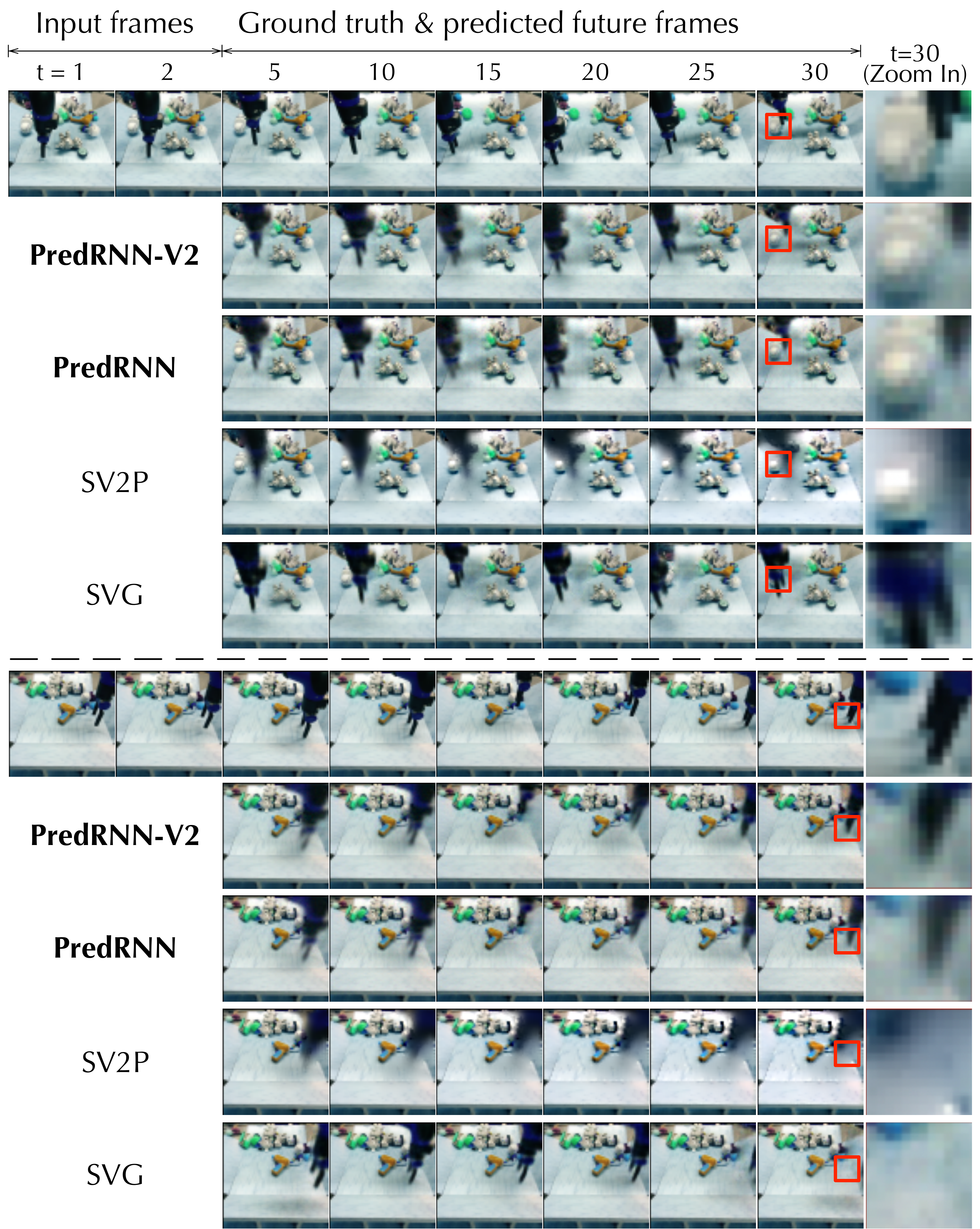}
  \caption{Prediction examples on the BAIR dataset, in which $28$ frames are predicted given the first $2$ frames and the entire action sequence at both encoding and forecasting timesteps. We zoom in the areas bounded by red boxes for clear comparison.}
  \label{fig:bair_showcase}
  \vspace{-5pt}
\end{figure}

We evaluate the action-conditioned PredRNN on the BAIR dataset that is widely used by previous methods, \textit{e.g.}, SVG \cite{denton2018stochastic} and SV2P \cite{babaeizadeh2017stochastic}. 
We follow the experimental setups in SV2P to train the models to predict $10$ frames into the future based on the first $2$ visual observations, as well as the entire action sequence covering both encoding and forecasting timesteps.
During the testing phase, instead, we extend the prediction time horizon to $28$ timesteps for long-term evaluation.
We mainly compare the performance of our approach with that of SVG and SV2P, which are competitive probabilistic models for stochastic video prediction. Specifically for these models, we draw $100$ prediction samples from the prior distribution given a single testing sequence and report the results of the sample with the best SSIM score.

From \tab{tab:bair}, by comparing the proposed models with the action-free PredRNN, we find that the action information plays a significant role in future image generation. The performance of action-free PredRNN is mainly affected by the lack of observable dynamics due to limited input sequence, as well as the unpredictable spatiotemporal changes when future actions are unknown. It can be concluded that \eqn{equ:action} provides an effective method to fuse the action information into the predictive network.

More importantly, in \fig{fig:bair_results}, we have the following observations.
First, the action-conditioned PredRNN and PredRNN-V2 outperform SVG and SV2P significantly, especially for long-term future prediction. Although probabilistic video prediction is useful for many downstream tasks such as epistemic POMDPs with implicit partial observability. The results indicate that in our cases in the BAIR dataset, the visual observations and action sequences provide rich information for future prediction.
Second, the proposed models significantly outperform the action-conditioned ConvLSTM network, where we also apply the action fusion operation in \eqn{equ:action} to each ConvLSTM unit. Again, the results validate the effectiveness of ST-LSTM over ConvLSTM.
Third, with memory decoupling and RSS, PredRNN-V2 presents a consistent improvement over PredRNN and achieves the state of the art. \fig{fig:bair_showcase} shows the qualitative results, in which our model predicts the future position of commanded gripper more precisely and largely enriches the details of objects.

\begin{figure}[t]
  \centering
  \includegraphics[width=\columnwidth]{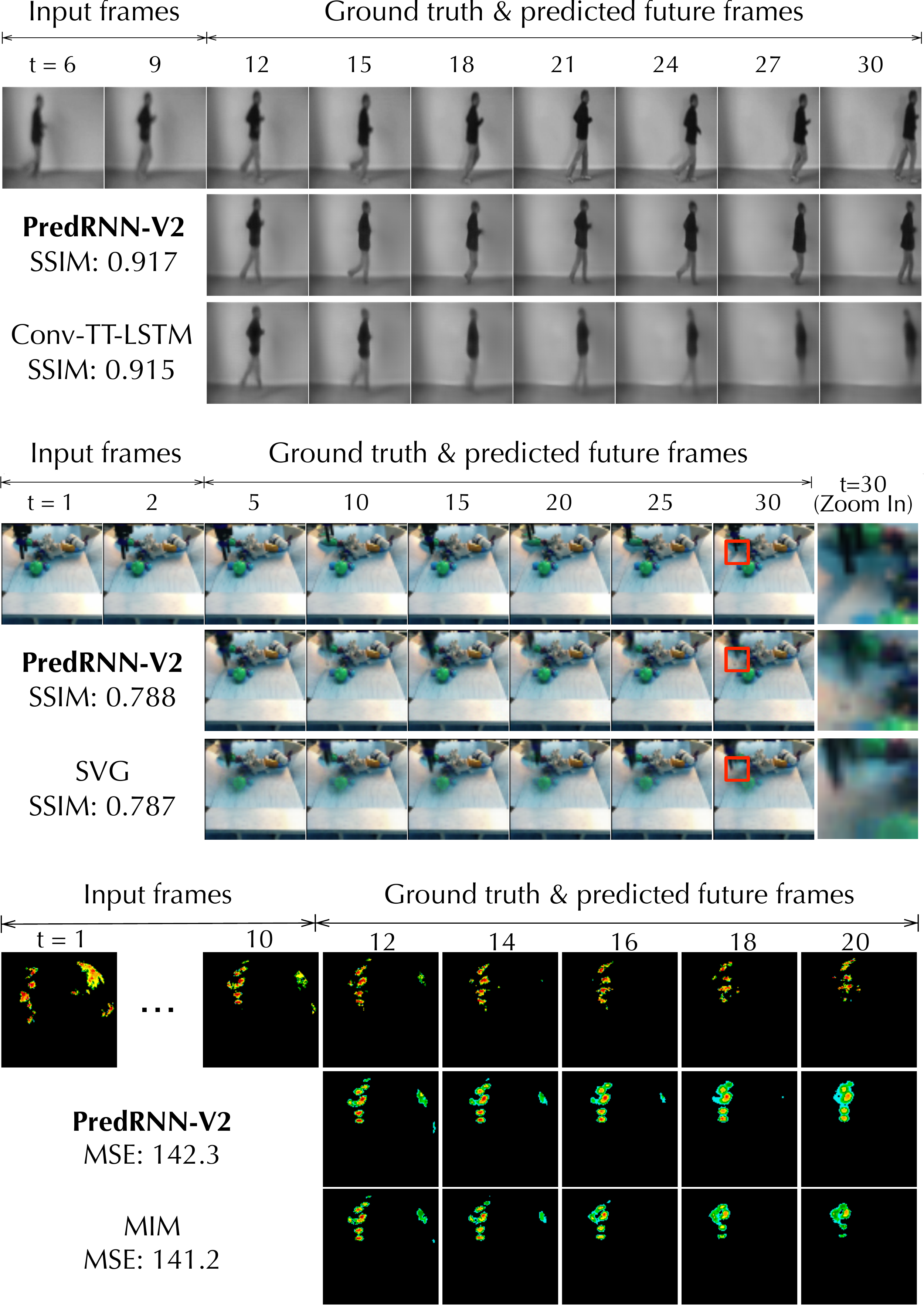}
  \vskip -0.05in
  \caption{
  \revise{Examples that PredRNN-V2 yields comparable quantitative results with the previous art. 
  }
  }
  \label{fig:comparable}
  \vspace{-5pt}
\end{figure}

\subsection{Model Limitations}

\revise{
To explore on what conditions the proposed model contributes the most, we visualize the examples that have comparable quantitative results in SSIM and MSE with the second-best models on the KTH, radar, and BAIR datasets.
As shown in \fig{fig:comparable}, on KTH and BAIR, \textit{i.e.}, datasets with clear visual structures, PredRNN-V2 significantly outperforms the other models in the visual quality of the generated future images. It makes more accurate predictions about future positions of the character (for KTH) and the robot arm (for BAIR).
However, as for the radar echo dataset, \fig{fig:comparable} reveals the limitation of our approach in dealing with the special cases of the long-tail distribution of the dataset. In these cases, the radar observations show very sparse but still significant areas of high intensity echoes.
Although PredRNN-V2 still achieves a better performance than the compared model in the forecast of future positions of the high-intensity areas, it cannot perfectly model the dynamics of the sparse echoes.
Precipitation nowcasting is an extremely difficult research problem for existing spatiotemporal predictive learning approaches.
A future research direction for mitigating the long-tail effect is to explicitly consider the complex physical properties and combine them with deep learning models.
}

\section{Conclusion}

In this paper, we propose a recurrent network named PredRNN for spatiotemporal predictive learning. 
With a new Spatiotemporal LSTM unit, PredRNN models the short-term deformations in spatial appearance and the long-term dynamics over multiple frames simultaneously.
The core of the Spatiotemporal LSTM is a zigzag memory flow that propagates across stacked recurrent layers vertically and through all temporal states horizontally, which enables the interactions of the hierarchical memory representations at different levels of PredRNN. 
Building upon the conference version of this paper, we introduce a new method to decouple the twisted memory states along the horizontal and the zigzag pathway of recurrent state transitions. It enables the model to benefit from learning distributed representations that could cover different aspects of spatiotemporal variations.
Furthermore, we also propose a new curriculum learning strategy named reverse scheduled sampling, which forces the encoding part of PredRNN to learn temporal dynamics from longer periods of the context frames. 
%
Our approach achieves state-of-the-art performance on synthetic and natural spatiotemporal datasets, including both action-free and action-conditioned predictive learning scenarios.


%



\ifCLASSOPTIONcompsoc
  \section*{Acknowledgments}
\else
  \section*{Acknowledgment}
\fi

This work was supported by the National Natural Science Foundation of China (62022050, 62106144 and 62021002). 
M. Long was supported by Beijing Nova Program (Z201100006820041) and BNRist Innovation Fund (BNR2021RC01002).
Y. Wang was supported by Shanghai Municipal Science and Technology Major Project (2021SHZDZX0102) and Shanghai Sailing Program (21Z510202133). 
The work was in part done when Y. Wang was a student at Tsinghua University. Y. Wang and H. Wu contributed equally to this work.

\ifCLASSOPTIONcaptionsoff
  \newpage
\fi



%

\bibliographystyle{IEEEtran}
\bibliography{IEEEabrv,yunbo.bib}

\begin{thebibliography}{10}
\providecommand{\url}[1]{#1}
\csname url@samestyle\endcsname
\providecommand{\newblock}{\relax}
\providecommand{\bibinfo}[2]{#2}
\providecommand{\BIBentrySTDinterwordspacing}{\spaceskip=0pt\relax}
\providecommand{\BIBentryALTinterwordstretchfactor}{4}
\providecommand{\BIBentryALTinterwordspacing}{\spaceskip=\fontdimen2\font plus
\BIBentryALTinterwordstretchfactor\fontdimen3\font minus
  \fontdimen4\font\relax}
\providecommand{\BIBforeignlanguage}[2]{{%
\expandafter\ifx\csname l@#1\endcsname\relax
\typeout{** WARNING: IEEEtran.bst: No hyphenation pattern has been}%
\typeout{** loaded for the language `#1'. Using the pattern for}%
\typeout{** the default language instead.}%
\else
\language=\csname l@#1\endcsname
\fi
#2}}
\providecommand{\BIBdecl}{\relax}
\BIBdecl

\bibitem{shi2015convolutional}
X.~Shi, Z.~Chen, H.~Wang, D.-Y. Yeung, W.-K. Wong, and W.-c. Woo,
  ``Convolutional {LSTM} network: A machine learning approach for precipitation
  nowcasting,'' in \emph{NeurIPS}, 2015, pp. 802--810.

\bibitem{wang2017predrnn}
Y.~Wang, M.~Long, J.~Wang, Z.~Gao, and S.~Y. Philip, ``Pred{RNN}: Recurrent
  neural networks for predictive learning using spatiotemporal lstms,'' in
  \emph{NeurIPS}, 2017, pp. 879--888.

\bibitem{xu2018predcnn}
Z.~Xu, Y.~Wang, M.~Long, and J.~Wang, ``Pred{CNN}: Predictive learning with
  cascade convolutions,'' in \emph{IJCAI}, 2018, pp. 2940--2947.

\bibitem{wang2019memory}
Y.~Wang, J.~Zhang, H.~Zhu, M.~Long, J.~Wang, and P.~S. Yu, ``Memory in memory:
  A predictive neural network for learning higher-order non-stationarity from
  spatiotemporal dynamics,'' in \emph{CVPR}, 2019, pp. 9154--9162.

\bibitem{wu2017learning}
J.~Wu, E.~Lu, P.~Kohli, B.~Freeman, and J.~Tenenbaum, ``Learning to see physics
  via visual de-animation,'' in \emph{NeurIPS}, 2017, pp. 153--164.

\bibitem{van2018relational}
S.~Van~Steenkiste, M.~Chang, K.~Greff, and J.~Schmidhuber, ``Relational neural
  expectation maximization: Unsupervised discovery of objects and their
  interactions,'' in \emph{ICLR}, 2018.

\bibitem{kipf2018neural}
T.~Kipf, E.~Fetaya, K.-C. Wang, M.~Welling, and R.~Zemel, ``Neural relational
  inference for interacting systems,'' in \emph{ICML}, 2018, pp. 2688--2697.

\bibitem{xu2019unsupervised}
Z.~Xu, Z.~Liu, C.~Sun, K.~Murphy, W.~T. Freeman, J.~B. Tenenbaum, and J.~Wu,
  ``Unsupervised discovery of parts, structure, and dynamics,'' in \emph{ICLR},
  2019.

\bibitem{wang2019eidetic}
Y.~Wang, L.~Jiang, M.-H. Yang, L.-J. Li, M.~Long, and L.~Fei-Fei, ``Eidetic {3D
  LSTM}: A model for video prediction and beyond,'' in \emph{ICLR}, 2019.

\bibitem{ha2018recurrent}
D.~Ha and J.~Schmidhuber, ``Recurrent world models facilitate policy
  evolution,'' in \emph{NeurIPS}, 2018, pp. 2450--2462.

\bibitem{hafner2018learning}
D.~Hafner, T.~Lillicrap, I.~Fischer, R.~Villegas, D.~Ha, H.~Lee, and
  J.~Davidson, ``Learning latent dynamics for planning from pixels,'' in
  \emph{ICML}, 2019, pp. 2555--2565.

\bibitem{finn2017deep}
C.~Finn and S.~Levine, ``Deep visual foresight for planning robot motion,'' in
  \emph{ICRA}, 2017, pp. 2786--2793.

\bibitem{ebert2017self}
F.~Ebert, C.~Finn, A.~X. Lee, and S.~Levine, ``Self-supervised visual planning
  with temporal skip connections,'' in \emph{CoRL}, 2017.

\bibitem{rumelhart1988learning}
D.~E. Rumelhart, G.~E. Hinton, and R.~J. Williams, ``Learning representations
  by back-propagating errors,'' \emph{Cognitive modeling}, vol.~5, no.~3, p.~1,
  1988.

\bibitem{werbos1990backpropagation}
P.~J. Werbos, ``Backpropagation through time: what it does and how to do it,''
  \emph{Proceedings of the IEEE}, vol.~78, no.~10, pp. 1550--1560, 1990.

\bibitem{hochreiter1997long}
S.~Hochreiter and J.~Schmidhuber, ``Long short-term memory,'' \emph{Neural
  computation}, vol.~9, no.~8, pp. 1735--1780, 1997.

\bibitem{Sutskever2011Generating}
I.~Sutskever, J.~Martens, and G.~E. Hinton, ``Generating text with recurrent
  neural networks,'' in \emph{ICML}, 2011.

\bibitem{Cho2014On}
K.~Cho, B.~Van~Merri{\"e}nboer, D.~Bahdanau, and Y.~Bengio, ``On the properties
  of neural machine translation: Encoder-decoder approaches,'' \emph{arXiv
  preprint arXiv:1409.1259}, 2014.

\bibitem{Graves2014Towards}
A.~Graves and N.~Jaitly, ``Towards end-to-end speech recognition with recurrent
  neural networks,'' in \emph{ICML}, 2014.

\bibitem{Ng15}
J.~Yue-Hei~Ng, M.~Hausknecht, S.~Vijayanarasimhan, O.~Vinyals, R.~Monga, and
  G.~Toderici, ``Beyond short snippets: Deep networks for video
  classification,'' in \emph{CVPR}, 2015, pp. 4694--4702.

\bibitem{donahue2015long}
J.~Donahue, L.~Anne~Hendricks, S.~Guadarrama, M.~Rohrbach, S.~Venugopalan,
  K.~Saenko, and T.~Darrell, ``Long-term recurrent convolutional networks for
  visual recognition and description,'' in \emph{CVPR}, 2015, pp. 2625--2634.

\bibitem{pascanu2013number}
R.~Pascanu, G.~Montufar, and Y.~Bengio, ``On the number of response regions of
  deep feed forward networks with piece-wise linear activations,'' \emph{arXiv
  preprint arXiv:1312.6098}, 2013.

\bibitem{Sutskever2014Sequence}
I.~Sutskever, O.~Vinyals, and Q.~V. Le, ``Sequence to sequence learning with
  neural networks,'' in \emph{NeurIPS}, 2014, pp. 3104--3112.

\bibitem{bengio2015scheduled}
S.~Bengio, O.~Vinyals, N.~Jaitly, and N.~Shazeer, ``Scheduled sampling for
  sequence prediction with recurrent neural networks,'' in \emph{NeurIPS},
  2015, pp. 1171--1179.

\bibitem{shi2017deep}
X.~Shi, Z.~Gao, L.~Lausen, H.~Wang, D.-Y. Yeung, W.-k. Wong, and W.-c. Woo,
  ``Deep learning for precipitation nowcasting: A benchmark and a new model,''
  in \emph{NeurIPS}, 2017, pp. 5617--5627.

\bibitem{Finn2016Unsupervised}
C.~Finn, I.~Goodfellow, and S.~Levine, ``Unsupervised learning for physical
  interaction through video prediction,'' in \emph{NeurIPS}, 2016, pp. 64--72.

\bibitem{wang2018predrnn++}
Y.~Wang, Z.~Gao, M.~Long, J.~Wang, and P.~S. Yu, ``{PredRNN}++: Towards a
  resolution of the deep-in-time dilemma in spatiotemporal predictive
  learning,'' in \emph{ICML}, 2018, pp. 5123--5132.

\bibitem{yu2020efficient}
W.~Yu, Y.~Lu, S.~Easterbrook, and S.~Fidler, ``Efficient and
  information-preserving future frame prediction and beyond,'' in \emph{ICLR},
  2020.

\bibitem{su2020convolutional}
J.~Su, W.~Byeon, F.~Huang, J.~Kautz, and A.~Anandkumar, ``Convolutional
  tensor-train {LSTM} for spatio-temporal learning,'' in \emph{NeurIPS}, 2020.

\bibitem{oprea2020review}
S.~Oprea, P.~Martinez-Gonzalez, A.~Garcia-Garcia, J.~A. Castro-Vargas,
  S.~Orts-Escolano, J.~Garcia-Rodriguez, and A.~Argyros, ``A review on deep
  learning techniques for video prediction,'' \emph{IEEE Transactions on
  Pattern Analysis and Machine Intelligence}, 2020.

\bibitem{Oh2015Action}
J.~Oh, X.~Guo, H.~Lee, R.~L. Lewis, and S.~Singh, ``Action-conditional video
  prediction using deep networks in atari games,'' in \emph{NeurIPS}, 2015, pp.
  2863--2871.

\bibitem{Mathieu2015Deep}
M.~Mathieu, C.~Couprie, and Y.~LeCun, ``Deep multi-scale video prediction
  beyond mean square error,'' in \emph{ICLR}, 2016.

\bibitem{tulyakov2018mocogan}
S.~Tulyakov, M.-Y. Liu, X.~Yang, and J.~Kautz, ``{M}o{C}o{GAN}: Decomposing
  motion and content for video generation,'' in \emph{CVPR}, 2018, pp.
  1526--1535.

\bibitem{srivastava2015unsupervised}
N.~Srivastava, E.~Mansimov, and R.~Salakhudinov, ``Unsupervised learning of
  video representations using {LSTM}s,'' in \emph{ICML}, 2015, pp. 843--852.

\bibitem{babaeizadeh2017stochastic}
M.~Babaeizadeh, C.~Finn, D.~Erhan, R.~H. Campbell, and S.~Levine, ``Stochastic
  variational video prediction,'' in \emph{ICLR}, 2018.

\bibitem{weissenborn2019scaling}
D.~Weissenborn, O.~T{\"a}ckstr{\"o}m, and J.~Uszkoreit, ``Scaling
  autoregressive video models,'' in \emph{ICLR}, 2019.

\bibitem{kumar2019videoflow}
M.~Kumar, M.~Babaeizadeh, D.~Erhan, C.~Finn, S.~Levine, L.~Dinh, and D.~Kingma,
  ``Videoflow: A flow-based generative model for video,'' in \emph{ICLR}, 2020.

\bibitem{xue2016visual}
T.~Xue, J.~Wu, K.~Bouman, and B.~Freeman, ``Visual dynamics: Probabilistic
  future frame synthesis via cross convolutional networks,'' in \emph{NeurIPS},
  2016.

\bibitem{zhang2017deep}
J.~Zhang, Y.~Zheng, and D.~Qi, ``Deep spatio-temporal residual networks for
  citywide crowd flows prediction.'' in \emph{AAAI}, 2017, pp. 1655--1661.

\bibitem{Goodfellow2014Generative}
I.~Goodfellow, J.~Pouget-Abadie, M.~Mirza, B.~Xu, D.~Warde-Farley, S.~Ozair,
  A.~Courville, and Y.~Bengio, ``Generative adversarial nets,'' in
  \emph{NeurIPS}, 2014, pp. 2672--2680.

\bibitem{Denton2015Deep}
E.~Denton, S.~Chintala, R.~Fergus \emph{et~al.}, ``Deep generative image models
  using a {L}aplacian pyramid of adversarial networks,'' in \emph{NeurIPS},
  2015, pp. 1486--1494.

\bibitem{bhattacharjee2017temporal}
P.~Bhattacharjee and S.~Das, ``Temporal coherency based criteria for predicting
  video frames using deep multi-stage generative adversarial networks,'' in
  \emph{NeurIPS}, 2017, pp. 4271--4280.

\bibitem{liang2017dual}
X.~Liang, L.~Lee, W.~Dai, and E.~P. Xing, ``Dual motion {GAN} for future-flow
  embedded video prediction,'' in \emph{ICCV}, 2017, pp. 1744--1752.

\bibitem{wu2020future}
Y.~Wu, R.~Gao, J.~Park, and Q.~Chen, ``Future video synthesis with object
  motion prediction,'' in \emph{CVPR}, 2020, pp. 5539--5548.

\bibitem{gur2020hierarchical}
S.~Gur, S.~Benaim, and L.~Wolf, ``Hierarchical patch vae-gan: Generating
  diverse videos from a single sample,'' in \emph{NeurIPS}, 2020.

\bibitem{liu2021deep}
B.~Liu, Y.~Chen, S.~Liu, and H.-S. Kim, ``Deep learning in latent space for
  video prediction and compression,'' in \emph{CVPR}, 2021, pp. 701--710.

\bibitem{Ranzato2014Video}
M.~Ranzato, A.~Szlam, J.~Bruna, M.~Mathieu, R.~Collobert, and S.~Chopra,
  ``Video (language) modeling: a baseline for generative models of natural
  videos,'' \emph{arXiv preprint arXiv:1412.6604}, 2014.

\bibitem{villegas2017learning}
R.~Villegas, J.~Yang, Y.~Zou, S.~Sohn, X.~Lin, and H.~Lee, ``Learning to
  generate long-term future via hierarchical prediction,'' in \emph{ICML},
  2017, pp. 3560--3569.

\bibitem{wichers2018hierarchical}
N.~Wichers, R.~Villegas, D.~Erhan, and H.~Lee, ``Hierarchical long-term video
  prediction without supervision,'' in \emph{ICML}, 2018, pp. 6038--6046.

\bibitem{kim2019variational}
T.~Kim, S.~Ahn, and Y.~Bengio, ``Variational temporal abstraction,''
  \emph{NeurIPS}, vol.~32, pp. 11\,570--11\,579, 2019.

\bibitem{denton2018stochastic}
E.~Denton and R.~Fergus, ``Stochastic video generation with a learned prior,''
  in \emph{ICML}, 2018, pp. 1174--1183.

\bibitem{lee2018stochastic}
A.~X. Lee, R.~Zhang, F.~Ebert, P.~Abbeel, C.~Finn, and S.~Levine, ``Stochastic
  adversarial video prediction,'' \emph{arXiv preprint arXiv:1804.01523}, 2018.

\bibitem{villegas2019high}
R.~Villegas, A.~Pathak, H.~Kannan, D.~Erhan, Q.~V. Le, and H.~Lee, ``High
  fidelity video prediction with large stochastic recurrent neural networks,''
  in \emph{NeurIPS}, 2019, pp. 81--91.

\bibitem{castrejon2019improved}
L.~Castrejon, N.~Ballas, and A.~Courville, ``Improved conditional vrnns for
  video prediction,'' in \emph{ICCV}, 2019, pp. 7608--7617.

\bibitem{franceschi2020stochastic}
J.-Y. Franceschi, E.~Delasalles, M.~Chen, S.~Lamprier, and P.~Gallinari,
  ``Stochastic latent residual video prediction,'' in \emph{ICML}, 2020, pp.
  3233--3246.

\bibitem{wu2021greedy}
B.~Wu, S.~Nair, R.~Martin-Martin, L.~Fei-Fei, and C.~Finn, ``Greedy
  hierarchical variational autoencoders for large-scale video prediction,'' in
  \emph{CVPR}, 2021, pp. 2318--2328.

\bibitem{Villegas2017Decomposing}
R.~Villegas, J.~Yang, S.~Hong, X.~Lin, and H.~Lee, ``Decomposing motion and
  content for natural video sequence prediction,'' in \emph{ICLR}, 2017.

\bibitem{bei2021learning}
X.~Bei, Y.~Yang, and S.~Soatto, ``Learning semantic-aware dynamics for video
  prediction,'' in \emph{CVPR}, 2021, pp. 902--912.

\bibitem{denton2017unsupervised}
E.~Denton and V.~Birodkar, ``Unsupervised learning of disentangled
  representations from video,'' in \emph{NeurIPS}, 2017, pp. 4417--4426.

\bibitem{hsieh2018learning}
J.-T. Hsieh, B.~Liu, D.-A. Huang, L.~F. Fei-Fei, and J.~C. Niebles, ``Learning
  to decompose and disentangle representations for video prediction,'' in
  \emph{NeurIPS}, 2018, pp. 517--526.

\bibitem{bodla2021hierarchical}
N.~Bodla, G.~Shrivastava, R.~Chellappa, and A.~Shrivastava, ``Hierarchical
  video prediction using relational layouts for human-object interactions,'' in
  \emph{CVPR}, 2021.

\bibitem{zablotskaia2020unsupervised}
P.~Zablotskaia, E.~A. Dominici, L.~Sigal, and A.~M. Lehrmann, ``Unsupervised
  video decomposition using spatio-temporal iterative inference,'' \emph{arXiv
  preprint arXiv:2006.14727}, 2020.

\bibitem{greff2019multi}
K.~Greff, R.~L. Kaufman, R.~Kabra, N.~Watters, C.~Burgess, D.~Zoran,
  L.~Matthey, M.~Botvinick, and A.~Lerchner, ``Multi-object representation
  learning with iterative variational inference,'' in \emph{International
  Conference on Machine Learning}.\hskip 1em plus 0.5em minus 0.4em\relax PMLR,
  2019, pp. 2424--2433.

\bibitem{guen2020disentangling}
V.~L. Guen and N.~Thome, ``Disentangling physical dynamics from unknown factors
  for unsupervised video prediction,'' in \emph{CVPR}, 2020.

\bibitem{patraucean2015spatio}
V.~Patraucean, A.~Handa, and R.~Cipolla, ``Spatio-temporal video autoencoder
  with differentiable memory,'' in \emph{ICLR Workshop}, 2016.

\bibitem{Lotter2016Deep}
W.~Lotter, G.~Kreiman, and D.~Cox, ``Deep predictive coding networks for video
  prediction and unsupervised learning,'' in \emph{ICLR}, 2017.

\bibitem{Kalchbrenner2016Video}
N.~Kalchbrenner, A.~Oord, K.~Simonyan, I.~Danihelka, O.~Vinyals, A.~Graves, and
  K.~Kavukcuoglu, ``Video pixel networks,'' in \emph{ICML}, 2017, pp.
  1771--1779.

\bibitem{byeon2018contextvp}
W.~Byeon, Q.~Wang, R.~Kumar~Srivastava, and P.~Koumoutsakos, ``Contextvp: Fully
  context-aware video prediction,'' in \emph{ECCV}, 2018, pp. 753--769.

\bibitem{oliu2018folded}
M.~Oliu, J.~Selva, and S.~Escalera, ``Folded recurrent neural networks for
  future video prediction,'' in \emph{ECCV}, 2018, pp. 716--731.

\bibitem{xu2018structure}
J.~Xu, B.~Ni, Z.~Li, S.~Cheng, and X.~Yang, ``Structure preserving video
  prediction,'' in \emph{CVPR}, 2018, pp. 1460--1469.

\bibitem{wu2021motionrnn}
H.~Wu, Z.~Yao, J.~Wang, and M.~Long, ``Motionrnn: A flexible model for video
  prediction with spacetime-varying motions,'' in \emph{CVPR}, 2021.

\bibitem{graves2014neural}
A.~Graves, G.~Wayne, and I.~Danihelka, ``Neural turing machines,'' \emph{arXiv
  preprint arXiv:1410.5401}, 2014.

\bibitem{sukhbaatar2015end}
S.~Sukhbaatar, J.~Weston, R.~Fergus \emph{et~al.}, ``End-to-end memory
  networks,'' in \emph{NeurIPS}, 2015, pp. 2440--2448.

\bibitem{graves2016hybrid}
A.~Graves, G.~Wayne, M.~Reynolds, T.~Harley, I.~Danihelka,
  A.~Grabska-Barwi{\'n}ska, S.~G. Colmenarejo, E.~Grefenstette, T.~Ramalho,
  J.~Agapiou \emph{et~al.}, ``Hybrid computing using a neural network with
  dynamic external memory,'' \emph{Nature}, vol. 538, no. 7626, pp. 471--476,
  2016.

\bibitem{hinton1984distributed}
G.~E. Hinton, ``Distributed representations,'' 1984.

\bibitem{bengio2000taking}
S.~Bengio and Y.~Bengio, ``Taking on the curse of dimensionality in joint
  distributions using neural networks,'' \emph{IEEE Transactions on Neural
  Networks}, vol.~11, no.~3, pp. 550--557, 2000.

\bibitem{maaten2008visualizing}
L.~v.~d. Maaten and G.~Hinton, ``Visualizing data using t-{SNE},''
  \emph{Journal of machine learning research}, vol.~9, no. Nov, pp. 2579--2605,
  2008.

\bibitem{krogh1995neural}
A.~Krogh and J.~Vedelsby, ``Neural network ensembles, cross validation, and
  active learning,'' in \emph{NeurIPS}, 1995, pp. 231--238.

\bibitem{zhou2009ensemble}
Z.-H. Zhou, ``Ensemble learning.'' \emph{Encyclopedia of biometrics}, vol.~1,
  pp. 270--273, 2009.

\bibitem{chiappa2017recurrent}
S.~Chiappa, S.~Racaniere, D.~Wierstra, and S.~Mohamed, ``Recurrent environment
  simulators,'' in \emph{ICLR}, 2017.

\bibitem{hafner2020dream}
D.~Hafner, T.~Lillicrap, J.~Ba, and M.~Norouzi, ``Dream to control: Learning
  behaviors by latent imagination,'' in \emph{ICLR}, 2020.

\bibitem{Sch2004Recognizing}
C.~Schuldt, I.~Laptev, and B.~Caputo, ``Recognizing human actions: a local svm
  approach,'' in \emph{ICPR}, 2004.

\bibitem{Traffic4cast-web-2019}
IARAI, ``Traffic4cast 2019: Traffic map movie forecasting.''
  \url{https://www.iarai.ac.at/traffic4cast/2019-competition/}, 2019.

\bibitem{Kingma2014Adam}
D.~Kingma and J.~Ba, ``Adam: A method for stochastic optimization,'' in
  \emph{ICLR}, 2015.

\bibitem{Wang2004Image}
Z.~Wang, A.~C. Bovik, H.~R. Sheikh, and E.~P. Simoncelli, ``Image quality
  assessment: from error visibility to structural similarity,'' \emph{IEEE
  transactions on image processing}, vol.~13, no.~4, pp. 600--612, 2004.

\bibitem{zhang2018unreasonable}
R.~Zhang, P.~Isola, A.~A. Efros, E.~Shechtman, and O.~Wang, ``The unreasonable
  effectiveness of deep features as a perceptual metric,'' in \emph{CVPR},
  2018, pp. 586--595.

\bibitem{de2016dynamic}
B.~De~Brabandere, X.~Jia, T.~Tuytelaars, and L.~Van~Gool, ``Dynamic filter
  networks,'' in \emph{NeurIPS}, 2016, pp. 667--675.

\bibitem{ronneberger2015u}
O.~Ronneberger, P.~Fischer, and T.~Brox, ``U-{N}et: Convolutional networks for
  biomedical image segmentation,'' in \emph{MICCAI}, 2015.

\end{thebibliography}

%



\vspace{-30pt}
\begin{IEEEbiography}[{\includegraphics[width=1in,clip,keepaspectratio]{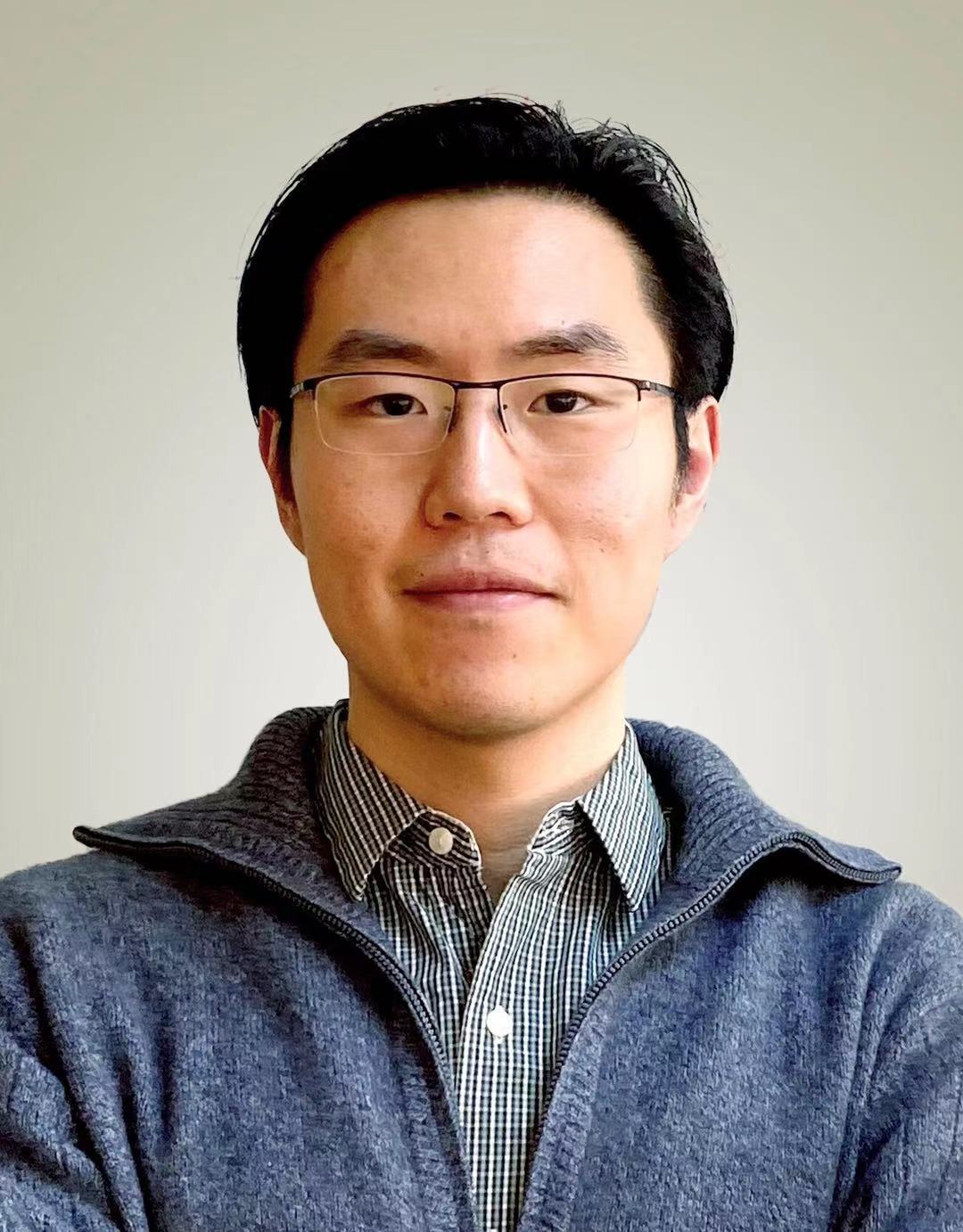}}]{Yunbo Wang} received the BE degree from Xi'an Jiaotong University in 2012, and the ME and PhD degrees from Tsinghua University in 2015 and 2020. He received the CCF Outstanding Doctoral Dissertation Award in 2020, advised by Philip S. Yu and Mingsheng Long. He is now an assistant professor at the AI Institute and the Department of Computer Science at Shanghai Jiao Tong University. He does research in deep learning, especially predictive learning, spatiotemporal modeling, and model-based decision making. 
\end{IEEEbiography}

\vspace{-30pt}
\begin{IEEEbiography}[{\includegraphics[width=1in,height=1.25in,clip,keepaspectratio]{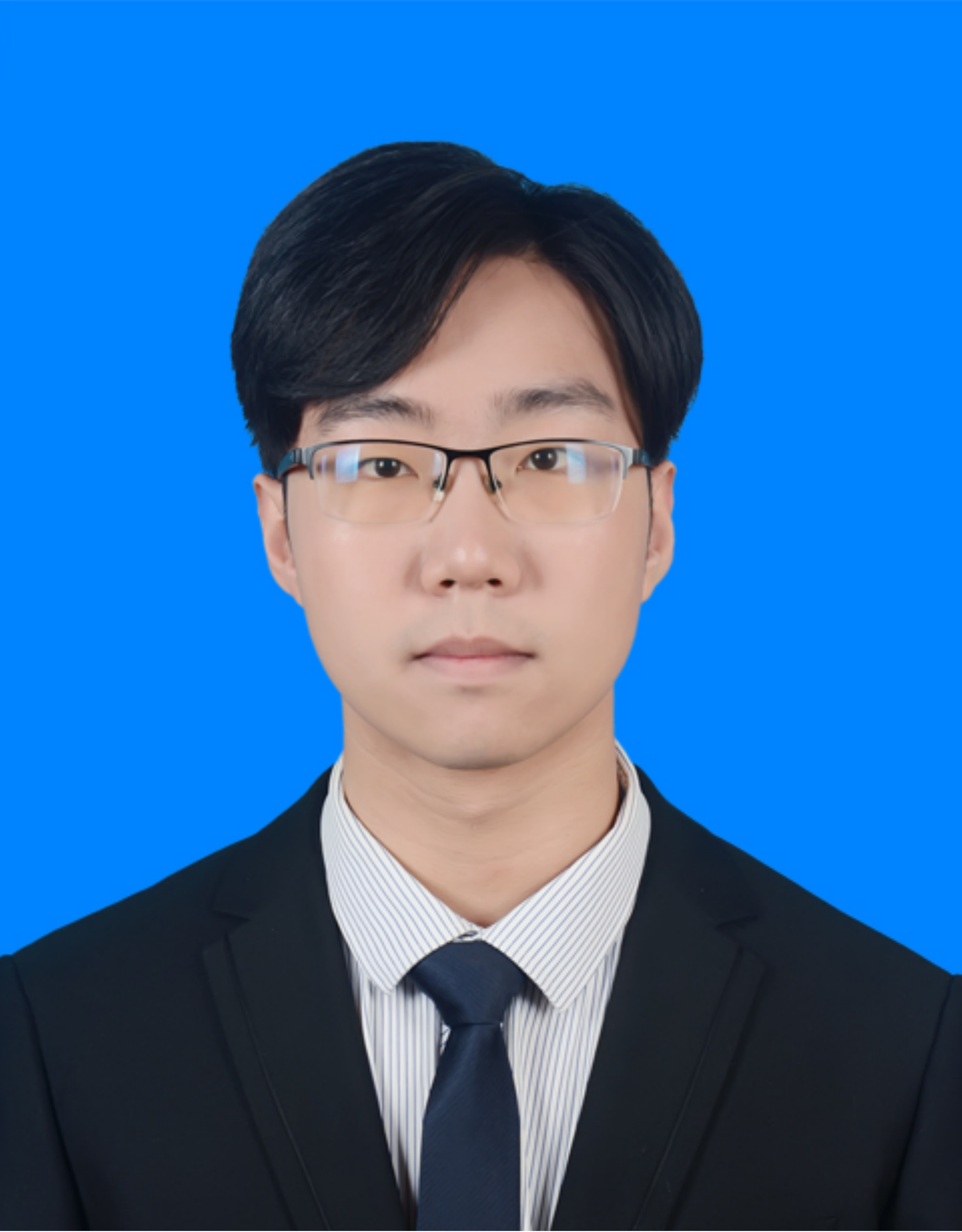}}]{Haixu Wu} received the BE degree in computer software from Tsinghua University in 2020. He is working towards the ME degree in computer software at Tsinghua University. His research interests include machine learning and computer vision.
\end{IEEEbiography}

\vspace{-30pt}
\begin{IEEEbiography}[{\includegraphics[width=1in,height=1.25in,clip,keepaspectratio]{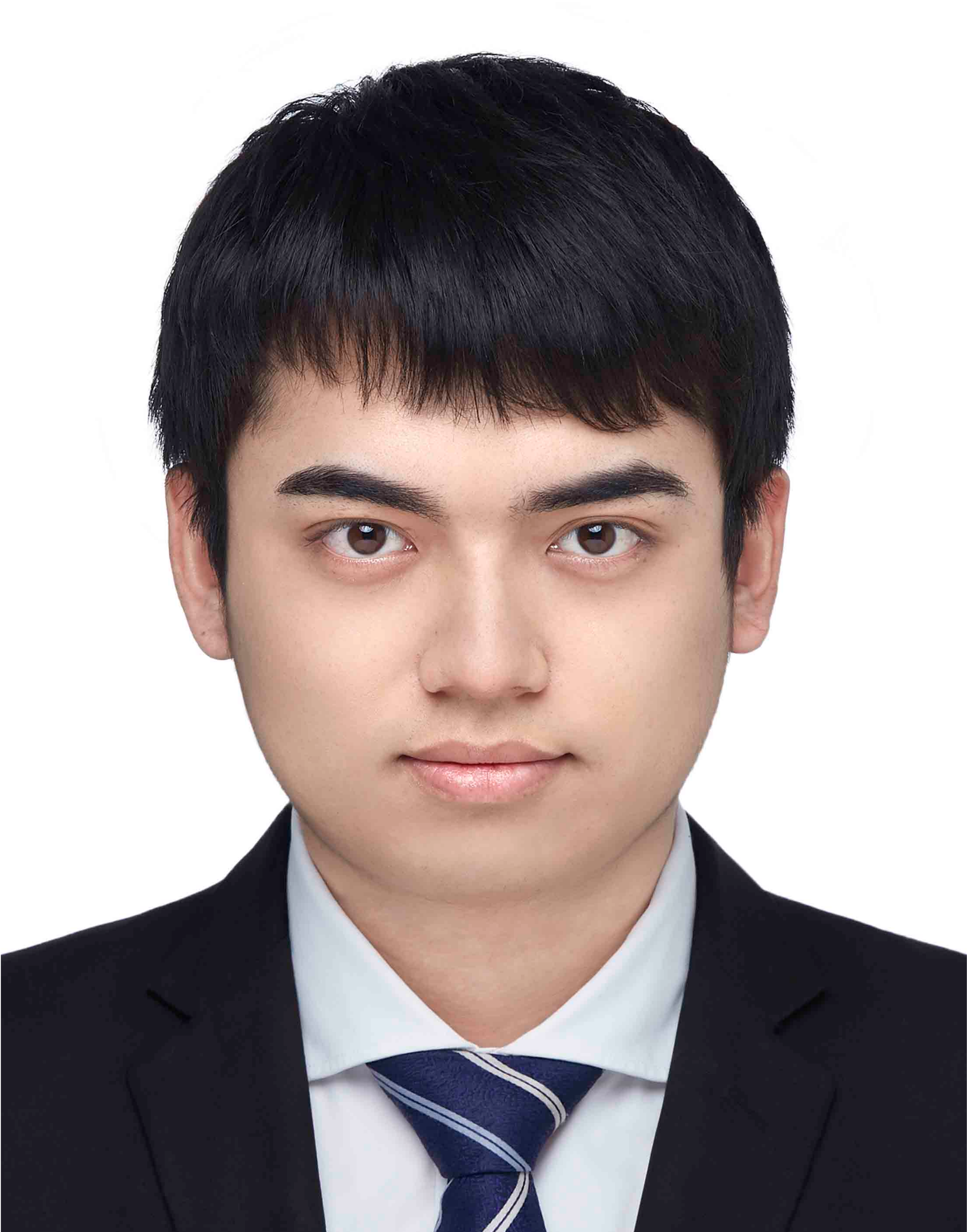}}]{Jianjin Zhang} received the ME degree in computer software from Tsinghua University in 2019. His research interests lie on the intersection of machine learning, time series analysis, and natural language processing.
\end{IEEEbiography}

\vspace{-30pt}
\begin{IEEEbiography}[{\includegraphics[width=1in,height=1.25in,clip,keepaspectratio]{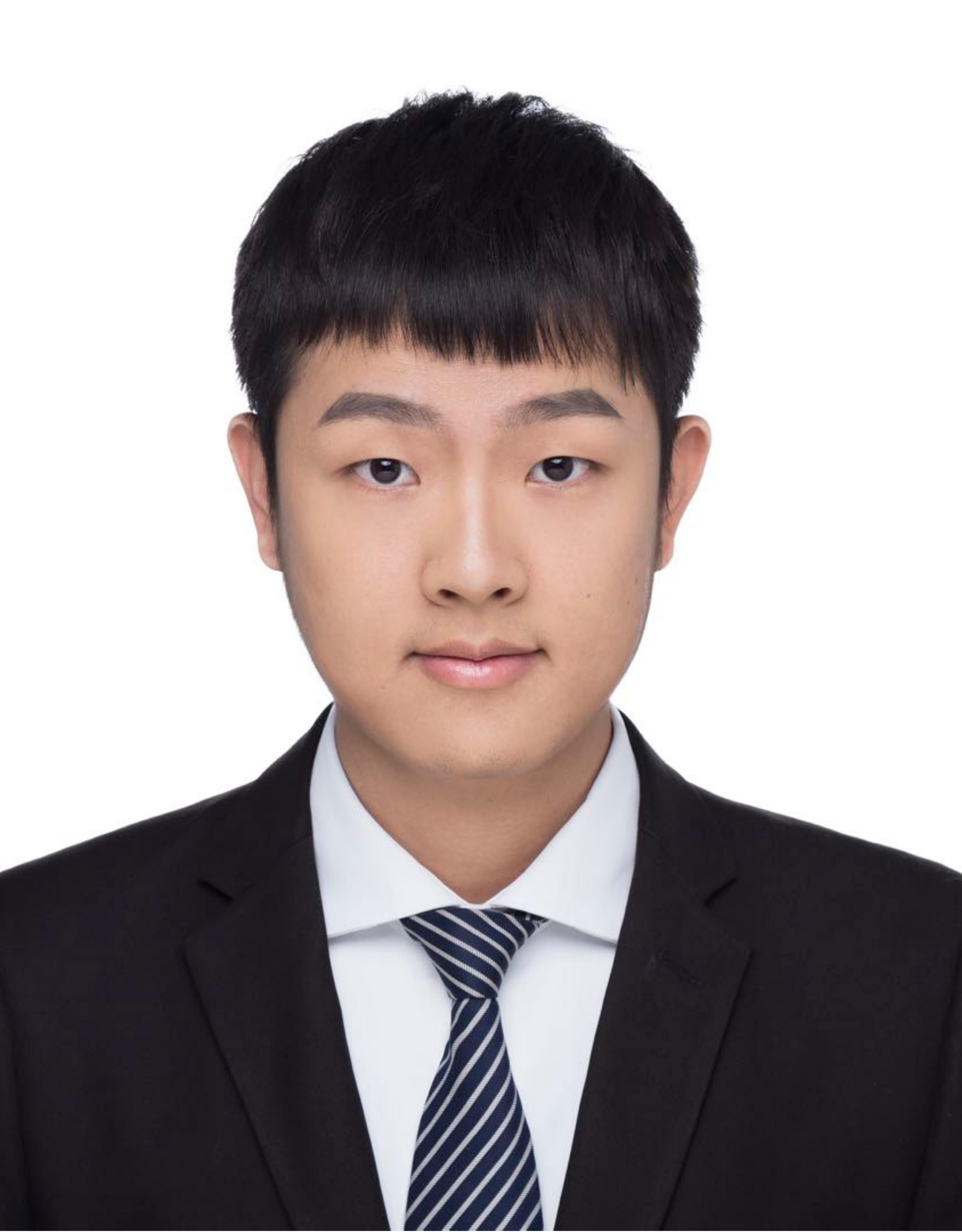}}]{Zhifeng Gao} received the ME degree in computer software from Tsinghua University in 2019. His research interests include machine learning and big data systems.
\end{IEEEbiography}

\vspace{-30pt}
\begin{IEEEbiography}[{\includegraphics[width=1in,height=1.25in,clip,keepaspectratio]{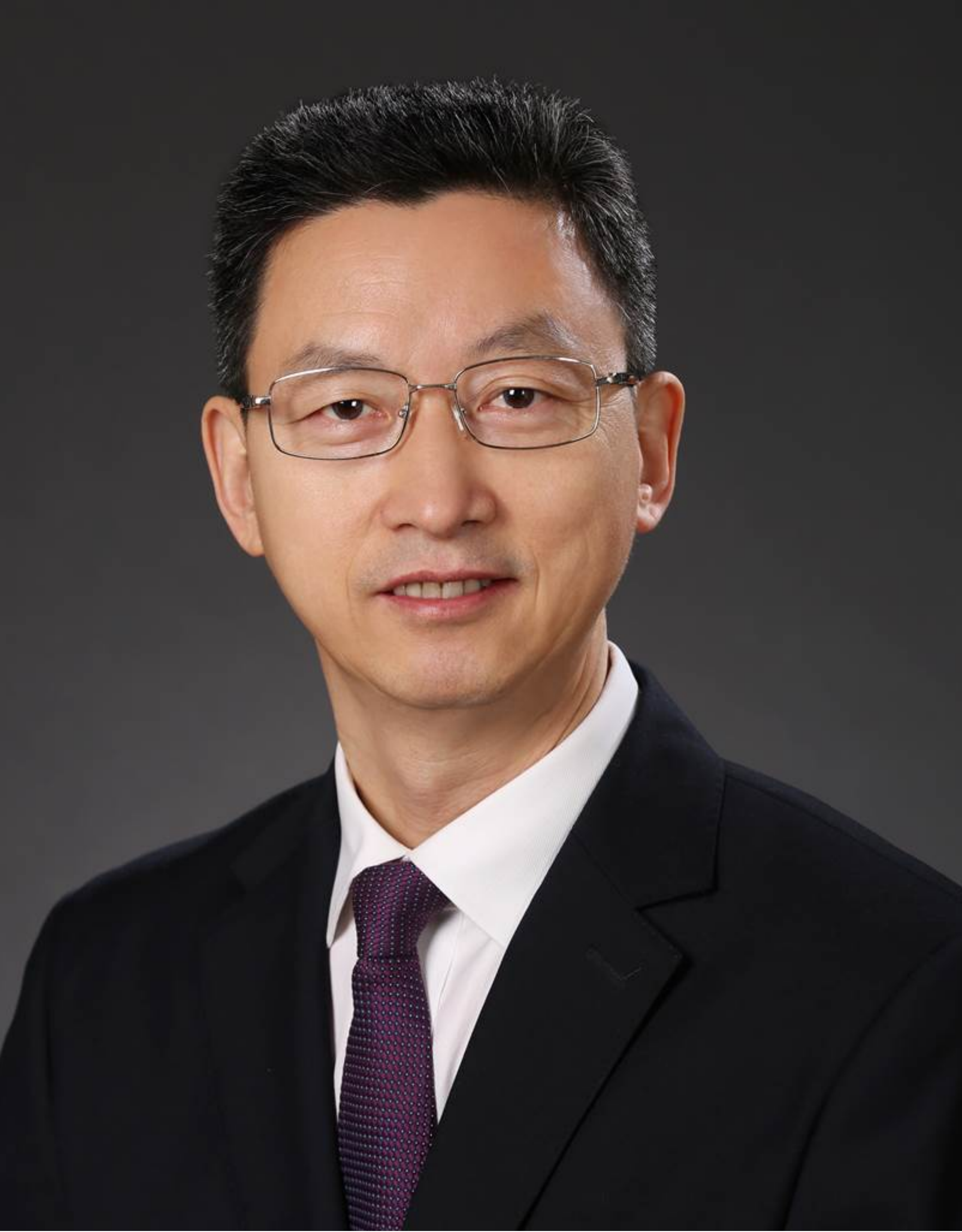}}]{Jianmin Wang} received the BE degree from Peking University, China, in 1990, and the ME and PhD degrees in computer software from Tsinghua University, China, in 1992 and 1995,
respectively. He is a full professor with the School
of Software, Tsinghua University. His research interests include big data management systems and large-scale data analytics. He led to develop a product data \& lifecycle management system, which has been deployed in hundreds of enterprises in China. He is leading to develop a big data platform in the National Engineering Lab for Big Data Software.
\end{IEEEbiography}

\vspace{-30pt}
\begin{IEEEbiography}[{\includegraphics[width=1in,height=1.25in,clip,keepaspectratio]{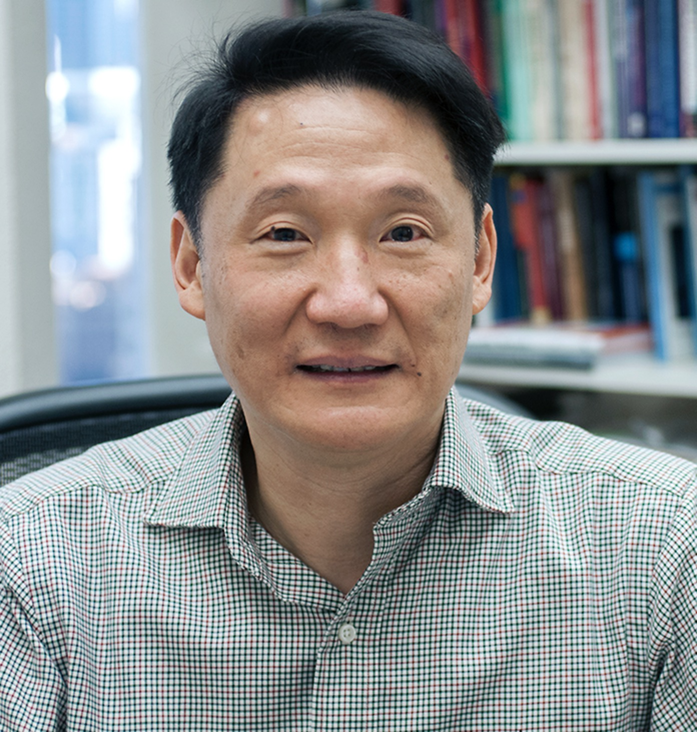}}]{Philip S. Yu} received his BS in electrical engineering from the National Taiwan University, and his MS and PhD also in electrical engineering from Stanford University in 1978. He is a Distinguished Professor at the University of Illinois at Chicago and Tsinghua University. Yu holds over 300 US patents, is ACM Fellow and IEEE Fellow, is Editor-in-Chief of ACM Transactions on Knowledge Discovery from Data, and has been awarded several awards by IBM and the IEEE. Yu's research interests are in the fields of data mining, social network, privacy preserving data publishing, data stream, database systems, and Internet applications and technologies. 
\end{IEEEbiography}

\vspace{-30pt}
\begin{IEEEbiography}[{\includegraphics[width=1in,height=1.25in,clip,keepaspectratio]{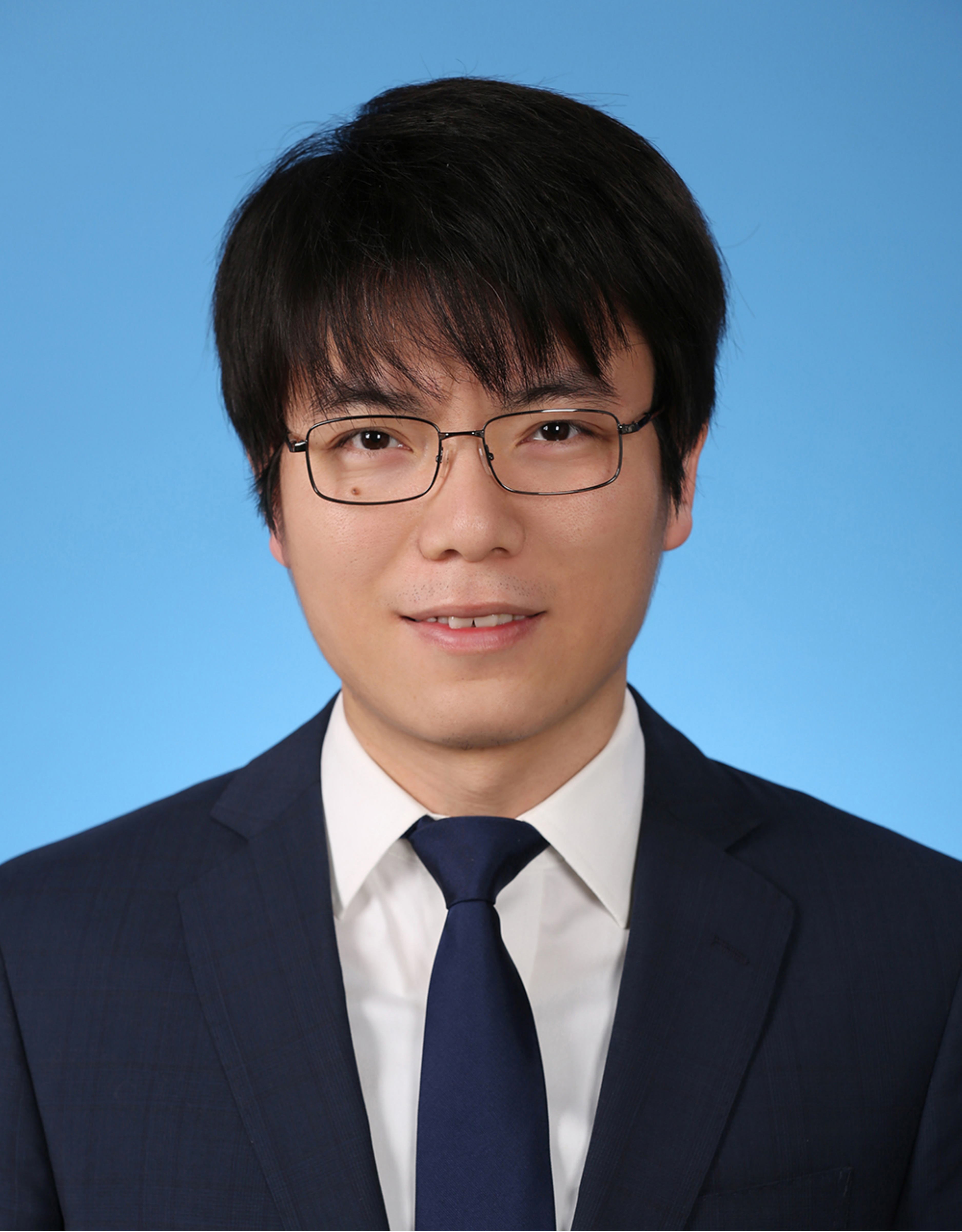}}]{Mingsheng Long} received the BE degree in electrical engineering and the PhD degree in computer science from Tsinghua University in 2008 and 2014 respectively. He was a visiting researcher in computer science, UC Berkeley from 2014 to 2015. He is an associate professor with the School of Software, Tsinghua University. He serves as Area Chairs of major machine learning conferences (ICML, NeurIPS, and ICLR). His research is persistently dedicated to theories and algorithms of machine learning, with special interests in transfer, adaptation, and data-efficient learning, foundation deep models, learning with spatiotemporal and scientific knowledge.
\end{IEEEbiography}

\end{document}